\documentclass{article} % For LaTeX2e
\usepackage{iclr2026_conference,times}

\usepackage{amsmath,amsfonts,bm}

\def\eqref#1{equation~\ref{#1}}
\def\1{\bm{1}}

\DeclareMathAlphabet{\mathsfit}{\encodingdefault}{\sfdefault}{m}{sl}
\SetMathAlphabet{\mathsfit}{bold}{\encodingdefault}{\sfdefault}{bx}{n}

\usepackage{hyperref}
\usepackage{url}

\usepackage{dsfont}
\usepackage{graphicx}

\usepackage[utf8]{inputenc} % allow utf-8 input
\usepackage[T1]{fontenc}    % use 8-bit T1 fonts
\usepackage{hyperref}       % hyperlinks
\usepackage{url}            % simple URL typesetting
\usepackage{booktabs}       % professional-quality tables
\usepackage[linesnumbered,ruled,vlined]{algorithm2e} 
\usepackage{amsfonts}       % blackboard math symbols
\usepackage{nicefrac}       % compact symbols for 1/2, etc.
\usepackage{microtype}      % microtypography
\usepackage{longtable}
\usepackage{times}
\usepackage{latexsym}
\usepackage[table,xcdraw]{xcolor}
\usepackage{cuted}

\usepackage{makecell}
\usepackage{multirow}
\usepackage{hyperref}
\usepackage{url}
\usepackage{subcaption}
\usepackage{algorithmic}
\usepackage{amssymb}
\usepackage[most]{tcolorbox}
\usepackage{pifont}
\usepackage{color}
\usepackage{amsmath}
\usepackage{makecell}
\usepackage{newfloat}
\usepackage{listings}
\usepackage{afterpage}
\usepackage{colortbl} 
\usepackage{wrapfig}
\usepackage{soul}  
\usepackage{threeparttable}
\usepackage{array}    
\usepackage{ragged2e}  
\usepackage{longtable}
\usepackage{tabularx}

\usepackage{rotating}

\usepackage{array} % for newcolumntype

\newcolumntype{L}[1]{>{\RaggedRight\arraybackslash}m{#1}} % 左对齐 + 垂直居中
\newcolumntype{C}[1]{>{\centering\arraybackslash}m{#1}}  % 居中 + 垂直居中
\newcolumntype{R}[1]{>{\RaggedLeft\arraybackslash}m{#1}} % 右对齐 + 垂直居中

\definecolor{c1}{HTML}{003371}
\definecolor{c2}{HTML}{057748}

\iclrfinalcopy
\title{EduPersona: Benchmarking Subjective Ability Boundaries of Virtual Student Agents}

\author{\textbf{Buyuan Zhu}$^{1*}$, \ 
\textbf{Shiyu Hu}$^{1*}$, \ 
\textbf{Yiping Ma}$^{2,3}$, \ 
\textbf{Yuanming Zhang}$^{4}$, \ 
\textbf{Kang Hao Cheong}$^{1,5\dagger}$\\
\textsuperscript{1}School of Physical and Mathematical Sciences, Nanyang Technological University,\\
\textsuperscript{2}Lab of Artificial Intelligence for Education, East China Normal University,\\
\textsuperscript{3}School of Computer Science and Technology, East China Normal University,\\
\textsuperscript{4}State Key Laboratory of Robotics and Systems, Harbin Institute of Technology\\
\textsuperscript{5}College of Computing and Data Science, Nanyang Technological University\\
\tt\small buyuan001@e.ntu.edu.sg \quad shiyu.hu@ntu.edu.sg \\
\tt\small 52275901020@stu.ecnu.edu.cn \quad zym@alu.hit.edu.cn \\
\tt\small kanghao.cheong@ntu.edu.sg\\
\footnotesize{$^{*}$ Equal Contribution \quad $^{\dagger}$ Corresponding Author}\\
}

\begin{document}

\maketitle

\begin{abstract}
As large language models (LLMs) are increasingly integrated into education, virtual student agents are becoming vital for classroom simulation and teacher training. Yet their classroom-oriented subjective abilities remain largely unassessed, limiting understanding of model boundaries and hindering trustworthy deployment.  
We present \textbf{EduPersona}, a large-scale benchmark spanning two languages, three subjects, and ten persona types based on the Big Five theory. The dataset contains 1,308 authentic classroom dialogue rounds, corresponding to 12,814 teacher–student Q\&A turns, and is further expanded through persona stylization into roughly 10$\times$ larger scale (128k turns), providing a solid foundation for evaluation. Building on this resource, we decompose hard-to-quantify subjective performance into three progressive tasks: TASK1 basic coherence (whether behavior, emotion, expression, and voice align with classroom context), TASK2 student realism, and TASK3 long-term persona consistency, thereby establishing an evaluation framework grounded in educational theory and research value.  
We conduct systematic experiments on three representative LLMs, comparing their original versions with ten persona-fine-tuned variants trained on EduPersona. Results show consistent and significant average improvements across all tasks: TASK1 \textbf{33.6\% $\uparrow$}, TASK2 \textbf{30.6\% $\uparrow$}, and TASK3 \textbf{14.9\% $\uparrow$}. These improvements highlight the dataset’s effectiveness and research value, while also revealing the heterogeneous difficulty of persona modeling.
In summary, EduPersona delivers the first classroom benchmark centered on subjective abilities, establishes a decoupled and verifiable research paradigm, and we will open-source both the dataset and the framework to support the broader research community in advancing trustworthy and human-like AI for education.
\end{abstract}

\section{Introduction}
\label{sec:introduction}

With the rapid proliferation of large language models (LLMs) in education~\citep{wang2024large,tan2025artificial,tan2025comprehensive}, virtual student agents are emerging as key tools for classroom simulation and teacher training~\citep{dai2024designing}. They offer education researchers low-cost, controllable, and reproducible experimental environments, while giving the AI community new opportunities to examine human-like interaction and role-playing. Yet existing evaluation frameworks remain focused on objective tasks such as question answering and accuracy~\citep{lu2022learn,huang2023c,ang2023socratic}, overlooking the subjective abilities essential to classroom practice.
In authentic educational interactions, students must exhibit multiple layers of subjective traits~\cite{wang2024boosting,seo2025augmented}. First is \textbf{basic coherence}, aligning observable behaviors, emotions, expressions, and voice with linguistic outputs~\citep{hayashi2024modeling}. Second is \textbf{student realism}, reflected in naturalness, credible identity, and adherence to classroom norms such as admitting ignorance, requesting hints, or self-correcting~\citep{sanyal2025investigating}. Third is \textbf{persona consistency}, sustaining stable traits and stylistic patterns across both short- and long-term dialogues~\citep{ma2024llms}. Together, these layers form a progression—from external behaviors to perceptual authenticity and long-term stability—that mirrors the educational logic of “classroom performance → student traits → individual stability,” while providing AI research with a systematic framework for subjective ability evaluation.

\begin{figure*}[t!]
\centering
% \vspace{-10pt}
  \includegraphics[width=\textwidth]{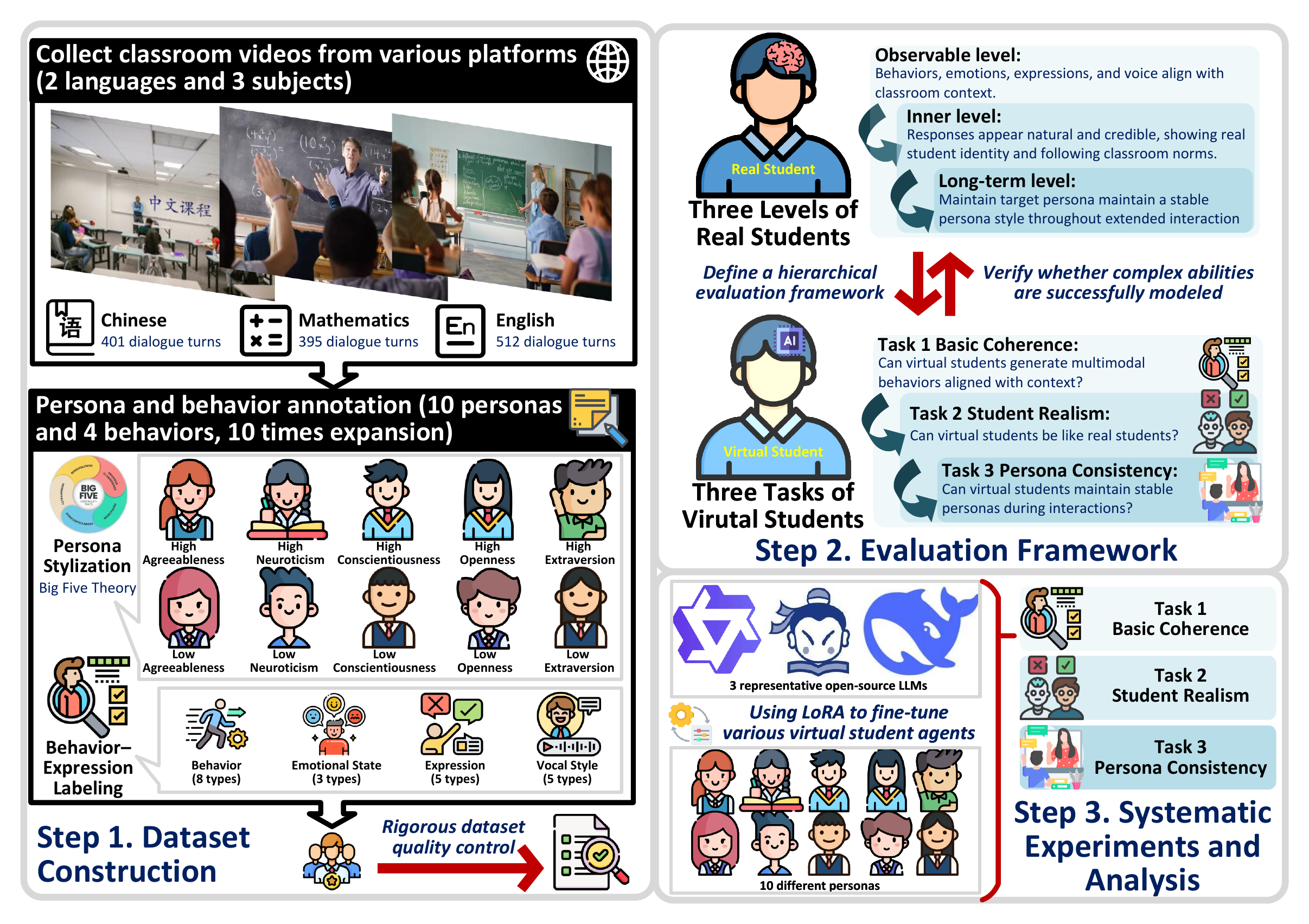}
  % \vspace{-15pt}
  \caption{\textbf{Workflow Overview of EduPersona.} It consists of three steps: (i) dataset construction (cross-subject and cross-lingual classroom dialogues with persona expansion and multimodal labeling, Sec.~\ref{sec:dataset}); (ii) a three-task evaluation framework (covering coherence, realism, and consistency, Sec.~\ref{sec:evaluation}); and (iii) systematic experiments and analysis (comparing original and fine-tuned models with cross-model comparisons and case studies, Sec.~\ref{sec:experiments}). 
  Together, these steps establish the first classroom benchmark focused on subjective abilities, systematically outlining the capability boundaries of virtual student agents.
  }
  \label{fig:overview}
  % \vspace{-12pt}
\end{figure*}

To address this challenge, we introduce \textbf{EduPersona}, the first large-scale benchmark spanning four dimensions: cross-lingual (Chinese and English), cross-subject (Chinese, Mathematics, and English), cross-behavior (four categories of classroom labels), and cross-persona (ten student traits extended from the Big Five Personality Theory). The dataset contains 1,308 authentic classroom dialogue rounds, corresponding to 12,814 teacher–student Q\&A turns. It is further expanded through persona stylization into roughly 10$\times$ larger scale (128k turns), providing a solid foundation for evaluation. The central innovation of EduPersona lies in decoupling subjective performance into three progressive evaluation tasks: starting from basic coherence, advancing to student realism, and culminating in long-term persona consistency. This design transforms traditionally unquantifiable subjective performance into an operational and reproducible measurement scheme (see Fig.~\ref{fig:overview}).

We conduct systematic experiments on three representative open-source LLMs—Qwen3 \citep{yang2025qwen3}, InternLM3 \citep{cai2024internlm2}, and DeepSeek-R1 \citep{guo2025deepseek}—comparing their original versions with ten EduPersona fine-tuned variants. Results show significant gains across all tasks: +33.6\% in basic coherence, +30.6\% in student realism, and +14.9\% in persona consistency. These findings validate EduPersona’s effectiveness and research value while highlighting the heterogeneous difficulty of persona modeling. Importantly, subjective abilities do not scale monotonically with model size or reasoning capacity but reveal independent challenges and capability gaps for virtual student agents in educational contexts.

\textbf{Contributions.}
(1) We construct the first large-scale classroom dataset covering two languages, three subjects, four categories of classroom behaviors, and ten persona traits based on extensions of the Big Five Personality Theory (Sec.~\ref{sec:dataset});
(2) We propose a progressive three-task evaluation framework that decomposes subjective abilities into coherence, realism, and consistency, transforming them into operational and comparable measures (Sec.~\ref{sec:evaluation});
(3) We conduct systematic experiments on multiple mainstream LLMs, showing both the improvements from fine-tuning and the persistent bottlenecks, while revealing that these abilities are not monotonically correlated with model scale or reasoning power (Sec.~\ref{sec:experiments}).
We will release EduPersona to support reproducibility and extension, advancing trustworthy and human-like virtual student research in both education and AI communities.

%%%%%%%%%%%%%%%%%%%%%%%%%%%%%%%%%

\section{Related Work}
\label{sec:related-work}

We review prior research through the lens of our three evaluation tasks and highlight why systematic modeling of virtual student agents requires going beyond existing approaches.  
(1) \textbf{Basic Behavioral Coherence:}  
Existing educational datasets (e.g., ScienceQA~\citep{lu2022learn}, C-Eval~\citep{huang2023c}, SocraticQ~\citep{ang2023socratic}, MathQA~\citep{amini2019mathqa}) have advanced knowledge assessment but remain largely single-turn or exam-oriented, lacking modeling of the IRF (Initiation–Response–Feedback) structure central to classrooms. Recent multimodal efforts explore VQA~\citep{lee2025multimodality,xiao2025eduvqa}, emotion recognition~\citep{song2025emotional}, and engagement detection~\citep{xie2025msc}, yet they focus on perception rather than coherence across verbal and non-verbal dimensions. Task~1 (Sec.~\ref{subsec:task1}) addresses this gap.
(2) \textbf{Student Realism:}  
Persona-driven dialogue studies such as PersonaChat~\citep{zhang2018personalizing}, PersonalDialog~\citep{zheng2019personalized}, and MBTI-based generation~\citep{kar2025convergence} illustrate role-conditioned generation, but they rely on simplified tags and are situated in open-domain settings. They cannot answer the classroom-specific question: does a model’s response resemble that of a real student? Task~2 (Sec.~\ref{subsec:task2}) formalizes this evaluation.
(3) \textbf{Persona Consistency:}  
Maintaining stable traits over long interactions remains challenging. Traditional metrics (BLEU, ROUGE) correlate poorly with persona preservation, and alignment methods (RLHF~\citep{ouyang2022training}, Constitutional AI~\citep{bai2022constitutional}) or bias detection~\citep{chen2024persona} provide only partial insights. Systematic evaluation of persona stability in classroom dialogue is still absent, which Task~3 (Sec.~\ref{subsec:task3}) directly operationalizes.
Overall, while prior work has progressed in knowledge testing, role-conditioned generation, and multimodal analytics, it lacks a unified, pedagogically grounded framework for jointly evaluating \textit{basic coherence}, \textit{student realism}, and \textit{persona consistency}. EduPersona is designed to fill this gap. 

\begin{figure*}[t!]
% \vspace{-45pt}
\centering
\includegraphics[width=\textwidth]{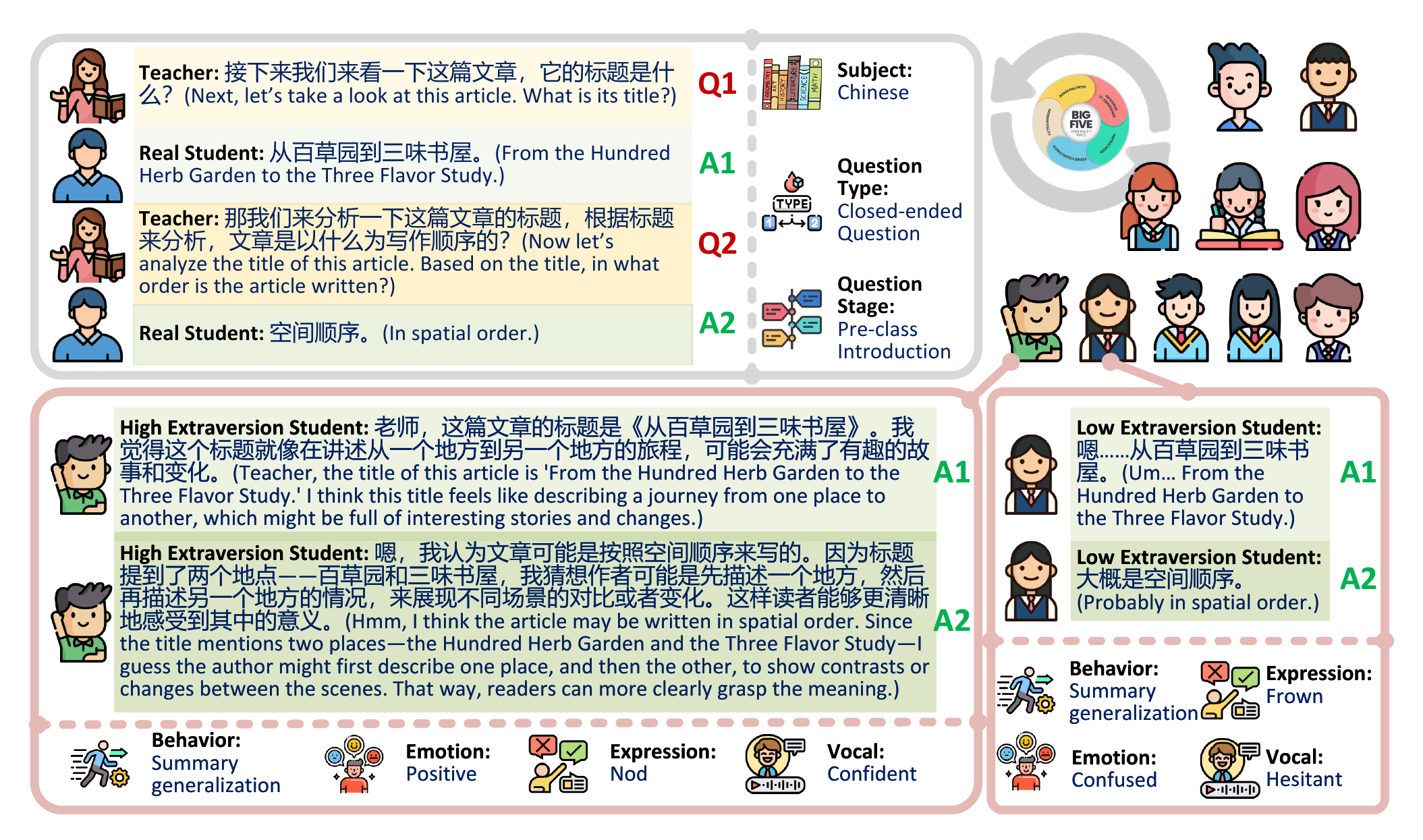}
% \vspace{-20pt}
\caption{\textbf{Chinese classroom example with persona-conditioned responses.} The top panel shows a real IRF snippet (with English translation), and the bottom presents virtual-student outputs under high/low extraversion with behavior–expression labels. This illustrates the EduPersona pipeline (raw dialogue $\rightarrow$ persona stylization $\rightarrow$ behavior–expression labeling) and highlights how different personas yield linguistic and non-verbal differences within the same teaching context.}
  \label{fig:example}
% \vspace{-12pt}
\end{figure*}

%%%%%%%%%%%%%%%%%%%%%%%%%%%%%%%%%
\section{Dataset Construction}
\label{sec:dataset}

This section introduces the construction and formal representation of the \textbf{EduPersona} dataset, which serves as the unified foundation for the subsequent evaluation tasks. We denote the final dataset as $\mathcal{D}=\{d_i\}$, $d_i=(x_i,\,y_i,\,p_i,\,b_i,\,s_i)$, where $x_i$ denotes the classroom context (teacher’s initiation and feedback), $y_i$ the student response, $p_i\in P$ the persona label, $b_i\in B$ the behavior–expression label, and $s_i\in S$ the subject. This formulation explicitly separates multiple dimensions, providing a clear and unified modeling basis for the evaluation tasks. Further details on data collection, preprocessing, and the generation and verification of persona and behavior labels are provided in App.~\ref{app:dataset}.

\subsection{Data Collection and Preprocessing}
\label{subsec:collection}

\begin{wraptable}{r}{0.42\textwidth}
  \centering
  \vspace{-30pt}
  \caption{Statistics for $\mathcal{D}_{base}$.}
  \label{tab:dialogue_stats}
  \vspace{-5pt}
  \setlength{\tabcolsep}{2pt}
  \small
  \begin{tabular}{lccc}
    \toprule
    \textbf{Subject} & \textbf{Rounds} & \textbf{Turns} & \textbf{Turns / Rounds} \\
    \midrule
    \textbf{Chinese} & 401 & 1,531 & 3.82 \\
    \textbf{Math}    & 395 & 3,420 & 8.66 \\
    \textbf{English} & 512 & 7,863 & 15.36 \\
    \midrule
    \textbf{Total}   & 1,308 & 12,814 & 9.80 \\
    \bottomrule
  \end{tabular}
  \vspace{-5pt}
\end{wraptable}

We denote the raw classroom corpus as $\mathcal{D}_{base}$, serving as the foundation for subsequent data construction. The subject set is $S={\text{Chinese},,\text{Mathematics},,\text{English}}$, ensuring broad curricular and linguistic coverage.
As shown in Tab.~\ref{tab:dialogue_stats}, $\mathcal{D}_{base}$ contains \textbf{1,308} dialogue rounds with \textbf{12,814} teacher–student Q\&A turns, averaging 9.8 turns per round, reflecting rich classroom interactions. The Chinese subset is drawn from the \textit{National Primary and Secondary Smart Education Platform}~\citep{SmartEdu2024}, comprising 401 rounds from 32 verified junior secondary open-class videos. The mathematics subset integrates \textit{TIMSS-Math}~\citep{stigler2000video} (spanning classrooms from seven countries, predominantly English-speaking) and the \textit{NCTE} corpus~\citep{demszky2023nctetranscriptsdatasetelementary}, totaling 395 rounds. The English subset comes from the \textit{TSCC v2} corpus~\citep{caines-etal-2022-teacher}, contributing 512 rounds of authentic classroom dialogues.
Overall, $\mathcal{D}_{base}$ offers \textbf{cross-subject, cross-lingual, and cross-cultural} coverage while strictly complying with copyright and privacy rules, providing a diverse, reliable, and pedagogically grounded resource that underpins persona modeling, behavior annotation, and evaluation tasks.

\subsection{Persona and Behavior Annotation}
\label{subsec:labeling}

Building on $\mathcal{D}_{base}$, we enrich the dataset with two additional layers of annotation: \textit{persona stylization} and \textit{behavior–expression labeling}, resulting in the complete dataset $\mathcal{D}$. This expansion substantially enhances stylistic and interactional diversity, while providing a unified foundation for the subsequent evaluation tasks.

\textbf{Persona Stylization.}  
We adopt the Big Five personality theory to define the persona set as:
$P=\mathcal{F}\times\{H,L\}$, 
$\mathcal{F}=\{\text{Extraversion},\,\text{Agreeableness},\,\text{Conscientiousness},\,\text{Neuroticism},\,\text{Openness}\},$
where $H$ and $L$ denote high and low levels, yielding $|P|=10$ standardized persona types (see App.~\ref{subsubapp:ten_personas}). For each sample $(x,y)$ consisting of classroom context and student response, we define a rewriting function:
$
  g:(x,y,p)\mapsto y^{(p)},
$
where $p\in P$ is the target persona and $y^{(p)}$ the persona-conditioned output. The function $g$ preserves semantic content while adapting expression style. Each dialogue is thus expanded into 10 persona-specific versions, while teacher feedback remains unchanged. This expansion increases the dataset size by nearly an order of magnitude, enriching stylistic diversity for persona-aware evaluation.  

\textbf{Behavior–Expression Labeling.}  
Classroom discourse is inherently multimodal, involving both verbal responses $y^{(p)}$ and non-verbal signals. We construct a four-dimensional label space (see App.~\ref{subsubapp:behavior_labels}):
$
  B=B_{beh}\times B_{emo}\times B_{exp}\times B_{voi},
$
where $B_{beh}$ (Behavior), $B_{emo}$ (Emotion), $B_{exp}$ (Expression), and $B_{voi}$ (Voice) jointly capture learning orientation and observable signals. Each instance is denoted as $b=(beh,emo,exp,voi)\in B$. 
The labeling function is defined as:
$
  f:(x,\,y^{(p)},\,p)\mapsto b,
$
where GPT-4o infers $b$ given $(x,y^{(p)},p)$, with low-confidence or inconsistent predictions flagged for human review. This ensures high-quality alignment between verbal and non-verbal signals.  

In summary, the enriched dataset is formalized as:
$
  \mathcal{D}=\{(x,\,y^{(p)},\,p,\,b,\,s)\},
$
where $s\in S$ denotes the subject. By integrating the persona set $P$ and behavior label space $B$, $\mathcal{D}$ provides a systematic and extensible foundation for evaluating the three tasks of basic coherence, student realism, and persona consistency (see Fig.~\ref{fig:example}).

\subsection{Data Quality Verification and Statistics}
\label{subsec:data-quality}

\begin{figure*}[t!]
  \centering
  % \vspace{-20pt}
  \includegraphics[width=\linewidth]{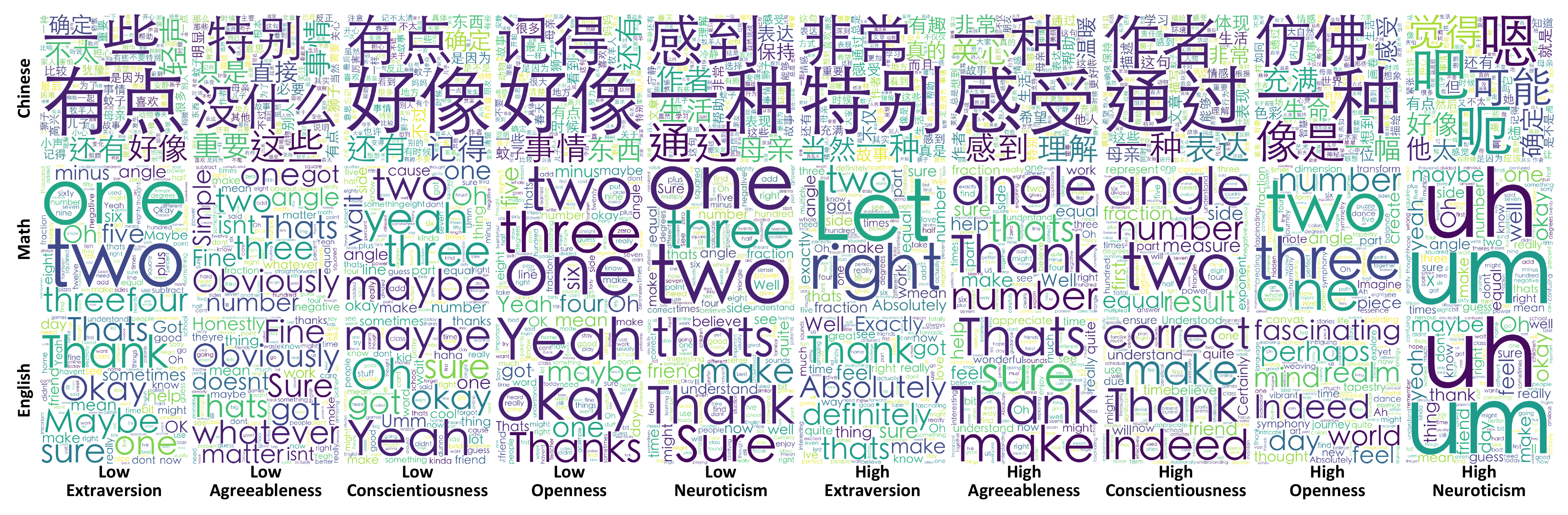}
  % \vspace{-20pt}
  \caption{\textbf{Cross-subject and persona linguistic variation.} The word clouds show high-frequency token distributions across Chinese, Math, and English under high/low persona settings, revealing distinct lexical preferences and expression patterns that provide linguistic features for evaluating student realism and persona consistency.
}
% \vspace{-10pt}
  \label{fig:wordcloud}
\end{figure*}

\textbf{Quality Control.}
Rigorous quality control was embedded throughout the dataset construction process to guarantee authenticity, reliability, and consistency. The preprocessing pipeline involved automated crawling of classroom videos and subtitles, text cleaning to remove narration and noise, and normalization to preserve semantic completeness. We reconstructed the IRF (Initiation–Response–Feedback) structure and assigned teacher–student roles, supported by a human-in-the-loop verification loop (automatic annotation, manual sampling, and iterative correction), which achieved 100\% verified role labels. For the Chinese subset, each dialogue instance was cross-checked against the corresponding public classroom video, while the mathematics and English subsets underwent multiple rounds of large-scale sampling and manual screening to filter out low-quality or inconsistent cases. All utterances were de-identified to eliminate personal information, and behavior–expression annotation achieved full coverage, with all labels strictly within the predefined vocabulary. These measures ensure the authenticity and reusability of $\mathcal{D}$, providing a reliable foundation for subsequent persona stylization and behavior modeling.

\begin{figure*}[t!]
% \vspace{-10pt}
\centering
\includegraphics[width=\textwidth]{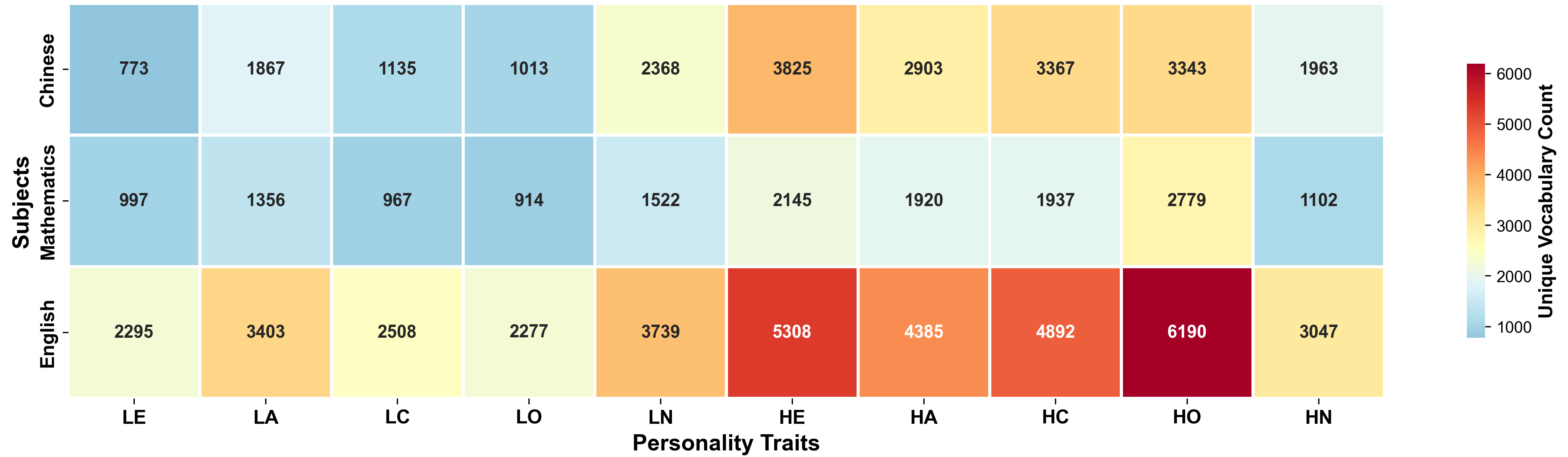}
% \vspace{-20pt}
\caption{\textbf{Vocabulary richness across subjects and personas.} The heatmap shows virtual students’ vocabulary coverage across three subjects and persona settings, indicating that both factors significantly shape lexical diversity.}
\label{fig:persona-heatmap}
% \vspace{-40pt}
\end{figure*}

\textbf{Statistical Analysis.}
Beyond quality assurance, we performed linguistic analyses to examine the effects of persona stylization. As shown in Fig.~\ref{fig:wordcloud}, word clouds illustrate cross-subject and cross-persona token distributions, revealing distinct lexical preferences and expression patterns under different Big Five settings. These differences highlight the interpretability of persona-conditioned outputs and provide observable linguistic cues. Moreover, Fig.~\ref{fig:persona-heatmap} presents a heatmap of vocabulary richness, showing that both subject domain and persona traits significantly affect lexical coverage. For example, English classes exhibit broader lexical diversity overall, while high extraversion and high openness personas consistently yield richer vocabulary across all subjects. These results demonstrate that the dataset not only achieves large scale and high consistency but also encodes quantifiable cross-subject and cross-persona linguistic variation, offering critical support for evaluation.

The final dataset thus achieves balanced coverage across subjects, languages, persona types, and behavior--expression dimensions. Its large scale and strong consistency make it well-suited for systematic evaluation of virtual student agents. Full distributions and examples are provided in the appendix for reproducibility and further extension.

%%%%%%%%%%%%%%%%%%%%%%%%%%%%%%%%%

\vspace{-5pt}
\section{Evaluation Framework}
\label{sec:evaluation}
\vspace{-5pt}

Evaluating virtual student agents poses the fundamental challenge of transforming inherently subjective abilities into measurable and reproducible statistics. To address this, we formalize the agent as a conditional generative model:
$
  M:(x_t,\,C,\,p,\,s)\mapsto Y_t,
$
where $x_t$ denotes the teacher’s input at turn $t$, $C$ the dialogue context (including prior turns), $p\in P$ the persona configuration of the student agent, $s\in S$ the subject domain, and $Y_t$ the generated response with associated labels. Equivalently, the model induces a conditional distribution $\mathbb{P}_M(Y_t \mid x_t, C, p, s)$, which is compared against a reference distribution $\mathbb{P}^\star(Y_t \mid x_t, C, p, s)$ derived from the dataset $\mathcal{D}$. The evaluation problem thus reduces to measuring the statistical divergence between $\mathbb{P}_M$ and $\mathbb{P}^\star$ under task-specific criteria.  
For experimental consistency, we partition the dataset $\mathcal{D}$ into a fine-tuning set $\mathcal{D}_{ft}$ and a held-out test set $\mathcal{D}_{test}$, following a fixed 6:4 split. Fine-tuning is performed exclusively on $\mathcal{D}_{ft}$, while all evaluations are conducted on $\mathcal{D}_{test}$. It ensures that reported results reflect genuine generalization performance rather than memorization of training data.  

To operationalize subjective ability assessment, we decompose the problem into three progressive tasks:  
(Task 1) \textbf{Basic coherence}: evaluating whether multimodal behavior–expression labels align with contextual semantics;  
(Task 2) \textbf{Student realism}: assessing whether responses resemble authentic student traits and classroom norms (e.g., admitting ignorance, requesting hints, or self-correcting);  
(Task 3) \textbf{Persona consistency}: examining whether persona traits and stylistic patterns are preserved across extended dialogues.  
Together, these layers form a progressive chain from observable behaviors, to perceptual authenticity, and ultimately to long-term stability. Each evaluation is stratified by subject $s$ and persona $p$, with confidence intervals and statistical tests ensuring robustness. In this way, EduPersona transforms vague notions such as “realism” or “consistency” into quantifiable distributional properties, enabling systematic and reproducible benchmarking of virtual student agents. See App.~\ref{app:evaluation} for further details.

\subsection{Task 1: Basic Coherence}
\label{subsec:task1}

At the most observable level, a virtual student agent should not only generate textual content but also produce behavior–expression patterns that are consistent with contextual semantics and the persona configuration. Given $(x_t, C, p, s)$, the model outputs a response $\hat y_t$ and a predicted behavior vector:
$
\hat b_t = (\hat{beh}_t,\, \hat{emo}_t,\, \hat{exp}_t,\, \hat{voi}_t) 
\in B = B_{beh} \times B_{emo} \times B_{exp} \times B_{voi},
$
where $B_{beh}$ (ICAP-based classroom behaviors), $B_{emo}$ (emotions), $B_{exp}$ (facial expressions), and $B_{voi}$ (vocal styles) form a \emph{closed vocabulary}. Reference labels $b_t=(beh_t,\,emo_t,\,exp_t,\,voi_t)$ are generated by GPT-4 under persona and context constraints and then human-audited (see Sec.~\ref{subsec:labeling}). We macro-average all metrics over the four dimensions, using the index set $\mathcal{I}=\{\text{beh},\text{emo},\text{exp},\text{voi}\}$ and $T$ evaluation instances; $\emptyset$ denotes no output, and $B_i$ the codebook of dimension $i$.

\textbf{Step 1. Response rate.} We first check whether any output is produced on each dimension:
\begin{equation}
\mathrm{RespRate}=\frac{1}{|\mathcal{I}|}\sum_{i\in\mathcal{I}}\frac{1}{T}\sum_{t=1}^{T}\mathds{1}[\hat b_{t,i}\neq\emptyset].
\end{equation}

\textbf{Step 2. Validity rate.} Conditional on producing outputs, we verify whether they fall within the predefined vocabulary:
\begin{equation}
\mathrm{ValidRate}=\frac{1}{|\mathcal{I}|}\sum_{i\in\mathcal{I}}
\frac{\sum_{t=1}^{T}\mathds{1}[\hat b_{t,i}\in B_i]}
{\max\!\left(1,\sum_{t=1}^{T}\mathds{1}[\hat b_{t,i}\neq\emptyset]\right)}.
\end{equation}

\textbf{Step 3. Label prediction quality.} Restricting to dimensions with outputs, we evaluate agreement with the reference annotations. To disentangle different error sources, we define three complementary accuracies (same numerator—the number of correct labels—but different denominators):
\begin{equation}
\mathrm{RawAcc}=\frac{1}{|\mathcal{I}|}\sum_{i\in\mathcal{I}}
\frac{\sum_{t=1}^{T}\mathds{1}[\hat b_{t,i}=b_{t,i}]}
{\max\!\left(1,\sum_{t=1}^{T}\mathds{1}[\hat b_{t,i}\neq\emptyset]\right)},
\end{equation}
\begin{equation}
\mathrm{ValAcc}=\frac{1}{|\mathcal{I}|}\sum_{i\in\mathcal{I}}
\frac{\sum_{t=1}^{T}\mathds{1}[\hat b_{t,i}=b_{t,i}]}
{\max\!\left(1,\sum_{t=1}^{T}\mathds{1}[\hat b_{t,i}\in B_i]\right)},
\end{equation}
\begin{equation}
\mathrm{OverallAcc}=\frac{1}{|\mathcal{I}|}\sum_{i\in\mathcal{I}}\frac{1}{T}\sum_{t=1}^{T}\mathds{1}[\hat b_{t,i}=b_{t,i}].
\end{equation}

These three metrics satisfy the inequality 
$\mathrm{OverallAcc} \le \mathrm{RawAcc} \le \mathrm{ValAcc}$, 
reflecting a progressive tightening from \emph{availability} $\rightarrow$ \emph{validity} $\rightarrow$ \emph{end-to-end correctness}. 
Task~1 thus provides the most objective baseline for behavioral alignment and supplies interpretable low-level signals upon which higher-level evaluations of student realism (Task 2) and persona consistency (Task 3) can be built.

\subsection{Task 2: Student Realism}
\label{subsec:task2}

Student realism evaluates whether a model’s response looks like a real student. This concept goes beyond linguistic fluency, requiring identity credibility (e.g., admitting ignorance, requesting hints, self-correction) and adherence to classroom interaction norms.
To ground the evaluation, we consulted ten experts from education and AI, who reviewed a sampled subset of responses and distilled a set of core dimensions $\mathcal{R}=\{r_1,\dots,r_m\}$, including linguistic naturalness, identity credibility, strategy appropriateness, and coordination with teacher feedback. These dimensions form the foundation for subsequent large-scale evaluation.
We then encode $\mathcal{R}$ into prompts to construct an evaluation function:
$
G_{\mathcal{R}}:(x_t,\,C,\,p,\,s,\,\hat y_t)\ \longmapsto\ \{\hat h,\,\hat{\mathbf{z}}\},
$
where $\hat h\in\{0,1\}$ denotes the overall student-likeness decision and $\hat{\mathbf{z}}\in\{0,1\}^{m}$ the dimension-wise outcomes. In this setup, GPT does \emph{not} serve as an independent judge but as a scalable extension of the expert-derived criteria, ensuring interpretability and reproducibility.
Results are aggregated across subjects $s$ and personas $p$, with macro-averaged scores reported at both overall and dimension levels, yielding a systematic, interpretable, and scalable evaluation of student realism.

\subsection{Task 3: Persona Consistency}
\label{subsec:task3}

Persona consistency requires virtual student agents to remain aligned with the target persona in both single-turn responses and extended dialogues. We define a standardized confidence function with range $[0,1]$:
$
J(\hat y_t,\,p)\in[0,1],
$
where $J(\hat y_t,\,p)=0$ indicates complete mismatch, $J(\hat y_t,\,p)=1$ indicates perfect alignment, and intermediate values reflect partial consistency.
Evaluation is conducted at two scales. For \textbf{short-term consistency}, each generated response $\hat y_t$ on the held-out test set $\mathcal{D}_{test}$ receives a persona score, and the average across samples is reported. For \textbf{long-term consistency}, models engage in fixed 10-turn classroom-style interactions driven by a \emph{Teacher-Policy} $\pi_T$ induced from the full dataset $\mathcal{D}$, which captures authentic instructional patterns such as IRF structures, scaffolding, and progressive difficulty. Scores are aggregated over the session to assess stability under sustained interaction.
A unified metric is applied to both settings:
$
\mathrm{Cons}=\frac{1}{|\mathcal{T}|}\sum_{t\in\mathcal{T}} J(\hat y_t,\,p),
$
where $\mathcal{T}=\{1,\dots,N\}$ for short-term consistency (with $N$ test samples) and $\mathcal{T}=\{1,\dots,10\}$ for long-term interactive sessions (fixed at 10 turns). This formulation allows us to examine both immediate persona alignment and its persistence throughout extended interactions.

%%%%%%%%%%%%%%%%%%%%%%%%%%%%%%%%%

\section{Experimental Design and Analysis}
\label{sec:experiments}

Building on EduPersona, we delineate the performance boundaries of virtual student agents across \emph{basic coherence}, \emph{student realism}, and \emph{persona consistency}. We study two complementary settings in a unified protocol: baseline evaluation, where three representative foundation LLMs $\mathcal{M}^{(i)}_{\text{base}}$ are directly assessed without additional adaptation, and persona-conditioned evaluation, where each base model $\mathcal{M}^{(i)}$ is fine-tuned over the Big Five–based persona set $P=\{p_1,\dots,p_{10}\}$ to obtain $\mathcal{M}^{(i)}_{\text{ft}}(p)$, yielding $3\times 10=30$ EduPersona-trained variants. All evaluations are run on the held-out test set $\mathcal{D}_{test}$, while fine-tuning uses only $\mathcal{D}_{ft}$; the complete dataset $\mathcal{D}$ is split $6{:}4$ (train:test) to emphasize generalization.
Within this setup, Sec.~\ref{subsec:architecture} outlines the model lineup and fine-tuning configuration; Secs.~\ref{subsec:exp1}–\ref{subsec:exp3} report task-wise results. Further implementation details are provided in App.~\ref{app:experiments}.

\subsection{Experimental Architecture and Model Selection}
\label{subsec:architecture}

We employ three open-source foundation LLMs as bases: Qwen3-8B ($\mathcal{M}^Q$), noted for strong Chinese–English instruction following \citep{yang2025qwen3}; InternLM3-8B-Instruct ($\mathcal{M}^I$), robust in Chinese educational scenarios with broad multilingual coverage \citep{cai2024internlm2}; and DeepSeek-R1-Distill-Qwen-14B ($\mathcal{M}^D$), distilled for enhanced mathematical and logical reasoning \citep{guo2025deepseek}. For each $\mathcal{M}^{(j)}$ and persona $p\in P=\{p_1,\dots,p_{10}\}$, we obtain a persona-specific variant $\mathcal{M}^{(j)}_{\text{ft}}(p)$, while the unadapted counterparts $\mathcal{M}^{(j)}_{\text{base}}$ serve as references.
Fine-tuning follows a consistent LoRA configuration (rank $r=16$, scaling $\alpha=32$) with AdamW, learning rate $\eta=3\times 10^{-4}$, per-device batch size $8$ and gradient accumulation $4$, for up to $5$ epochs. Each turn is encoded as a unified input $u_t=(x_t,\,C,\,p,\,s,\,b_t)$ to couple generative dialogue with closed-vocabulary behavior–expression codes. Inference settings and random seeds are aligned across models.

\subsection{Basic Coherence: Can virtual students generate multimodal behaviors aligned with context?}
\label{subsec:exp1}

Basic coherence requires virtual students not only to produce text but also to align behaviors, emotions, expressions, and vocal styles with the classroom context. We evaluate three model families using the five metrics defined in Sec.~\ref{subsec:task1}—response rate, validity rate, RawAcc, ValAcc, and OverallAcc—macro-averaged across four dimensions (behavior, emotion, expression, voice).

\begin{figure}[ht!]
  \centering
  % \vspace{-5pt}
  \includegraphics[width=\linewidth]{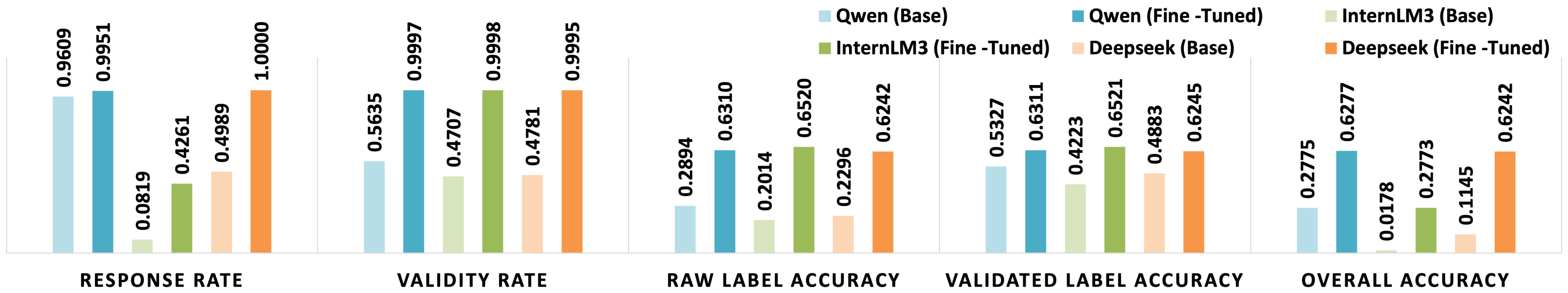}
  % \vspace{-5pt}
  \caption{\textbf{Impact of persona fine-tuning on basic coherence across five metrics.} Persona fine-tuning consistently improves basic coherence across all models, demonstrating the value of EduPersona. Fine-tuned Qwen and DeepSeek achieve OverallAcc above 0.62 with strong label alignment, while InternLM3 also benefits but remains constrained by a low response rate.}
  % \vspace{-5pt}
  \label{fig:task1-results}
\end{figure}

Fig.~\ref{fig:task1-results} shows that \textbf{persona fine-tuning substantially enhances multimodal alignment}. Qwen (0.9951/0.9997) and DeepSeek (1.0000/0.9995) achieve near-perfect response and validity rates, while InternLM3 retains high validity (0.9998) but suffers from low response coverage (0.4261). For label alignment, fine-tuned RawAcc and ValAcc converge in the 0.624–0.653 range, a large improvement over base versions (e.g., Qwen: 0.2894 $\rightarrow$ 0.6310). With validity nearly saturated, Raw and Val scores converge, whereas base models show wider Raw–Val gaps due to frequent OOV outputs (e.g., DeepSeek: 0.2296 $\rightarrow$ 0.4883). For end-to-end correctness, Qwen and DeepSeek reach OverallAcc of 0.6277 and 0.6242, far above their baselines (0.2775/0.1145), while InternLM3 remains at 0.2773, constrained by limited response coverage.

In summary, Task~1 shows that \textbf{persona fine-tuning substantially enhances basic coherence, enabling Qwen and DeepSeek to approach usable levels of multimodal alignment, though InternLM3 still faces a bottleneck in response generation}. 
Beyond overall gains, two further insights emerge based on detailed per-dimension results in App.~\ref{subapp:task1_figs}. 
First, a stable difficulty ordering is observed across dimensions, with Emotion being the easiest and Behavior the hardest, highlighting that residual errors stem from discourse- and intent-level challenges rather than random noise. 
Second, Qwen and DeepSeek converge in end-to-end accuracy after fine-tuning (OverallAcc $\approx 0.62$), suggesting that task structure and dataset design, rather than model scale alone, largely determine the performance ceiling. 
With validity already near saturation, future improvements will most likely come from increasing response coverage and strengthening fine-grained behavioral guidance.

\subsection{Student Realism: Can virtual students be like real students?}
\label{subsec:task2}

\begin{figure}[ht!]
% \vspace{-20pt}
\centering
\includegraphics[width=\textwidth]{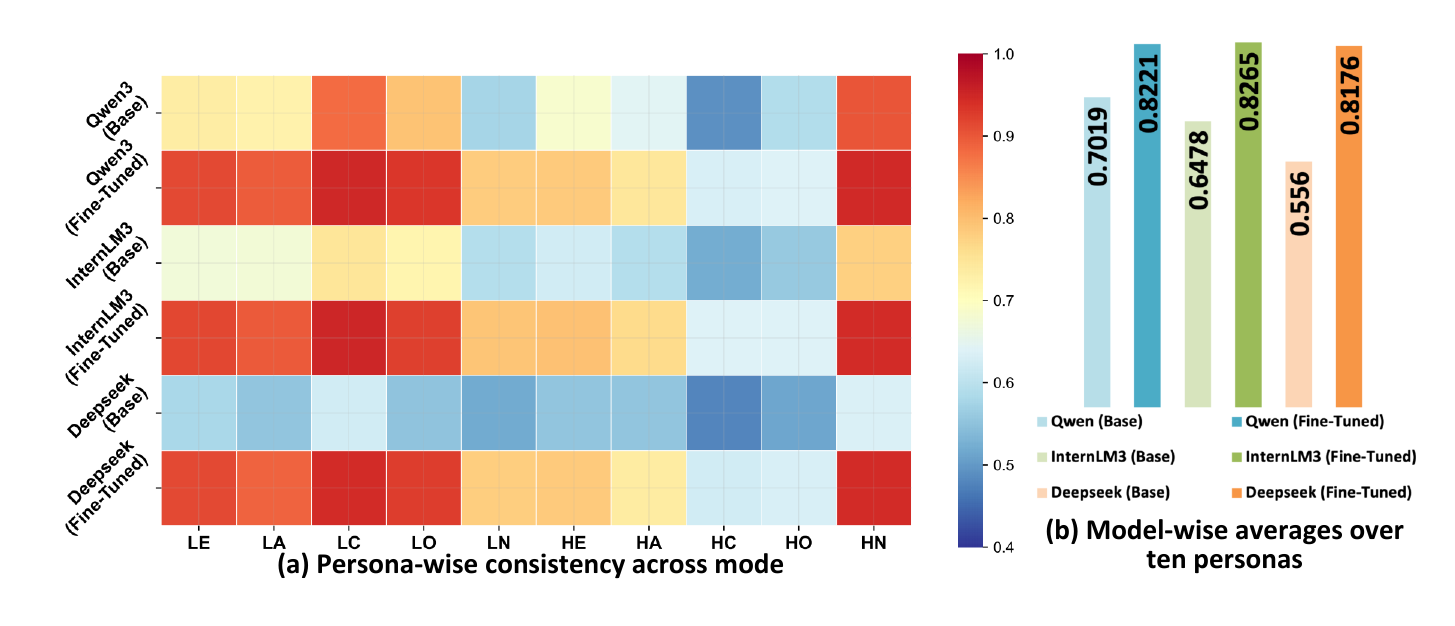}
% \vspace{-20pt}
\caption{\textbf{Persona fine-tuning markedly improves student realism.} EduPersona brings all three model families to a converged level around 0.82 while revealing stable persona-specific differences (HC/HO harder, HN/LC/LO easier).}
\label{fig:task2-combo}
% \vspace{-10pt}
\end{figure}

Student realism is a key criterion for evaluating whether virtual students resemble real learners. Using expert-derived dimensions operationalized through an automatic evaluator, we analyze results from both persona and model perspectives (Fig.~\ref{fig:task2-combo}).

\textbf{Persona-wise analysis (Fig.~\ref{fig:task2-combo}a).}
Fine-tuned models exhibit consistently stronger realism across all personas, yet heterogeneity persists. High Conscientiousness (HC) and High Openness (HO) remain lower both before and after fine-tuning, with modest gains, whereas High Neuroticism (HN), Low Conscientiousness (LC), and Low Openness (LO) achieve relatively high realism. This reflects the interaction between persona traits and model defaults: HC/HO emphasize objective, comprehensive, and teacher-aligned behaviors, overlapping with LLMs’ inherent answer-first tendency, thus appearing more machine-like. By contrast, HN/LC/LO manifest hesitation, partial responses, or self-corrections, which enhance perceived authenticity and yield higher realism.

\textbf{Model-wise analysis (Fig.~\ref{fig:task2-combo}b).}
At the baseline level, the three families differ substantially (Qwen 0.7019, InternLM3 0.6478, DeepSeek 0.556). After fine-tuning, however, all models converge to a narrow band around 0.82 (Qwen 0.8221, InternLM3 0.8265, DeepSeek 0.8176). This demonstrates that persona conditioning both elevates student realism and reduces inter-model disparities.

In summary, Task~2 shows that persona fine-tuning significantly enhances student realism and brings different models to a converged level, though persona-specific bottlenecks remain. Additional analysis in the App.~\ref{subapp:task2_figs} further reveals that the absolute gains vary across models, with DeepSeek benefiting most, and that cross-task consistency emerges: the easiest Emotion dimension in Task~1 corresponds to high realism personas (HN/LN/LO), while the hardest Behavior dimension aligns with low realism personas (HC/HO). These findings establish EduPersona’s effectiveness in enhancing high-level perception while also highlighting areas where persona-specific improvements are needed, providing a solid foundation for subsequent evaluations of persona consistency.

\subsection{Persona Consistency: Can virtual students maintain stable personas during interactions?}
\label{subsec:exp3}

Task~3 evaluates whether virtual students can maintain stable persona traits throughout interactions. Results show that fine-tuned models achieve substantially higher and more stable consistency at both persona and model levels, confirming the effectiveness of persona conditioning.

\begin{figure}[ht!]
% \vspace{-10pt}
\centering
\includegraphics[width=\textwidth]{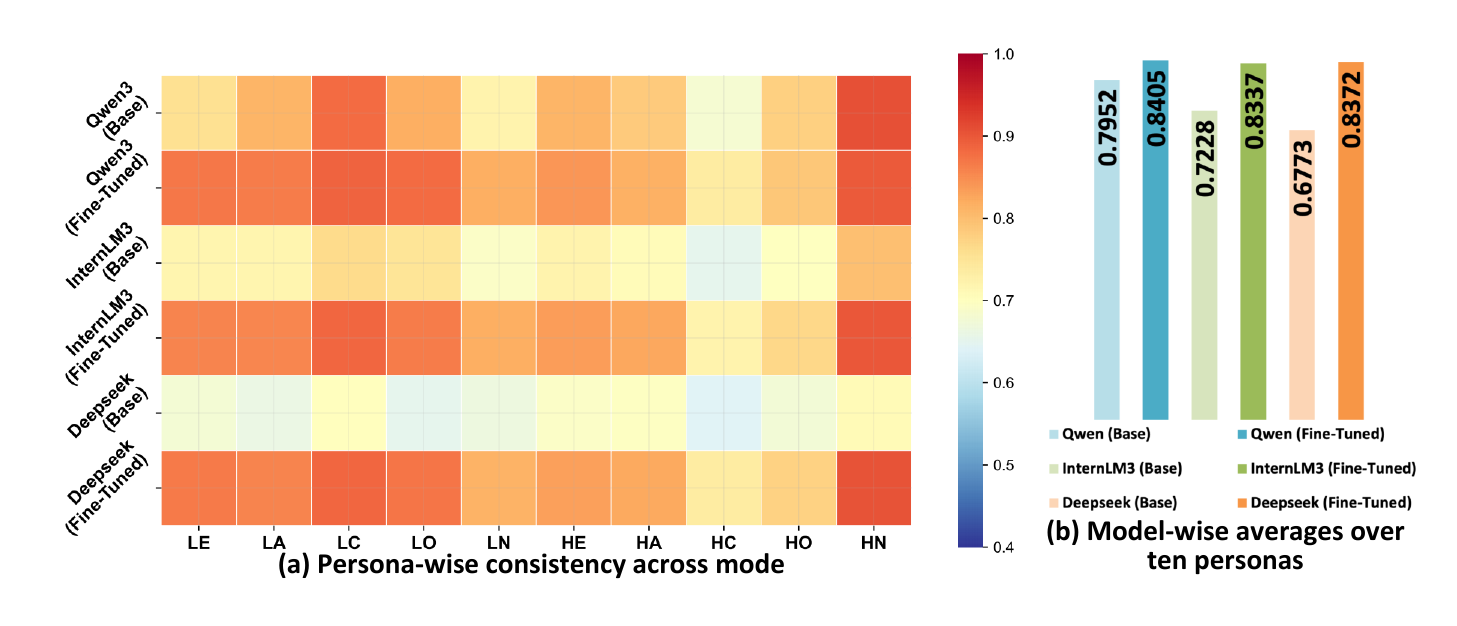}
% \vspace{-20pt}
\caption{\textbf{Persona fine-tuning enhances persona stability.} Fine-tuned models converge around 0.84 while showing consistent persona-specific differences (HC/HO harder, HN/LC/LO easier).}
\label{fig:task3-combo}
% \vspace{-10pt}
\end{figure}

\textbf{Persona-wise analysis (Fig.~\ref{fig:task3-combo}a).}
While fine-tuning consistently improves all ten personas, heterogeneity persists. High Conscientiousness (HC, 0.731) and High Openness (HO, 0.779) remain the most difficult to sustain, even after adaptation, whereas High Neuroticism (HN, 0.901), Low Conscientiousness (LC, 0.887), and Low Openness (LO, 0.873) achieve the highest stability. This echoes Task~2’s findings on realism, suggesting that structured, “idealized” personas are both less authentic and less consistent, while personas reflecting hesitation or partial responses are easier to maintain. Gains also vary across personas: Low Extraversion (+0.146), Low Openness (+0.133), and Low Agreeableness (+0.130) benefit most, whereas High Openness (+0.062) and High Conscientiousness (+0.073) improve the least.

\textbf{Model-wise analysis (Fig.~\ref{fig:task3-combo}b).}
At the baseline level, Qwen, InternLM3, and DeepSeek differ significantly (0.795, 0.723, and 0.677). After fine-tuning, however, all three converge to a narrow range of 0.833–0.841, showing that persona conditioning boosts consistency while reducing cross-family disparities.

\textbf{Multi-turn consistency.} To compare the persona retention ability of fine-tuned versus closed-source models, we conducted a 10-turn English classroom experiment, involving three LoRA-fine-tuned models (Qwen3, InternLM3, DeepSeek) and GPT-4o.  LoRA-fine-tuned models retained persona traits even after explicit prompts disappeared, with Qwen3-LoRA achieving the highest overall score ($0.920 \pm 0.042$). All three models maintained $\geq 0.8$ in later turns (6--10), whereas GPT-4o quickly drifted once persona instructions vanished, averaging only $0.480 \pm 0.262$ and dropping below 0.50 in later turns. This highlights the importance of fine-tuning for long-horizon persona stability.

In summary, Task~3 demonstrates that \textbf{persona consistency is more demanding than student realism, but persona fine-tuning significantly enhances stability while narrowing both model- and persona-level gaps}. Notably, HC and HO personas remain persistent bottlenecks across realism and consistency, whereas HN, LC, and LO are consistently easier to model. More detailed distributions and analyses are provided in App.~\ref{subapp:task3_figs}.

\subsection{Summary}
\label{subsec:exp-sum}

The three experiments depict the landscape of virtual students’ subjective abilities. In \textbf{basic coherence (Task~1)}, Qwen and DeepSeek converge around 0.62 after fine-tuning (0.6277/0.6242), while InternLM3 remains lower at 0.2773 due to limited response coverage (RespRate 0.4261; App.~\ref{subapp:task1_figs}). For \textbf{student realism (Task~2)}, baseline gaps (Qwen 0.7019, InternLM3 0.6478, DeepSeek 0.5560) narrow after fine-tuning, with all three converging near 0.82 (App.~\ref{subapp:task2_figs}). In \textbf{persona consistency (Task~3)}, models rise from 0.795/0.723/0.677 to ~0.84 (0.841/0.834/0.837), again showing significant gains and convergence (App.~\ref{subapp:task3_figs}). Thus, EduPersona boosts absolute performance while aligning cross-family outcomes.

The tasks also reveal a layered progression: Task~1 captures structural compliance, Task~2 addresses human-perceived realism, and Task~3 tests long-horizon persona stability. Persona-level trends are consistent—\textbf{HC/HO remain bottlenecks}, while \textbf{HN/LC/LO are easier to emulate}, echoing Task~1’s “Emotion easier, Behavior harder” pattern. A 10-turn study further shows LoRA-fine-tuned models retain traits after prompts vanish (e.g., Qwen3-LoRA 0.920±0.042), while GPT-4o drifts (0.480±0.262), underscoring that \textbf{long-horizon stability depends on fine-tuning rather than model scale}. In sum, performance ceilings are shaped more by persona type and interaction length than by parameter size, pointing to future work on coverage, behavior supervision, and embedding natural hesitation–repair–collaboration strategies into idealized personas.

\section{Conclusion}
\label{sec:conclusion}

This paper introduced \textbf{EduPersona}, a framework for evaluating virtual students across three subjective abilities: basic coherence, student realism, and persona consistency. Experiments show that persona fine-tuning markedly improves performance, with models converging at 0.62, 0.82, and 0.84. Stable persona-specific gaps persist: high conscientiousness and openness are harder to emulate, while high neuroticism, low conscientiousness, and low openness are easier to sustain. These results validate the framework and reveal structural links between persona traits and model behaviors. Overall, EduPersona demonstrates a progression from structural compliance to perceived realism and long-horizon stability, underscoring the importance of the “behavior” dimension and sustained dialogue. While fine-tuning narrows disparities, challenges remain in response coverage, idealized personas, and long-term stability. Future work should strengthen behavioral supervision, foster natural interaction, and extend to interdisciplinary contexts, positioning EduPersona as a pathway toward trustworthy educational agents.

% \clearpage

\section*{Ethics Statement}
This work complies with ethical standards and data usage regulations. The EduPersona dataset is constructed from publicly available classroom resources, with all data anonymized to remove IDs and other identifiers. Persona configurations are derived from the well-established Big Five personality theory and serve only as abstract stylistic constructs for modeling language and behavior; they do not encode demographic or sensitive attributes. Thus, this study introduces no new ethical risks and instead provides a safe, controlled, and reproducible framework for evaluating virtual students in educational contexts.

\section*{Reproducibility Statement}
To ensure reproducibility, we will open-source the complete codebase, annotation guidelines, preprocessing pipeline, and trained models. The released EduPersona resources will include the full data schema (prompts, vocabularies, labels, and evaluation metrics). Third-party raw media will not be redistributed due to licensing restrictions; instead, we provide scripts to re-derive processed text from public resources. Appendix~\ref{app:dataset} details data collection and preprocessing, Appendix~\ref{subsubapp:ten_personas} and \ref{subsubapp:persona_prompt} describe persona configurations and stylization prompts, and Appendix~\ref{app:evaluation} provides the evaluation prompts. Additional fine-grained analyses are reported in Appendix~\ref{subapp:task1_figs}–\ref{subapp:task3_figs} to further substantiate our conclusions. All code, data, and models will be released upon acceptance to foster long-term reproducibility and extensibility in educational agent research.

\bibliography{iclr2026_conference}
\bibliographystyle{iclr2026_conference}

\clearpage

\appendix

\renewcommand{\thetable}{A\arabic{table}}
\renewcommand{\thefigure}{A\arabic{figure}}
\renewcommand{\thealgocf}{A.\arabic{algocf}}  % 注意是 thealgocf 不是 thealgorithm
\setcounter{figure}{0}  
\setcounter{table}{0}  
\setcounter{algocf}{0} 

\section*{Appendix}
This appendix provides implementation details of the EduPersona corpus and the processing steps that support the evaluation framework in the main paper, along with additional fine-grained experimental results that further substantiate our conclusions.

% \section{The Use of Large Language Models (LLMs)}
% During manuscript preparation, Large Language Models (LLMs) were employed solely for language refinement and stylistic polishing. 

\section{Dataset Details}
\label{app:dataset}

\subsection{Detailed Data Collection and Preprocessing}
\label{subapp:dataset}

The EduPersona corpus is constructed from publicly accessible classroom resources licensed for research, covering three subjects (Chinese, Mathematics, English) across two languages (Chinese/English). Sources include: (i) curated Chinese open-class videos and transcripts from a national education platform, (ii) international mathematics discourse corpora, and (iii) English classroom transcripts. Only segments with clear instructional intent are retained; all personally identifiable information is masked, and collection complies with the original platforms’ terms of use. No redistribution of third-party raw media is performed.

\textbf{Preprocessing and structuring.}
A unified pipeline ensures cross-subject comparability. Videos and subtitles (or ASR transcripts where subtitles are unavailable) are aligned at the utterance level with normalized punctuation and casing. Teacher–student roles and IRF structures are reconstructed, narration and meta-comments are removed, and turn boundaries are adjusted to avoid truncation. Identifiers such as IDs are masked at source to guarantee non-attributability. A human-in-the-loop process (automatic tagging, sampled review, manual correction) yields verified role labels.

\textbf{Persona expansion and behavior labeling.}
To enable persona-conditioned dialogue modeling, each student response is expanded into ten variants using the Big Five dimensions $\mathcal{F}={\text{Extraversion},\ \text{Agreeableness},\ \text{Conscientiousness},\ \text{Neuroticism},\ \text{Openness}}$ with high/low polarity, giving $P=\mathcal{F}\times{H,L}$. Each original $(x,y)$ is rewritten into $y^{(p)}$ that preserves semantics while adapting to persona $p$. In addition, each $(x,y^{(p)})$ is labeled with a four-dimensional vector $b=(beh, emo, exp, voi)$ covering \emph{Behavior}, \emph{Emotion}, \emph{Expression}, and \emph{Voice}, constrained to a fixed vocabulary (App.~\ref{subsubapp:behavior_labels}). Low-confidence or contradictory cases are resolved through human auditing.

\textbf{Split, statistics, and quality control.}
The full dataset $\mathcal{D}$ is stratified by subject and persona into fine-tuning ($\mathcal{D}{ft}$) and test sets ($\mathcal{D}{test}$) with a 6{:}4 ratio. All evaluation tasks (Tasks~1–3) are performed exclusively on $\mathcal{D}{test}$.
Before persona expansion, the base corpus $\mathcal{D}{base}$ contains \textbf{1,308} dialogue rounds with \textbf{12,814} teacher–student Q\&A turns. By subject: Chinese contributes 401 rounds from 32 lessons (1,531 Q\&A turns; avg.\ 3.82/round), Mathematics 395 rounds (3,420 Q\&A turns; avg.\ 8.66/round), and English 512 rounds (7,863 Q\&A turns; avg.\ 15.36/round). After stylization, each turn yields ten persona-conditioned variants, expanding the dataset tenfold.
Quality control is embedded at every stage: role integrity verified against source materials, persona fidelity checked by independent judges, label validity enforced with 100\% in-vocabulary coverage, and near-duplicates removed via semantic similarity. These measures ensure that EduPersona is reliable, diverse, and ethically compliant, providing a solid foundation for the evaluation framework and subsequent experiments.

\subsection{Data Access and Ethics}
\label{subapp:ethics}
We will release full preprocessing and labeling code, along with a legally redistributable subset that mirrors the complete schema (including prompts, vocabularies, and metric definitions), thereby supporting reproducibility of all reported experiments. Third-party raw media are not redistributed; instead, scripts are provided to re-derive processed text from publicly available resources where permitted.

All dialogues are fully de-identified, with names, IDs, and any sensitive information removed. Persona variants are abstract stylistic constructs derived from the Big Five framework and do not encode demographic attributes. We explicitly caution against any attempt at re-identification or demographic inference. The EduPersona corpus is designed under principles of compliance, transparency, and responsible use, ensuring safe deployment in educational and AI research contexts.

%  这里对应放一下数据来源，具体的处理流程之类的解释信息，以及数据来源合规之类的说明

\label{subapp:sources}

\subsection{More Information about Annotation}
\label{subapp:labeling}

% 这里需要给出所有标注准则的说明，比如我们的10个人格的定义，行为标签具体是啥之类的信息。此外，要给出这部分的prompt

\subsubsection{Ten Persona Configurations Based on Big Five Theory}
\label{subsubapp:ten_personas}

Our persona framework is grounded in the Big Five personality theory (BFAS scale), extending it into ten standardized configurations that represent authentic student behavioral patterns in classroom settings. Each persona is defined through multiple dimensions to ensure comprehensive and scientifically-based characterization.

\begin{longtable}{p{3.5cm}p{10cm}}
\caption{Ten persona configurations based on Big Five personality theory.}
\label{tab:detailed_persona_configs}\\
\toprule
\textbf{Persona Type} & \textbf{Detailed Characteristics} \\
\midrule
\endfirsthead

\multicolumn{2}{c}%
{{\bfseries Table \thetable\ continued from previous page}} \\
\toprule
\textbf{Persona Type} & \textbf{Detailed Characteristics} \\
\midrule
\endhead

\midrule \multicolumn{2}{r}{{Continued on next page}} \\ \midrule
\endfoot

\endlastfoot
\textbf{High Extraversion} & 
\textbf{Core traits}: Take charge, have a strong personality, warm up quickly to others \\
& \textbf{Behavioral traits}: Active participation, strong social skills, comfortable expression, high exhibition, obvious dominance \\
& \textbf{Language style}: Fluent and confident, detailed elaboration, positive attitude, clear structure, likes to lead conversations \\
& \textbf{Classroom behavior}: Strong desire to participate, actively answers questions, proactively shows themselves, frequent interaction with teachers \\
& \textbf{Response patterns}: Detailed answers with additional explanations, fluent language but may include repetitions and filler words \\
\midrule
\textbf{Low Extraversion} & 
\textbf{Core traits}: Do not have an assertive personality, hard to get to know, keep others at a distance \\
& \textbf{Behavioral traits}: Introverted and cautious, doesn't like to show off, socially conservative, low participation \\
& \textbf{Language style}: Concise and conservative, cautious responses, avoids excessive expression, passive responses \\
& \textbf{Classroom behavior}: Less proactive participation, speaks softly, doesn't want to be center of attention \\
& \textbf{Response patterns}: Brief answers, sometimes needs teacher encouragement to speak, lighter tone \\
\midrule
\textbf{High Agreeableness} & 
\textbf{Core traits}: Sympathize with others' feelings, take an interest in other people's lives, feel others' emotions \\
& \textbf{Behavioral traits}: Cooperative and caring, considerate and patient, positive feedback, understanding, high empathy \\
& \textbf{Language style}: Gentle and friendly, detailed elaboration, accurate expression, caring for others \\
& \textbf{Classroom behavior}: Happy to help classmates, actively participates in discussions, pays attention to others' feelings \\
& \textbf{Response patterns}: Gentle language, accurate expression, rarely makes mistakes, considers others' feelings \\
\midrule
\textbf{Low Agreeableness} & 
\textbf{Core traits}: Can't be bothered with others' needs, take advantage of others, not interested in other people's problems \\
& \textbf{Behavioral traits}: Competitive, direct and frank, self-focused, less compromising, lacks empathy \\
& \textbf{Language style}: Direct and clear, sometimes slightly harsh, focuses on personal views \\
& \textbf{Classroom behavior}: More focused on own performance, may lack patience with others' opinions \\
& \textbf{Response patterns}: Direct answers, sometimes slightly harsh, more focused on expressing own views \\
\midrule
\textbf{High Conscientiousness} & 
\textbf{Core traits}: Keep things tidy, like order, carry out plans \\
& \textbf{Behavioral traits}: Serious and responsible, organized, clear goals, persistent, highly self-disciplined \\
& \textbf{Language style}: Clear and logical, accurate and error-free, strong logic, complete expression \\
& \textbf{Classroom behavior}: Listens carefully, well-prepared when answering questions, solid grasp of content \\
& \textbf{Response patterns}: Accurate and complete answers, clear logic, rarely makes mistakes \\
\midrule
\textbf{Low Conscientiousness} & 
\textbf{Core traits}: Waste time, find it difficult to get down to work, leave belongings around \\
& \textbf{Behavioral traits}: Careless, poor organization, easily distracted, lacks persistence \\
& \textbf{Language style}: Simple and direct, occasional errors, incomplete expression, sometimes inconsistent \\
& \textbf{Classroom behavior}: Easily distracted, unprepared when answering questions \\
& \textbf{Response patterns}: Unstable answers, sometimes right sometimes wrong, incomplete expression \\
\midrule
\textbf{High Neuroticism} & 
\textbf{Core traits}: Get upset easily, get angry easily, get easily agitated \\
& \textbf{Behavioral traits}: Nervous and anxious, emotional fluctuations, sensitive and irritable, emotionally unstable \\
& \textbf{Language style}: Hesitant and indecisive, repetitive backtracking, incoherent expression, full of uncertainty \\
& \textbf{Classroom behavior}: Easily nervous, sensitive to classroom environment, shows worry when answering \\
& \textbf{Response patterns}: Full of 'um', 'uh' filler words, repeats and backtracks, incoherent expression \\
\midrule
\textbf{Low Neuroticism} & 
\textbf{Core traits}: Rarely get irritated, not easily annoyed, feel comfortable with self \\
& \textbf{Behavioral traits}: Emotionally stable, calm and composed, strong stress adaptability, high self-acceptance \\
& \textbf{Language style}: Stable and natural, clear logic, calm expression, few emotional fluctuations \\
& \textbf{Classroom behavior}: Remains calm when facing problems, peaceful attitude when answering \\
& \textbf{Response patterns}: Stable and natural answers, clear logic, shows inner calm and confidence \\
\midrule
\textbf{High Openness} & 
\textbf{Core traits}: Quick to understand things, believe in the importance of art, can handle a lot of information \\
& \textbf{Behavioral traits}: Strong curiosity, imaginative, accepts new things, flexible thinking, values aesthetics \\
& \textbf{Language style}: Creative and rich expression, good at association, broad thinking \\
& \textbf{Classroom behavior}: Curious about new knowledge, good at asking questions, creative answers \\
& \textbf{Response patterns}: Creative answers, makes associations and extensions, can handle complex information \\
\midrule
\textbf{Low Openness} & 
\textbf{Core traits}: Have difficulty understanding abstract ideas, do not like poetry, seldom notice emotional aspects of art \\
& \textbf{Behavioral traits}: Conservative and traditional, relies on experience, low acceptance, rigid thinking \\
& \textbf{Language style}: Simple and direct, lacks extension, difficulty going deep, prefers concrete descriptions \\
& \textbf{Classroom behavior}: Tends to rely on existing knowledge, low acceptance of new content \\
& \textbf{Response patterns}: Simple and direct answers, lacks extension, prefers concrete answers over abstract analysis \\
\bottomrule
\end{longtable}

\subsubsection{Prompt Template for Persona Stylization}
\label{subsubapp:persona_prompt}

To implement personality-driven dialogue generation, we design a structured prompt template that formalizes the subjective task of persona stylization into a systematic workflow. The system prompt defines the model as an “expert in Big Five personality theory” and establishes rules that govern three major components:

First, \textbf{student speech processing} is grounded in the five BFAS dimensions, with placeholders specifying core traits, behavioral tendencies, linguistic patterns, classroom manifestations, and response styles. This multi-dimensional mapping ensures that student responses preserve semantic meaning while being rewritten into styles consistent with the target persona.

Second, \textbf{teacher speech processing} follows a conservative strategy: original content is preserved whenever possible, with minimal modifications applied only in four predefined cases (e.g., coherence or disambiguation). This guarantees that instructional intent remains intact.

Finally, \textbf{quality and output requirements} enforce strict formatting rules and consistency checks, balancing stylistic fidelity with semantic preservation. By integrating these layers, EduPersona transforms persona-conditioned dialogue generation into a reproducible and scalable process, enabling consistent expansion and evaluation in large-scale educational applications.

\begin{tcolorbox}[title=\textbf{The Prompt for Persona Stylization.},breakable]

\textbf{System Prompt}

You are an expert deeply knowledgeable in student psychology and Big Five personality theory, particularly skilled at simulating different personality traits of students in classroom settings based on the Big Five personality framework.

Your task is to regenerate student speech that conforms to specific personality traits based on provided real teacher-student dialogues, while handling teacher speech according to strict rules.

\textbf{\#\#INSTRUCTIONS:}

- Student speech processing principles, reflecting target student personality traits - [TARGET PERSONALITY]:
  \begin{itemize}
  \item Core traits: [CORE TRAITS FROM BFAS SCALE]
  \item Behavioral traits: [BEHAVIORAL CHARACTERISTICS]  
  \item Language style: [LANGUAGE PATTERNS]
  \item Classroom behavior: [CLASSROOM MANIFESTATIONS]
  \item Response patterns: [TYPICAL RESPONSE STYLES]
  \item Audio clarity handling: When encountering unclear or inaudible portions, intelligently infer and complete missing content based on dialogue context, logical flow, and target personality traits
  \end{itemize}
- Teacher speech processing principles:
  \begin{enumerate}
  \item Keep teacher's original words unchanged by default
  \item Only make adjustments in the following situations:
    \begin{itemize}
    \item Teacher's referential content doesn't match student's new response
    \item Teacher's follow-up questions don't match student's new response structure
    \item Teacher's guidance obviously conflicts with student's new state
    \item Teacher's speech contains unclear or inaudible portions
    \end{itemize}
  \item When modifying, only adjust referential content and connection logic, maintain teacher's educational intent and professionalism
  \item Audio clarity handling: When encountering unclear or inaudible portions in teacher's speech, intelligently infer and complete the missing content based on educational context, teaching objectives, and professional pedagogical patterns
  \end{enumerate}

- Important requirements:
  \begin{enumerate}
  \item Strictly follow the scientific descriptions in the prompts to shape student personality traits, and student personality traits should remain consistent throughout the dialogue
  \item Student responses should match appropriate student knowledge level and vocabulary
  \item Maintain the educational significance and logical relationships of the dialogue
  \item For unclear or inaudible portions in the original dialogue (both teacher and student speech), use contextual inference and appropriate behavioral patterns to complete missing content naturally
  \end{enumerate}

- \textbf{DO NOT PROVIDE ANY OTHER OUTPUT TEXT OR EXPLANATION}. Output strictly in specified format without explanatory text.

\par\medskip
\hrule
\par\medskip

\textbf{User:}

Original teacher-student dialogue: [ORIGINAL DIALOGUE TEXT]

Please regenerate the dialogue based on the above information, where:

\textbf{Student speech requirements:}
\begin{itemize}
\item Completely regenerate according to [TARGET PERSONALITY] personality traits
\item Reflect typical performance and response patterns of this personality in classroom settings  
\item Maintain appropriate student knowledge level and expression style
\item For any unclear portions, infer and complete based on context and personality traits
\end{itemize}

\textbf{Teacher speech requirements:}
\begin{itemize}
\item Prioritize keeping original words unchanged
\item Only make adjustments when references don't match, follow-ups don't align, emotions conflict, or speech is unclear
\item For unclear portions, infer and complete based on educational context and teaching objectives
\item Maintain teaching objectives and professional expression unchanged
\end{itemize}

Please strictly follow the format below for output, maintain the same number of dialogue turns as the original dialogue, without any other explanatory text:

Teacher: [Teacher's words]\\
Student: [Student's words]\\  
Teacher: [Teacher's words]\\
Student: [Student's words]\\
...

\end{tcolorbox}

\subsubsection{Behavior–Expression Label Space}
\label{subsubapp:behavior_labels}

The behavior–expression annotation system employs a controlled vocabulary across four dimensions.
\medskip

\begin{longtable}{p{2cm} L{3.6cm} L{7.8cm}}
\caption{Behavior–expression controlled vocabulary with operational definitions.}
\label{tab:behavior_expression_labels}\\
\toprule

% \midrule
\textbf{Dimension} & \textbf{Label} & \textbf{Operational Definition} \\
\midrule
\endfirsthead

\multicolumn{3}{c}{{\bfseries Table \thetable\ (continued)}}\\
\toprule
\textbf{Dimension} & \textbf{Label} & \textbf{Operational Definition} \\
\midrule
\endhead

\midrule \multicolumn{3}{r}{{Continued on next page}}\\
\endfoot

\bottomrule
\endlastfoot

% ===== Behavior =====
\multirow{8}{*}{\textbf{Behavior}}
& Simple Response        & Answers with “yes/no”, “I don’t know”. \\
& Mechanical Repetition  & Repeats the teacher’s question or content verbatim. \\
& Standing Answer        & Independently provides a complete answer to the teacher’s question. \\
& Example Explanation    & Actively uses examples to explain knowledge. \\
& Summary Generalization & Summarizes the learned content; expresses personal understanding. \\
& Active Questioning     & Asks questions to express confusion or reflective thoughts. \\
& Supplementary Speech   & Expands or supplements others’ viewpoints. \\
& Opinion Expression     & Refutes or negotiates with others’ statements. \\
\midrule

% ===== Emotion =====
\multirow{3}{*}{\textbf{Emotion}}
& Positive & Positive/engaged/confident state. \\
& Confused & Unclear thinking or partial understanding but still trying. \\
& Negative & Disappointed, frustrated, helpless emotions. \\
\midrule

% ===== Expression =====
\multirow{5}{*}{\textbf{Expression}}
& Smile       & Pleasant, satisfied facial expression. \\
& Frown       & Confused or dissatisfied facial expression. \\
& Nod         & Agreeing/understanding/affirming action. \\
& Shake Head  & Disagreeing or confused action. \\
& Silence     & No obvious facial change; remains calm. \\
\midrule

% ===== Voice =====
\multirow{5}{*}{\textbf{Voice}}
% & Fast         & Fast speech rate (rushing/eager/nervous). \\
% & Slow         & Slow speech rate (thinking/unfamiliar/low mood). \\
& Hesitant     & “um/uh/I think/maybe”, disfluent. \\
& Confident    & Clear, continuous, steady tone. \\
& Enthusiastic & Higher pitch, emotional emphasis. \\
& Low-tone     & Low and weak voice. \\
& Neutral      & Medium rate and tone; no strong affect. \\

\end{longtable}

\clearpage

\section{Evaluation Prompt Template for Task 2 \& 3}
\label{app:evaluation}

Our evaluation adopts a dual-task framework combining automatic language detection with adaptive prompting to systematically measure \emph{dialogue realism} (Task 2) and \emph{persona consistency} (Task 3). This design ensures comparability and reproducibility across multilingual datasets.

\begin{tcolorbox}[title=\textbf{Prompt Template for Dialogue Evaluation (GPT-4o)},breakable]

\textbf{System Prompt}  

You are a professional dialogue evaluation expert, skilled in analyzing dialogue realism and persona consistency.  
Return the evaluation strictly in JSON format.

\par\medskip
\hrule
\par\medskip

\textbf{User Prompt Structure}

\textbf{Dialogue Context:}  
[FULL DIALOGUE CONTENT]

\textbf{Student's Final Response to Evaluate:}  
"[STUDENT'S LAST RESPONSE]"

\textbf{Target Persona Description:}  
"[TARGET PERSONALITY TRAITS DESCRIPTION]"

\textbf{Evaluation Tasks}

\textbf{Task 2 – Realism Assessment}  
Evaluate how realistic and natural the student's final response is, in line with human conversational characteristics.

\emph{Guidelines:}
\begin{itemize}
  \item Focus exclusively on the final response.
  \item Consider fluency, logical coherence, natural emotional expression; penalize mechanical or overly structured replies.
  \item Strictly penalize textbook-like or formalized answers.
  \item Response length is irrelevant: short, casual replies can be more realistic.
\end{itemize}

\emph{Scoring Scale:}
\begin{itemize}
  \item 0.9–1.0: Perfectly natural, conversational, spontaneous
  \item 0.7–0.8: Mostly natural, minor formal elements
  \item 0.5–0.6: Partly mechanical or structured
  \item 0.3–0.4: Clearly artificial or textbook-like
  \item 0.0–0.2: Completely mechanical or unrealistic
\end{itemize}

\textbf{Task 3 – Persona Consistency Assessment}  
Evaluate how naturally and realistically the student's final response reflects the \emph{target persona traits}, \textbf{relative to the provided persona description}.

\emph{Guidelines:}
\begin{itemize}
  \item Natural expression outweighs exaggerated or templated trait markers.
  \item Consistency must be judged with reference to the provided persona description.
  \item Consistency does not require repetition or extremity.
  \item Penalize repetitive, formulaic, or mechanized expressions.
  \item Allow minor deviations; focus on overall alignment.
  \item Length is not decisive; natural reflection of traits is key.
\end{itemize}

\emph{Scoring Scale:}
\begin{itemize}
  \item 0.9–1.0: Natural and realistic reflection of traits, no templated artifacts
  \item 0.7–0.8: Mostly consistent, slight templating
  \item 0.5–0.6: Partial consistency, mixed traits, less natural
  \item 0.3–0.4: Traits weak or inconsistent, but still natural
  \item 0.0–0.2: Contradicts traits and unnatural
\end{itemize}

\textbf{Important:} Evaluate only the final student response.  
Do not use previous dialogue turns as reference.

\textbf{Expected Output (strict JSON):}
\begin{verbatim}
{
  "task2_realism": {"score": 0.0-1.0},
  "task3_persona": {"score": 0.0-1.0}
}
\end{verbatim}

\end{tcolorbox}

This framework ensures robustness and reproducibility through three mechanisms. First, language is automatically detected via character-level statistics, enabling adaptive switching between Chinese and English prompts with corresponding persona descriptions. Second, a retry mechanism with exponential backoff and strict JSON parsing with fallback patterns safeguard evaluation reliability. Third, the dual-task design balances realism and persona consistency, scoring only the student’s final response to maintain contextual grounding while ensuring comparability across scenarios.

Overall, this evaluation framework transforms subjective persona assessment into a structured, replicable process, providing the methodological foundation for large-scale experiments and statistical analysis.

\clearpage

\section{Detailed Experimental Results}
\label{app:experiments}
% ================== Task 1 (24 charts, weighted averages) ==================
\subsection{Additional Analysis for Task 1}
\label{subapp:task1_figs}

\begin{figure*}[!htbp]
  \centering
  \captionsetup[subfigure]{font=small,justification=centering,singlelinecheck=false}
  
  % ========== Qwen 部分 ==========
  \textbf{Qwen (Base)}\\[2pt]
  \subcaptionbox{Behavior}[.24\linewidth]{%
    \includegraphics[width=\linewidth]{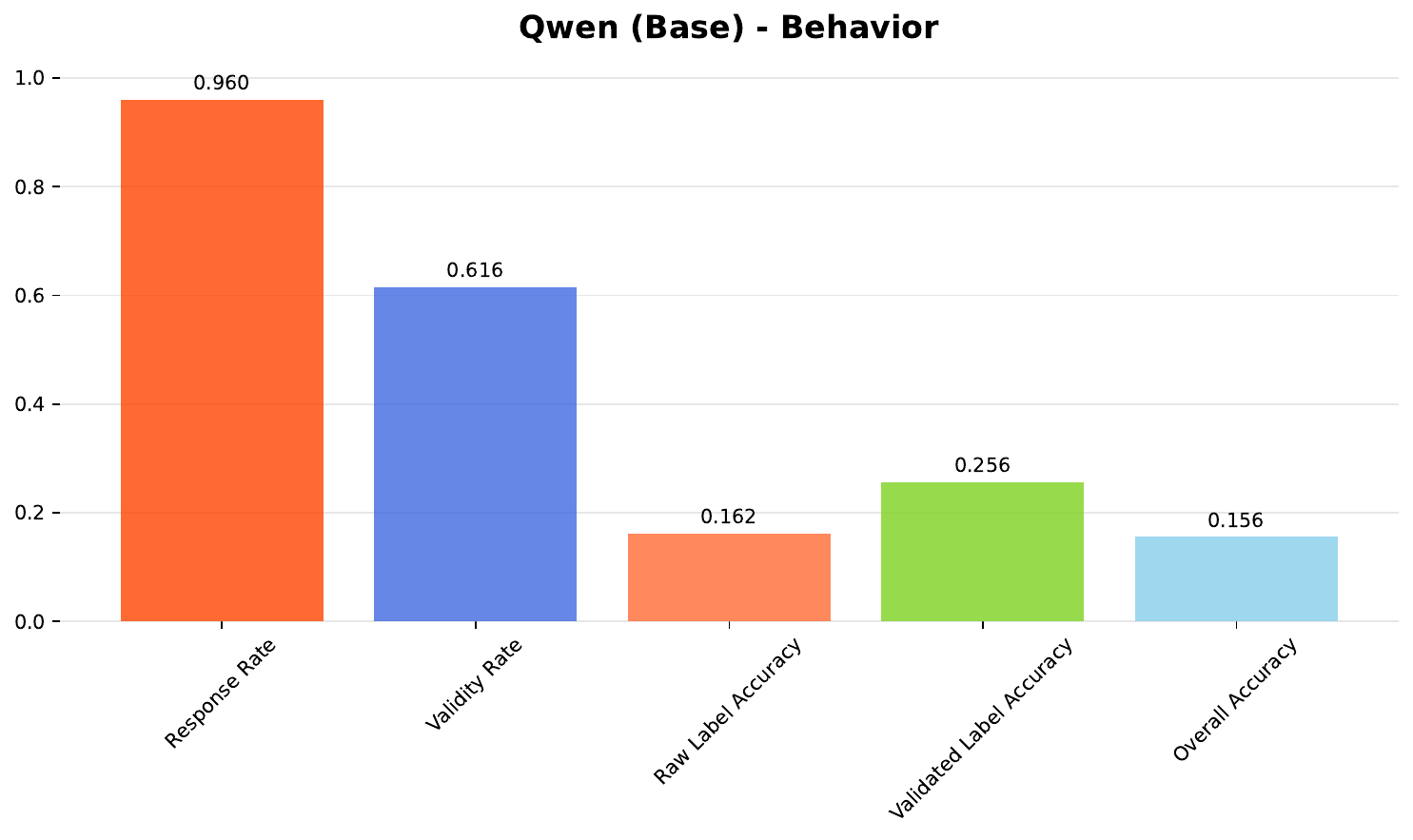}
  }\hfill
  \subcaptionbox{Emotion}[.24\linewidth]{%
    \includegraphics[width=\linewidth]{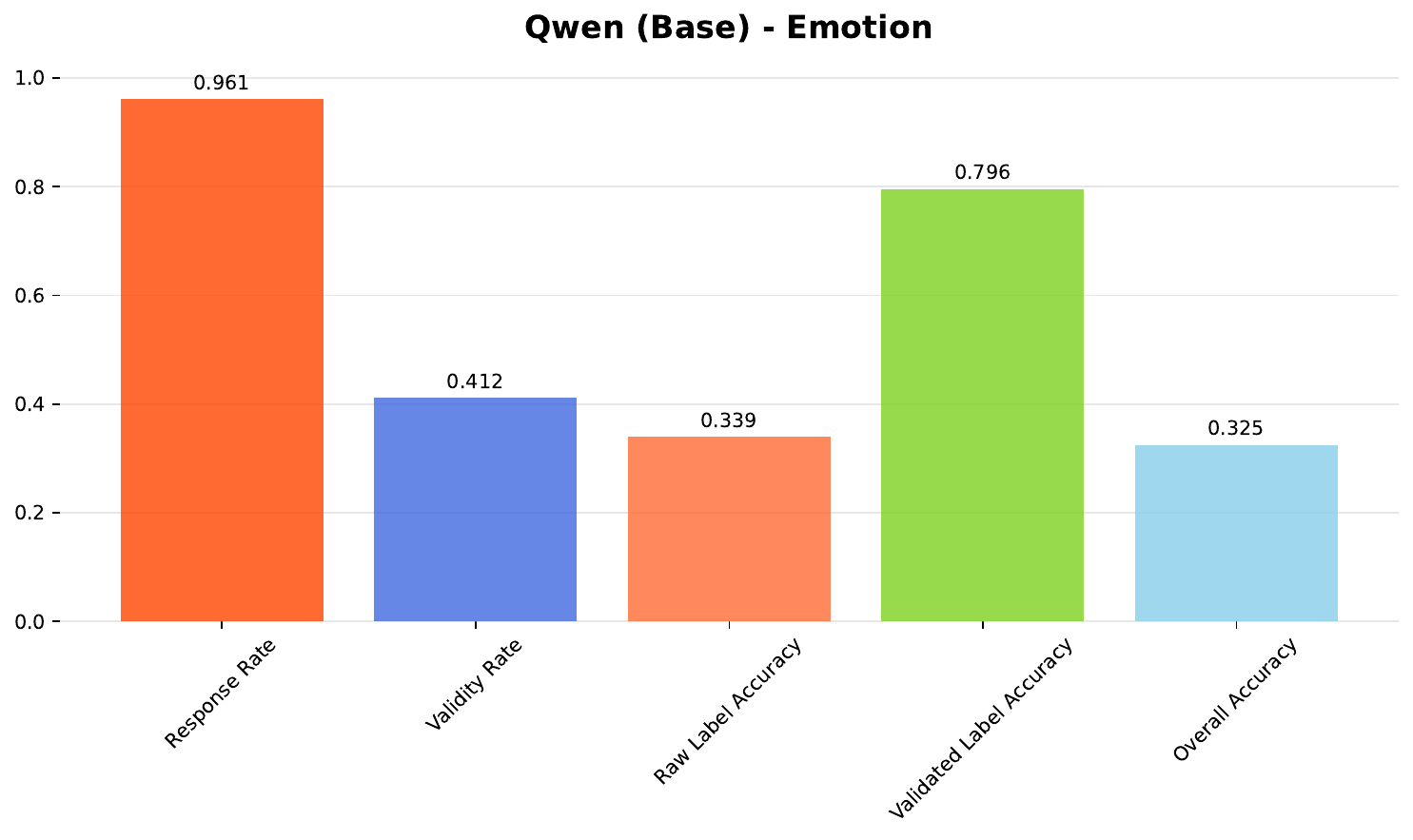}
  }\hfill
  \subcaptionbox{Expression}[.24\linewidth]{%
    \includegraphics[width=\linewidth]{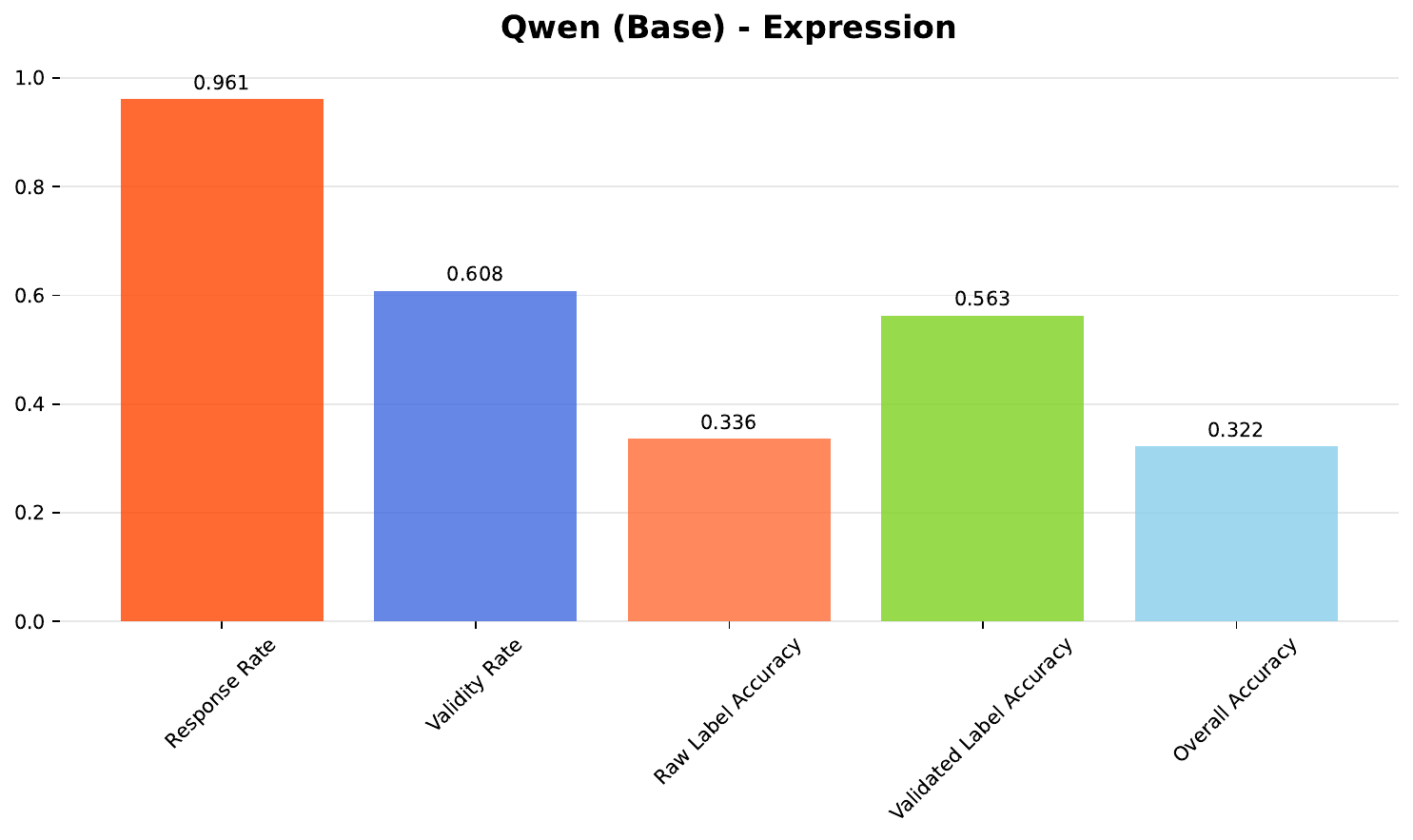}
  }\hfill
  \subcaptionbox{Voice}[.24\linewidth]{%
    \includegraphics[width=\linewidth]{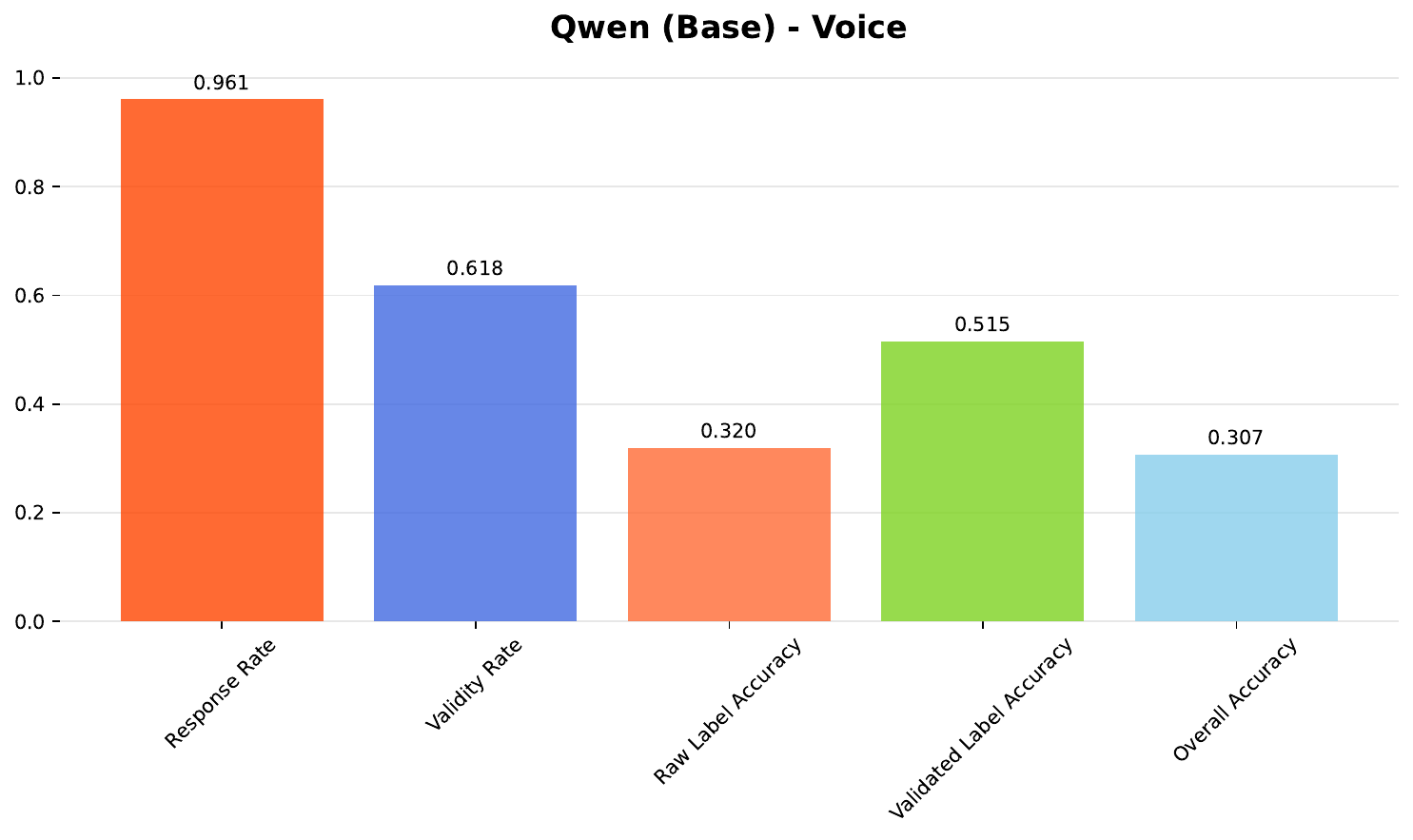}
  }

  \par\medskip

  \textbf{Qwen (Fine-Tuned)}\\[2pt]
  \subcaptionbox{Behavior}[.24\linewidth]{%
    \includegraphics[width=\linewidth]{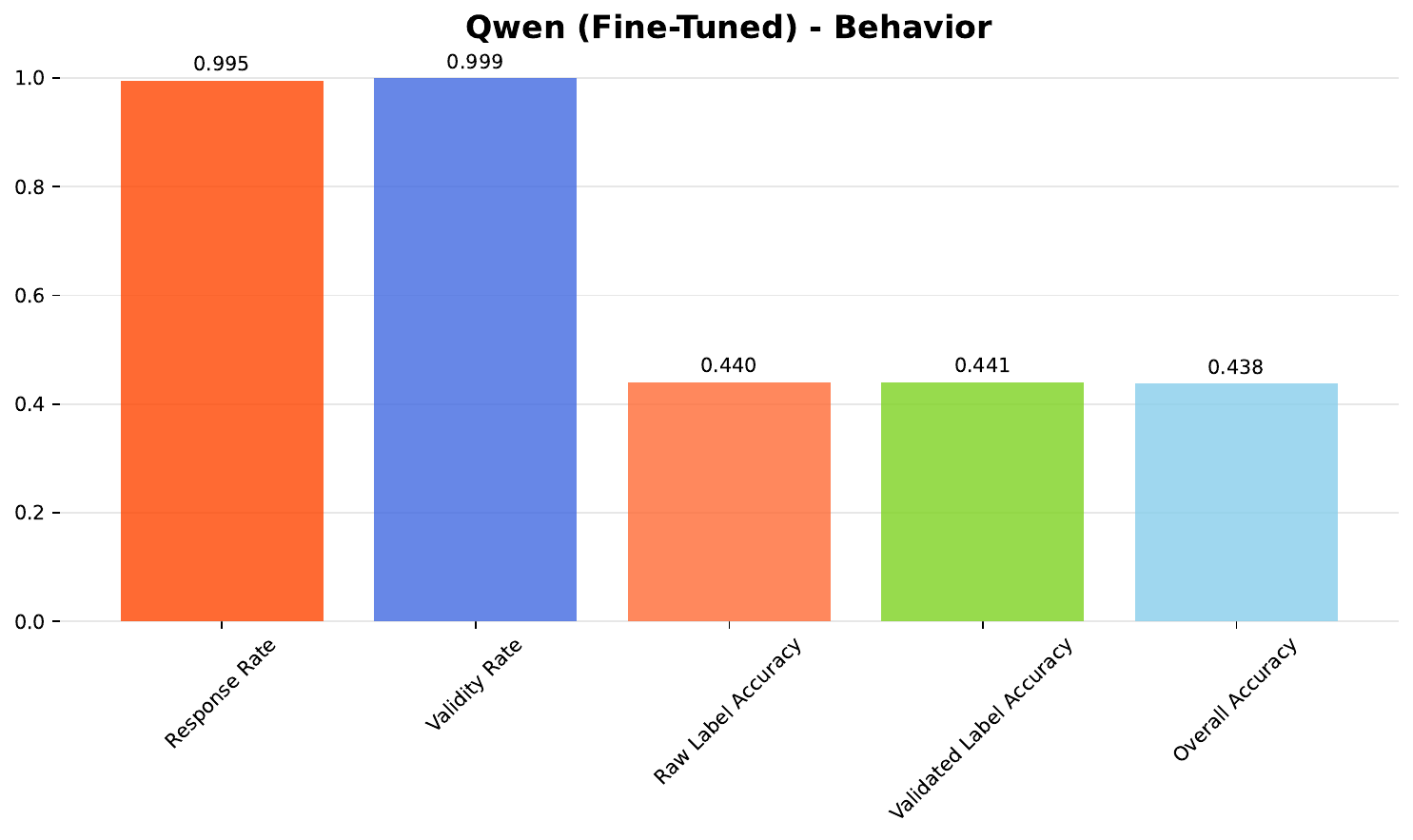}
  }\hfill
  \subcaptionbox{Emotion}[.24\linewidth]{%
    \includegraphics[width=\linewidth]{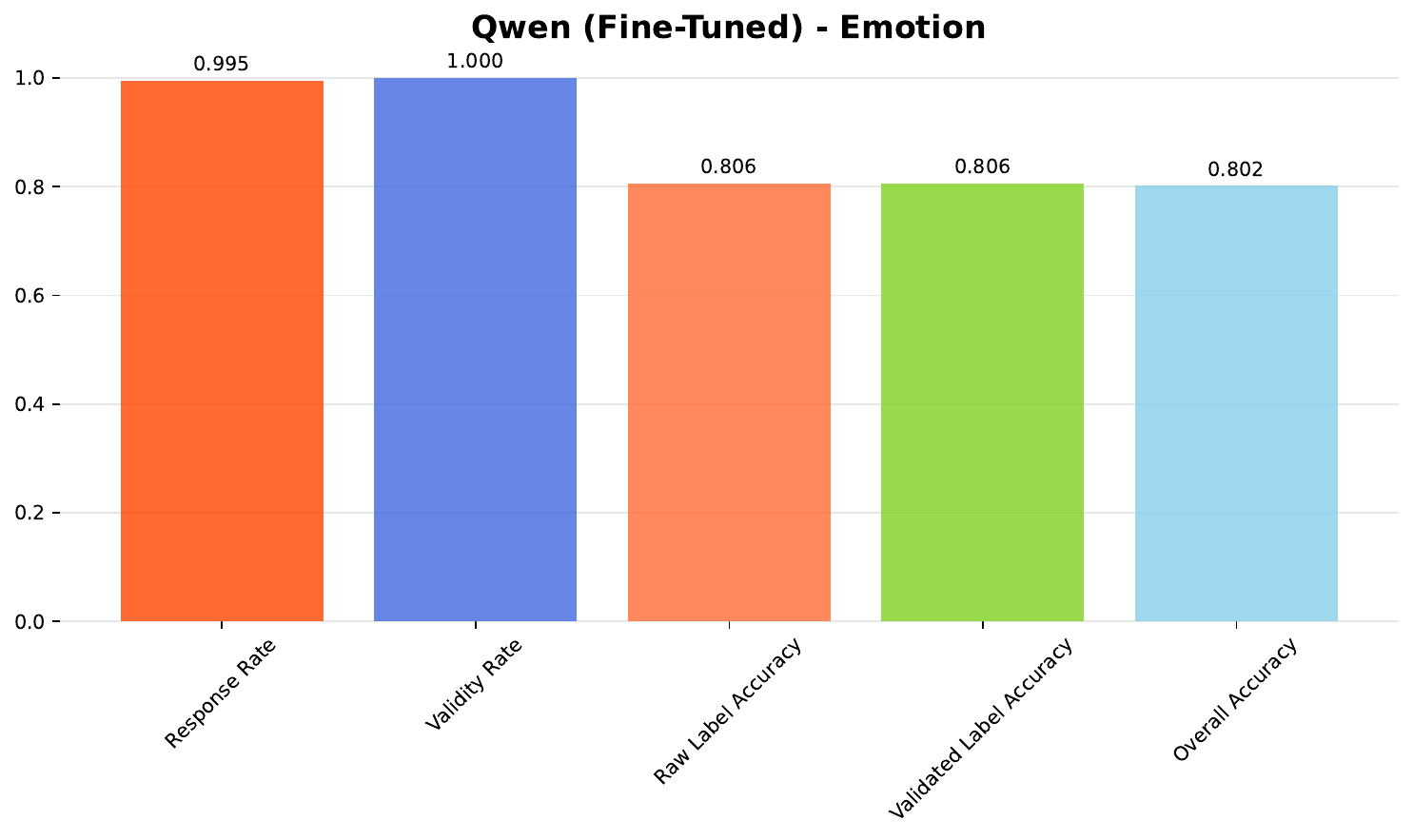}
  }\hfill
  \subcaptionbox{Expression}[.24\linewidth]{%
    \includegraphics[width=\linewidth]{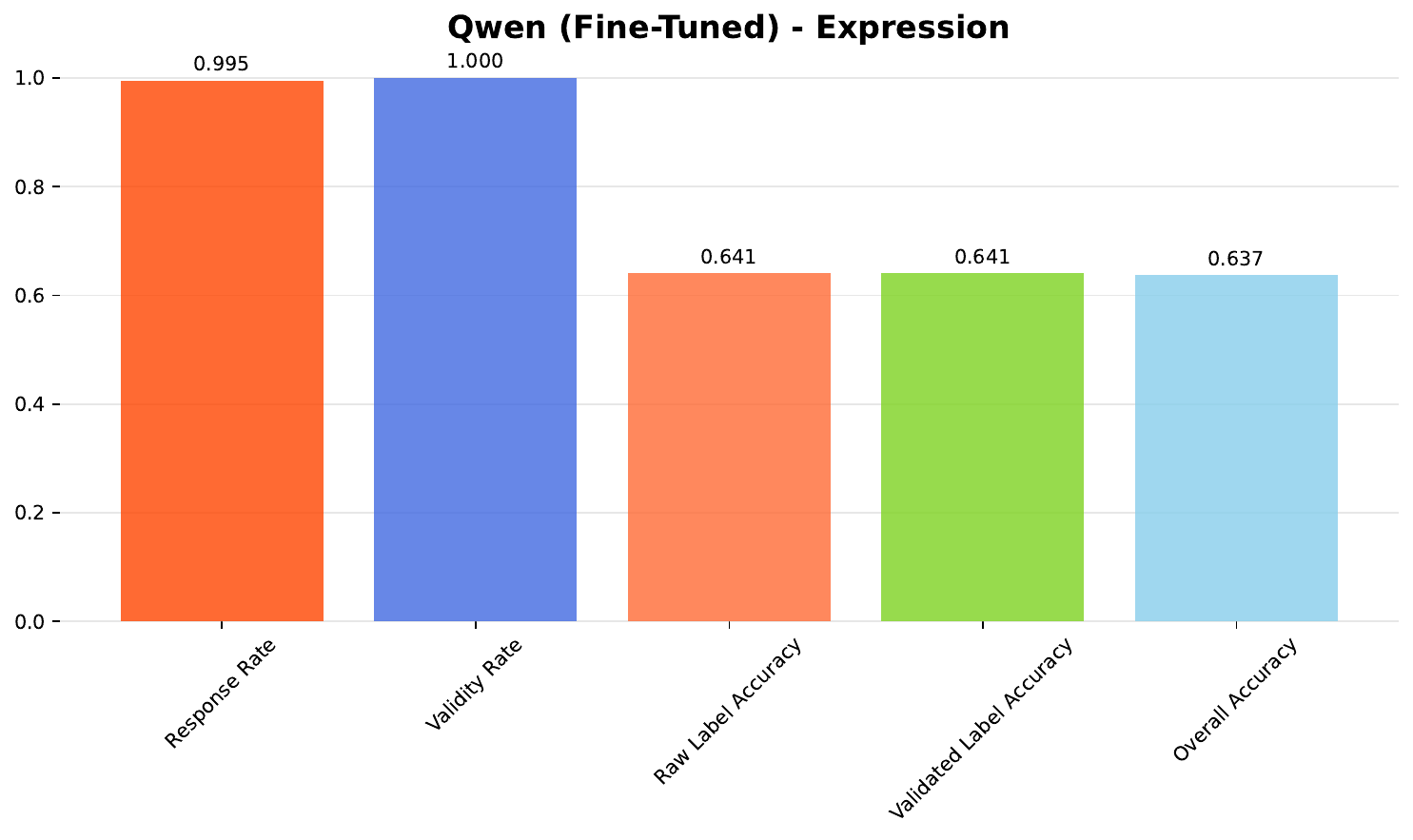}
  }\hfill
  \subcaptionbox{Voice}[.24\linewidth]{%
    \includegraphics[width=\linewidth]{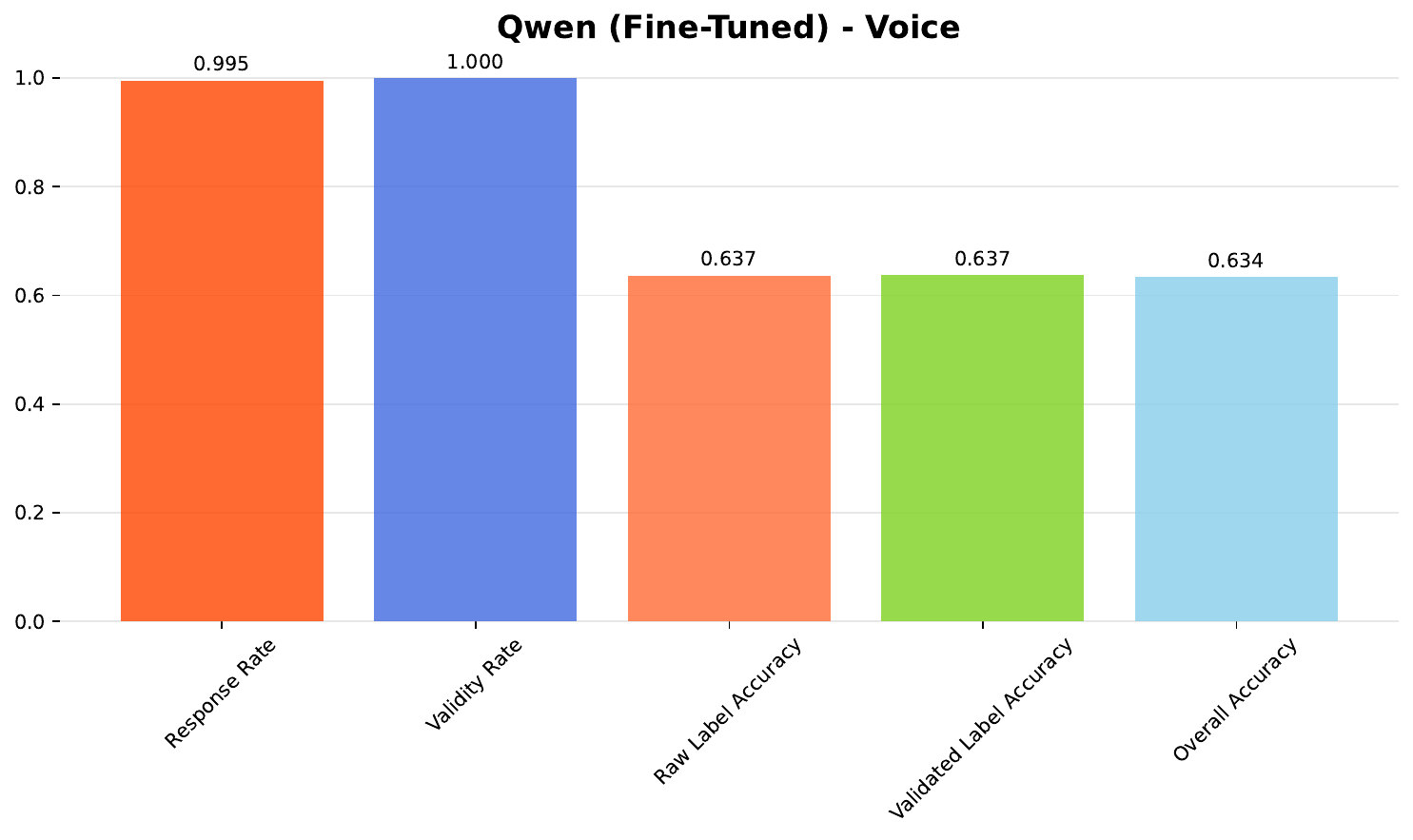}
  }

  \par\bigskip

  % ========== InternLM3 部分 ==========
  \textbf{InternLM3 (Base)}\\[2pt]
  \subcaptionbox{Behavior}[.24\linewidth]{%
    \includegraphics[width=\linewidth]{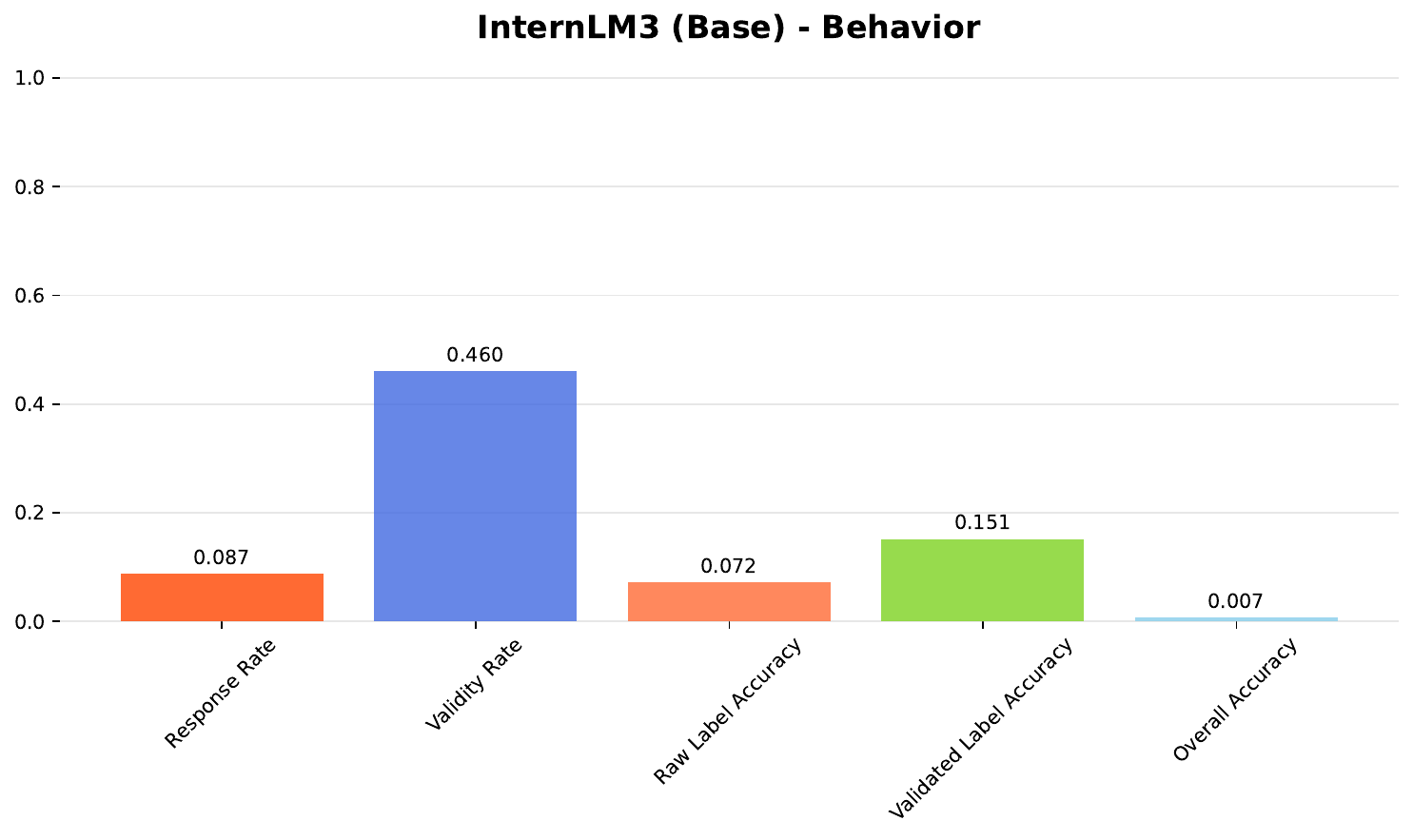}
  }\hfill
  \subcaptionbox{Emotion}[.24\linewidth]{%
    \includegraphics[width=\linewidth]{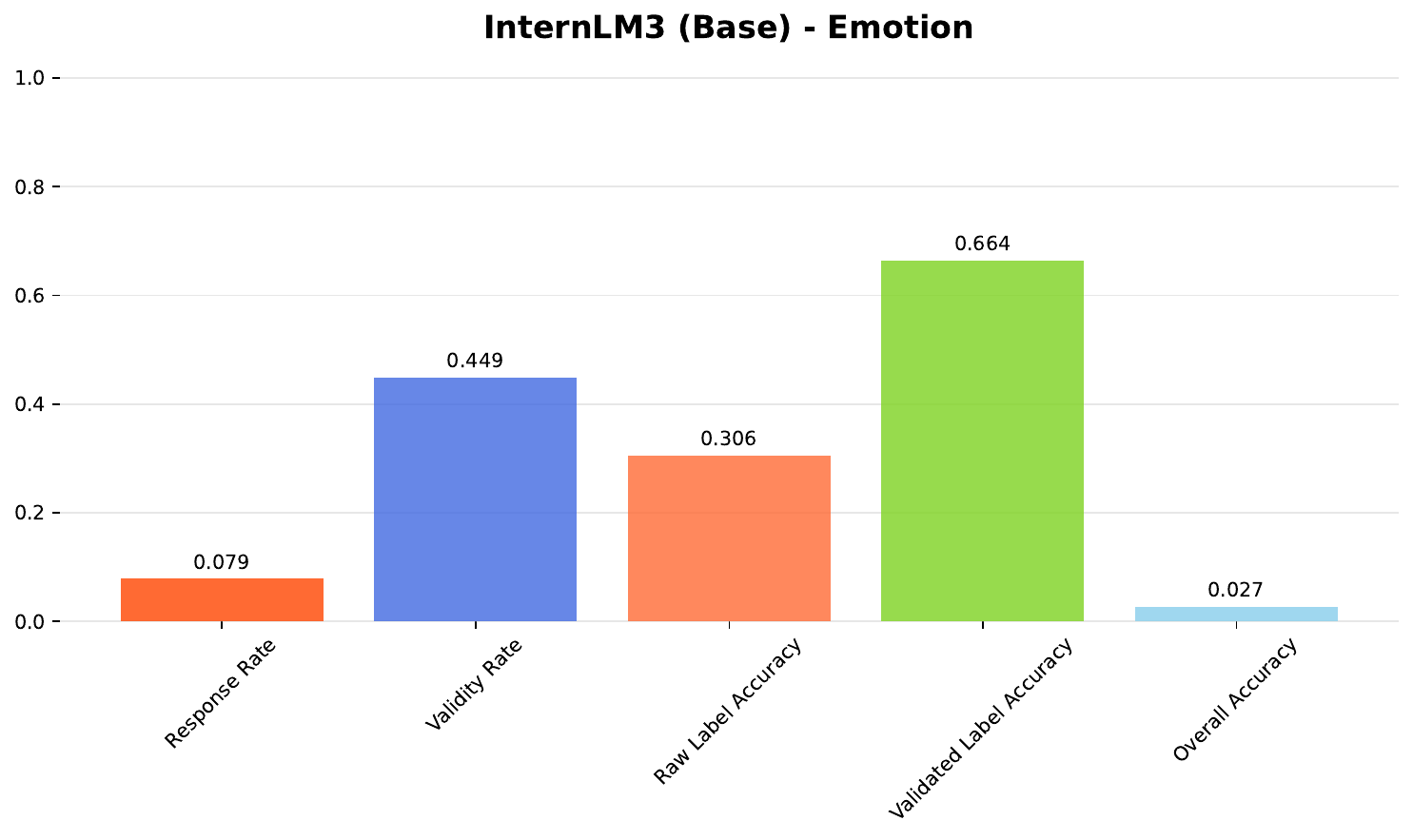}
  }\hfill
  \subcaptionbox{Expression}[.24\linewidth]{%
    \includegraphics[width=\linewidth]{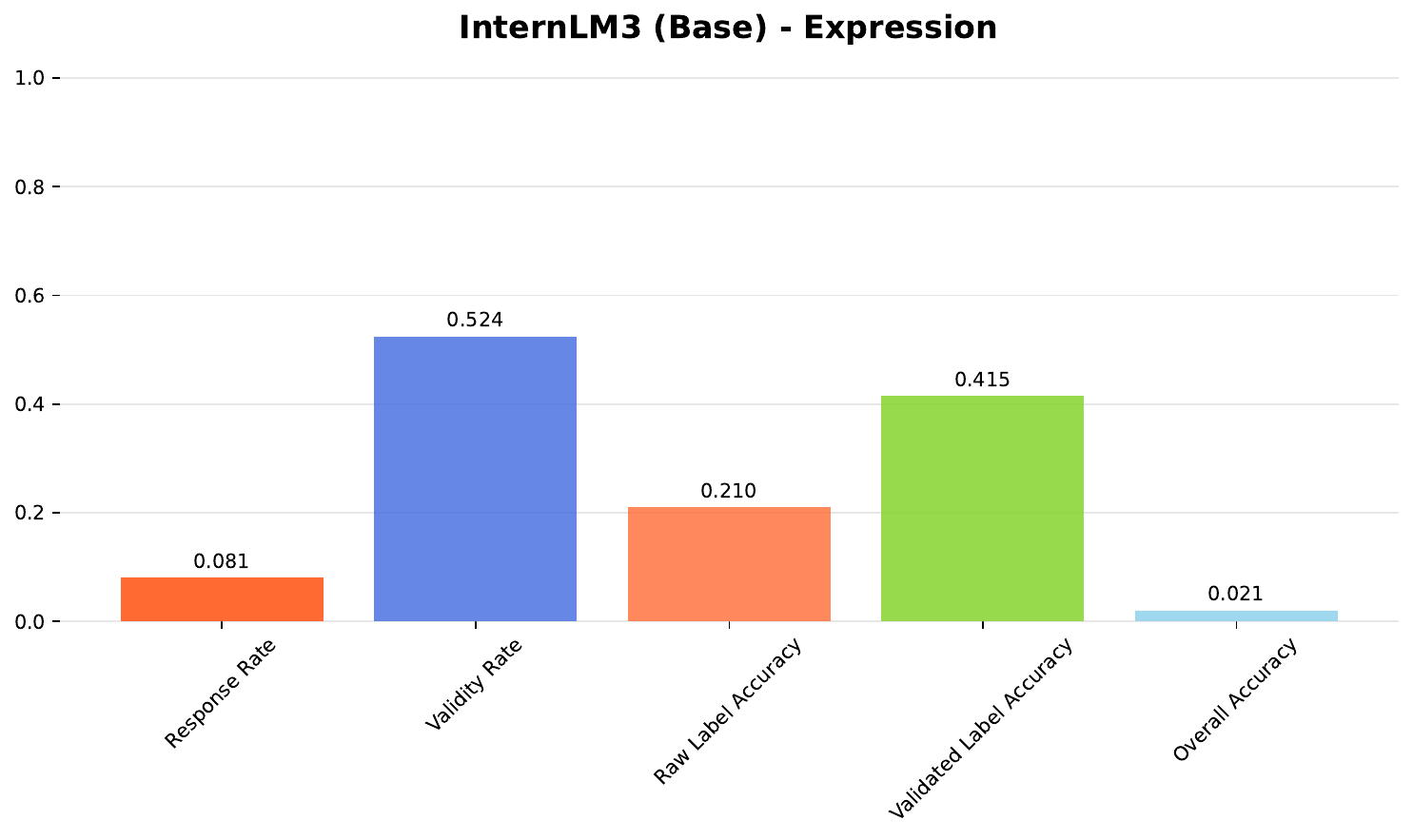}
  }\hfill
  \subcaptionbox{Voice}[.24\linewidth]{%
    \includegraphics[width=\linewidth]{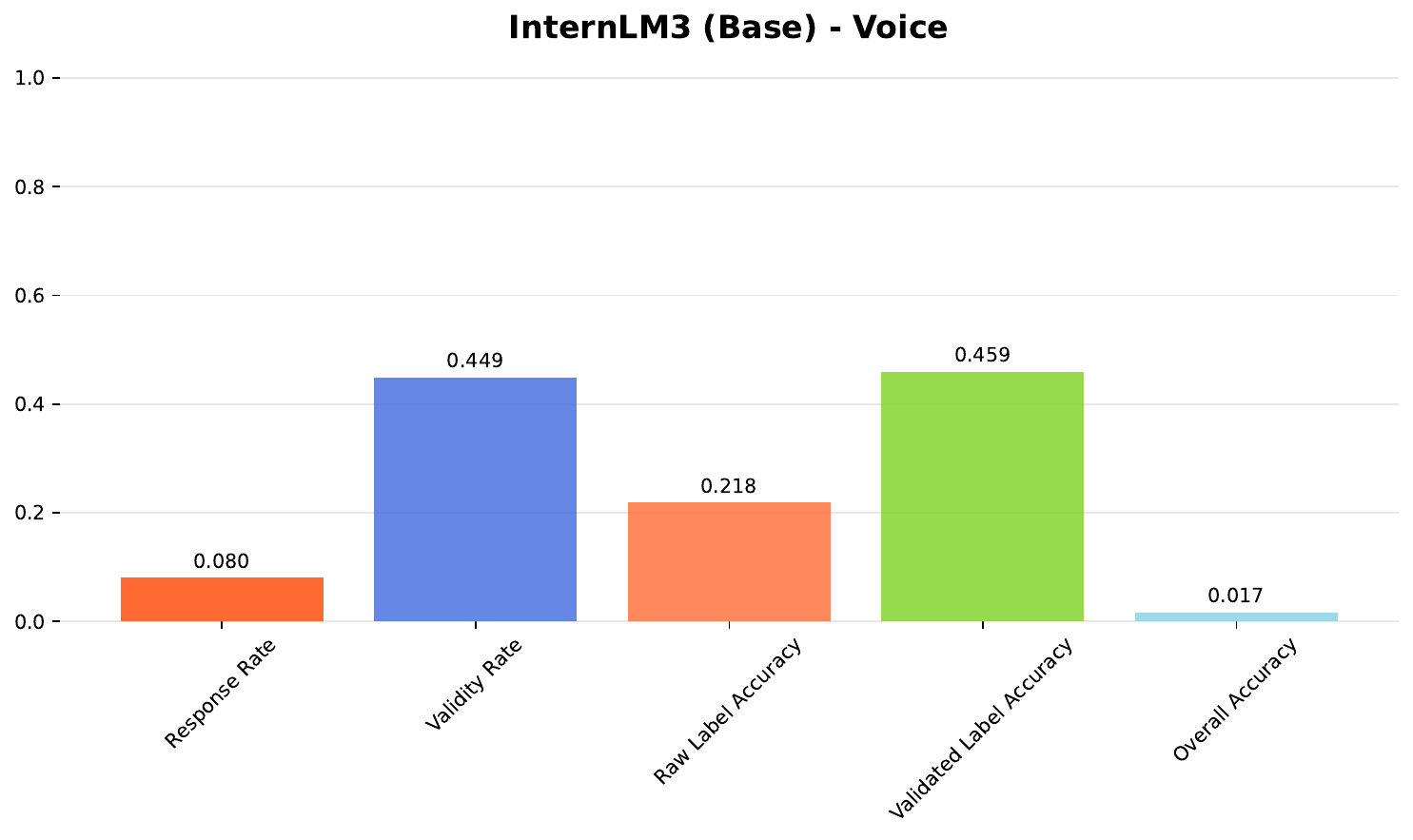}
  }

  \par\medskip

  \textbf{InternLM3 (Fine-Tuned)}\\[2pt]
  \subcaptionbox{Behavior}[.24\linewidth]{%
    \includegraphics[width=\linewidth]{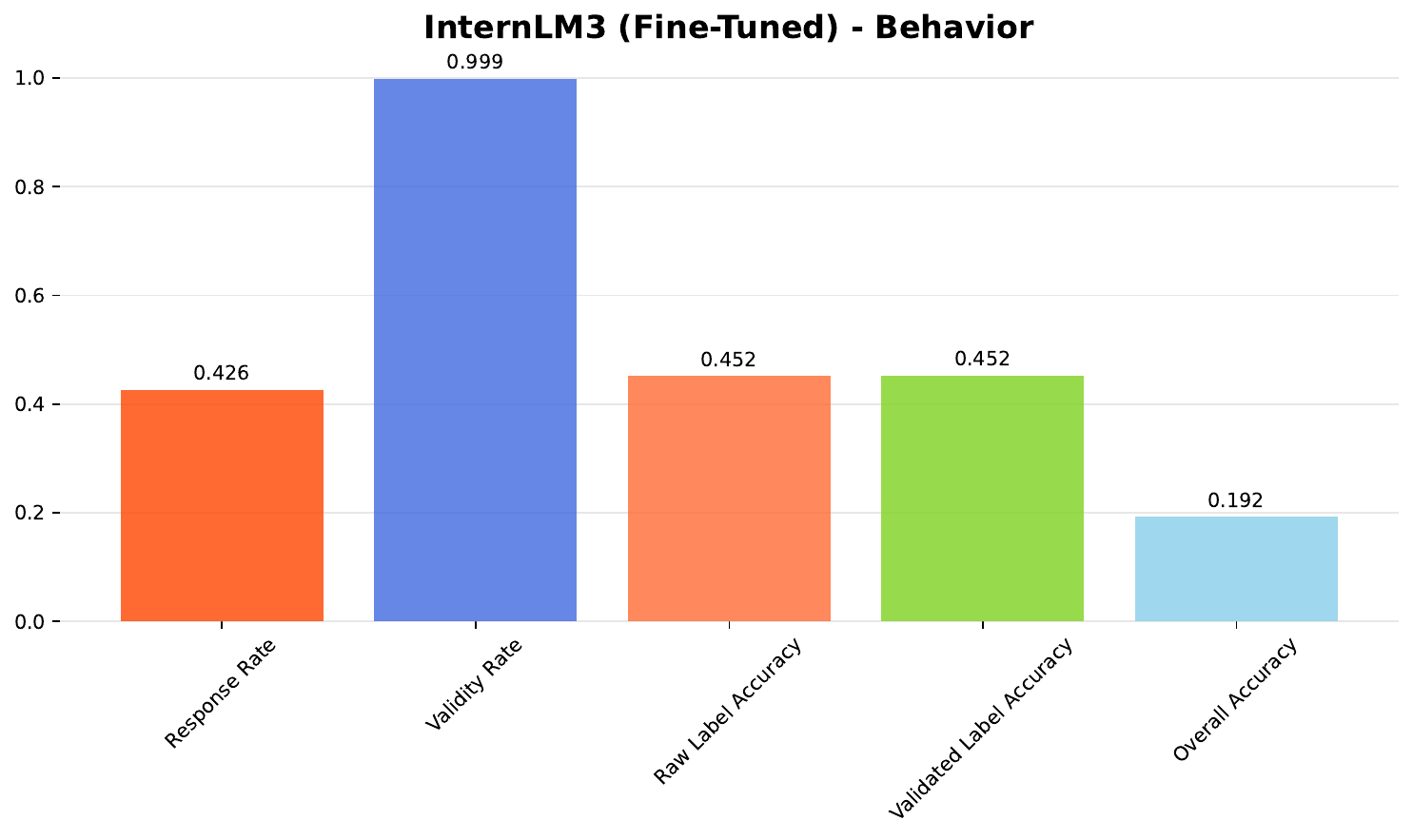}
  }\hfill
  \subcaptionbox{Emotion}[.24\linewidth]{%
    \includegraphics[width=\linewidth]{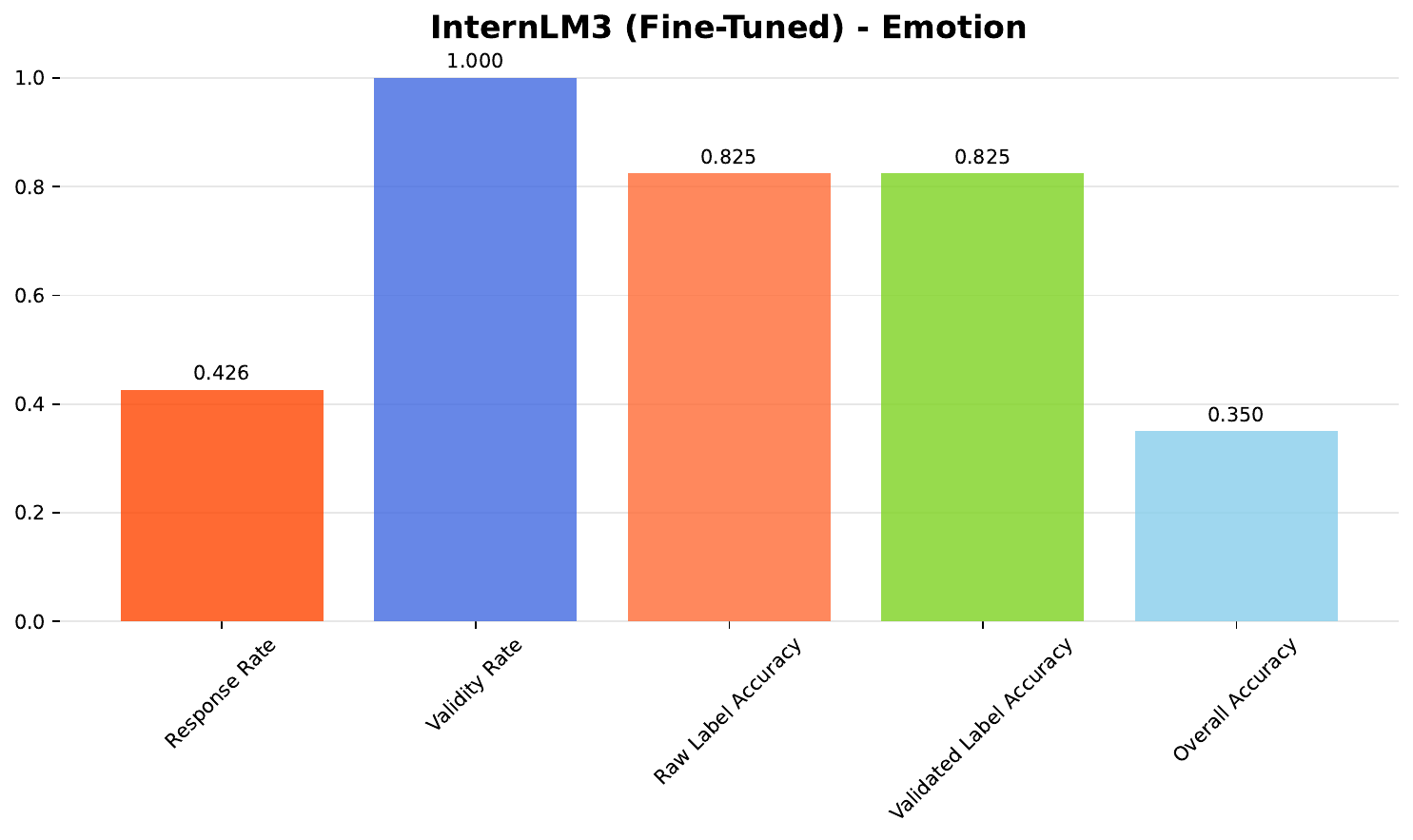}
  }\hfill
  \subcaptionbox{Expression}[.24\linewidth]{%
    \includegraphics[width=\linewidth]{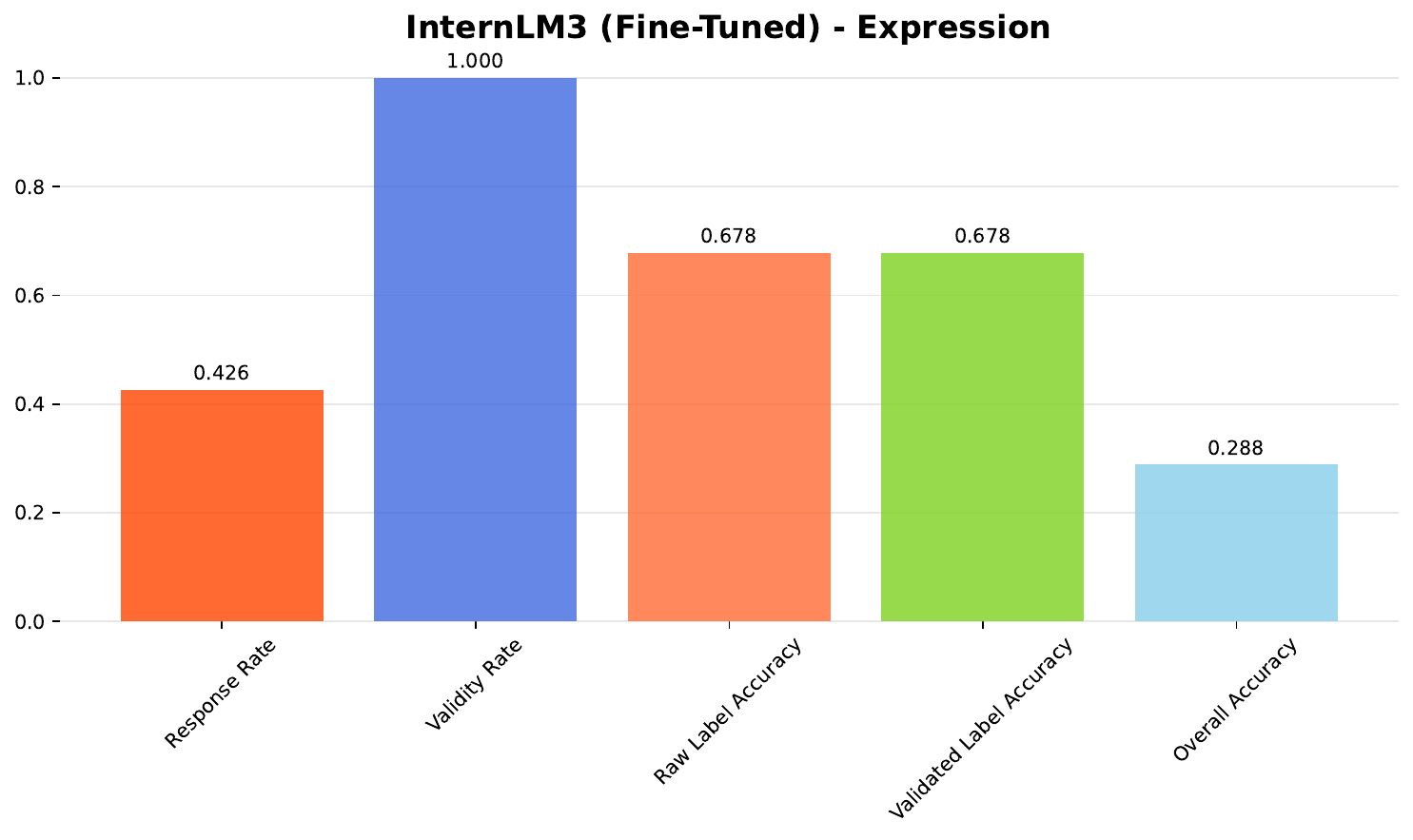}
  }\hfill
  \subcaptionbox{Voice}[.24\linewidth]{%
    \includegraphics[width=\linewidth]{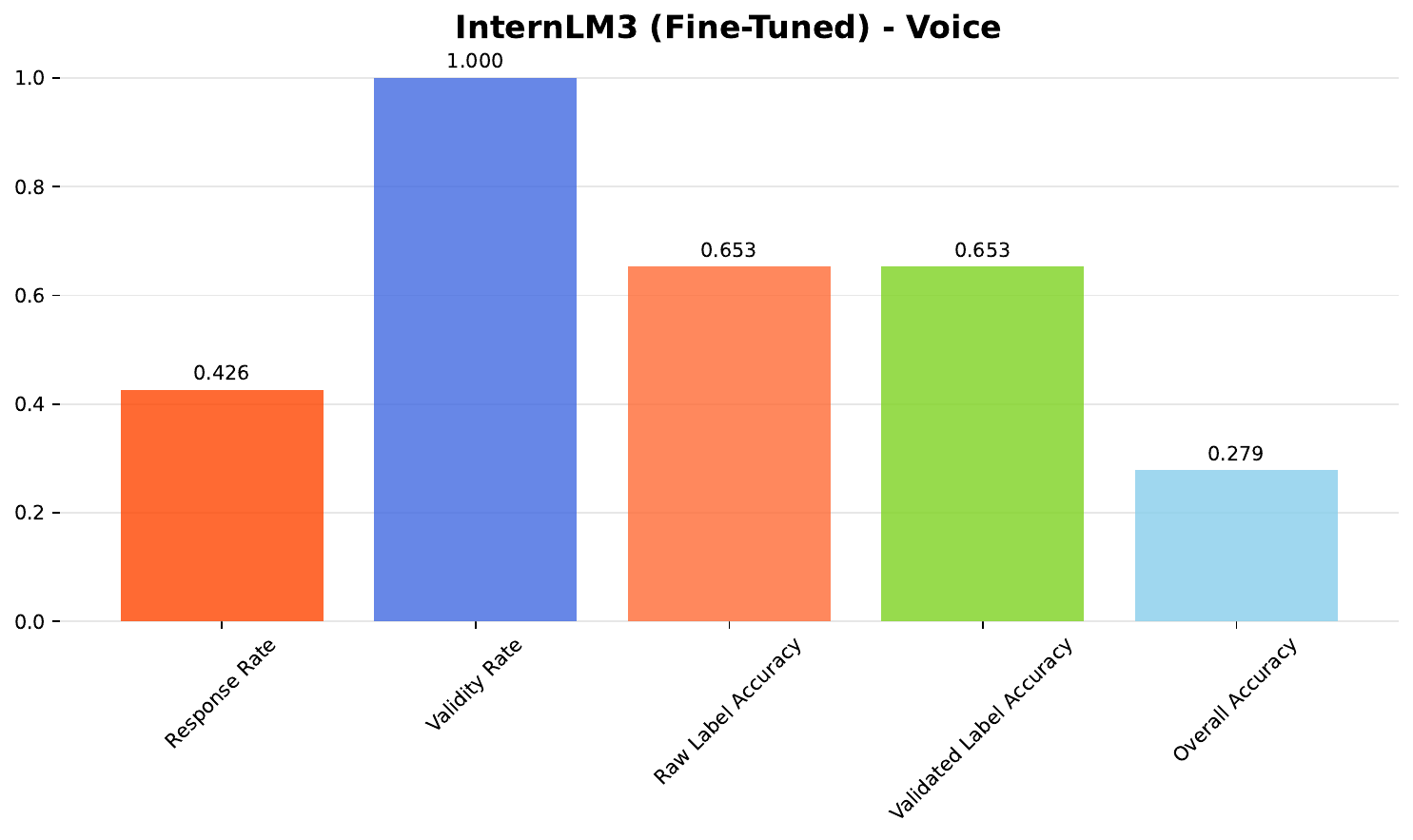}
  }

  \par\bigskip

  % ========== DeepSeek 部分 ==========
  \textbf{DeepSeek (Base)}\\[2pt]
  \subcaptionbox{Behavior}[.24\linewidth]{%
    \includegraphics[width=\linewidth]{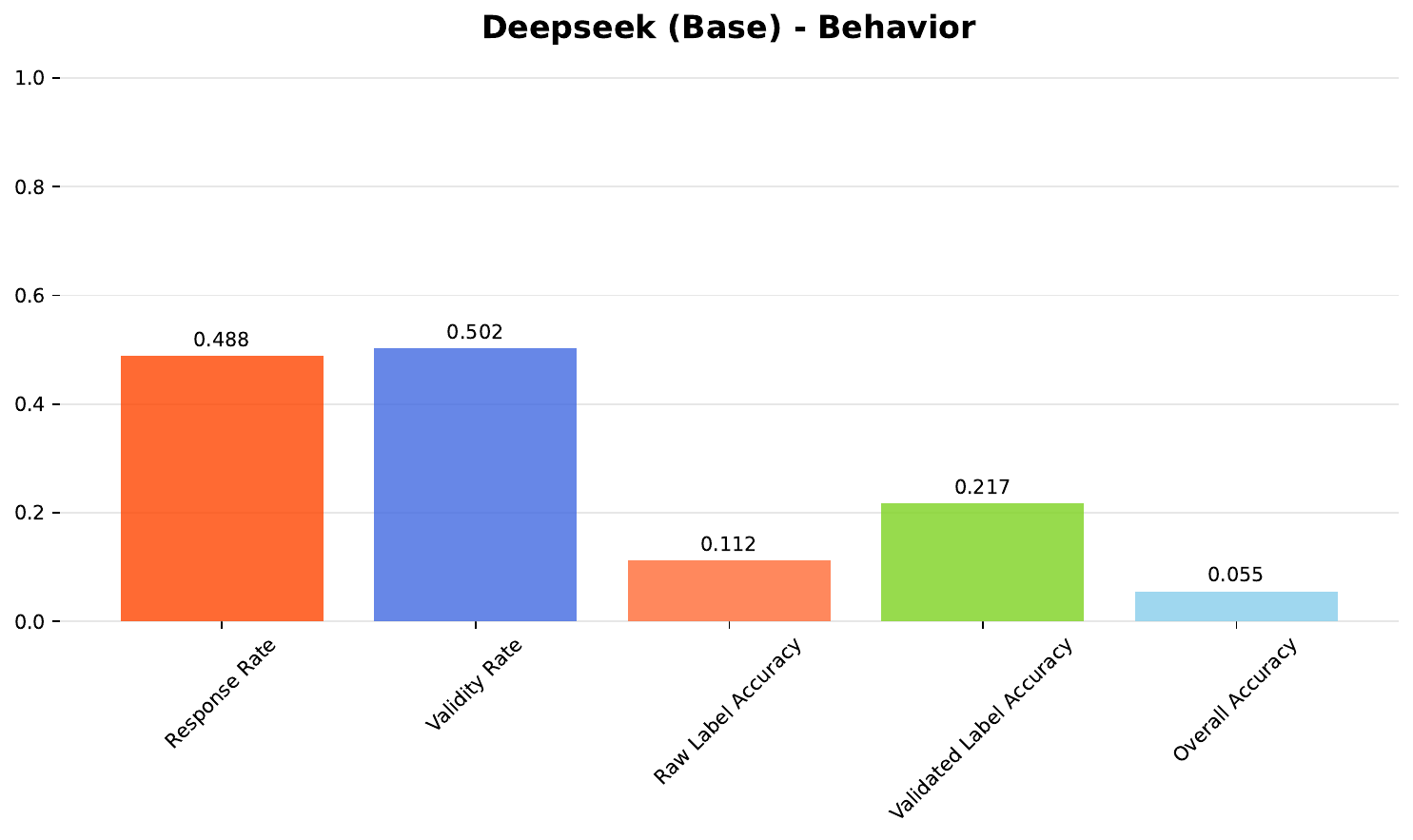}
  }\hfill
  \subcaptionbox{Emotion}[.24\linewidth]{%
    \includegraphics[width=\linewidth]{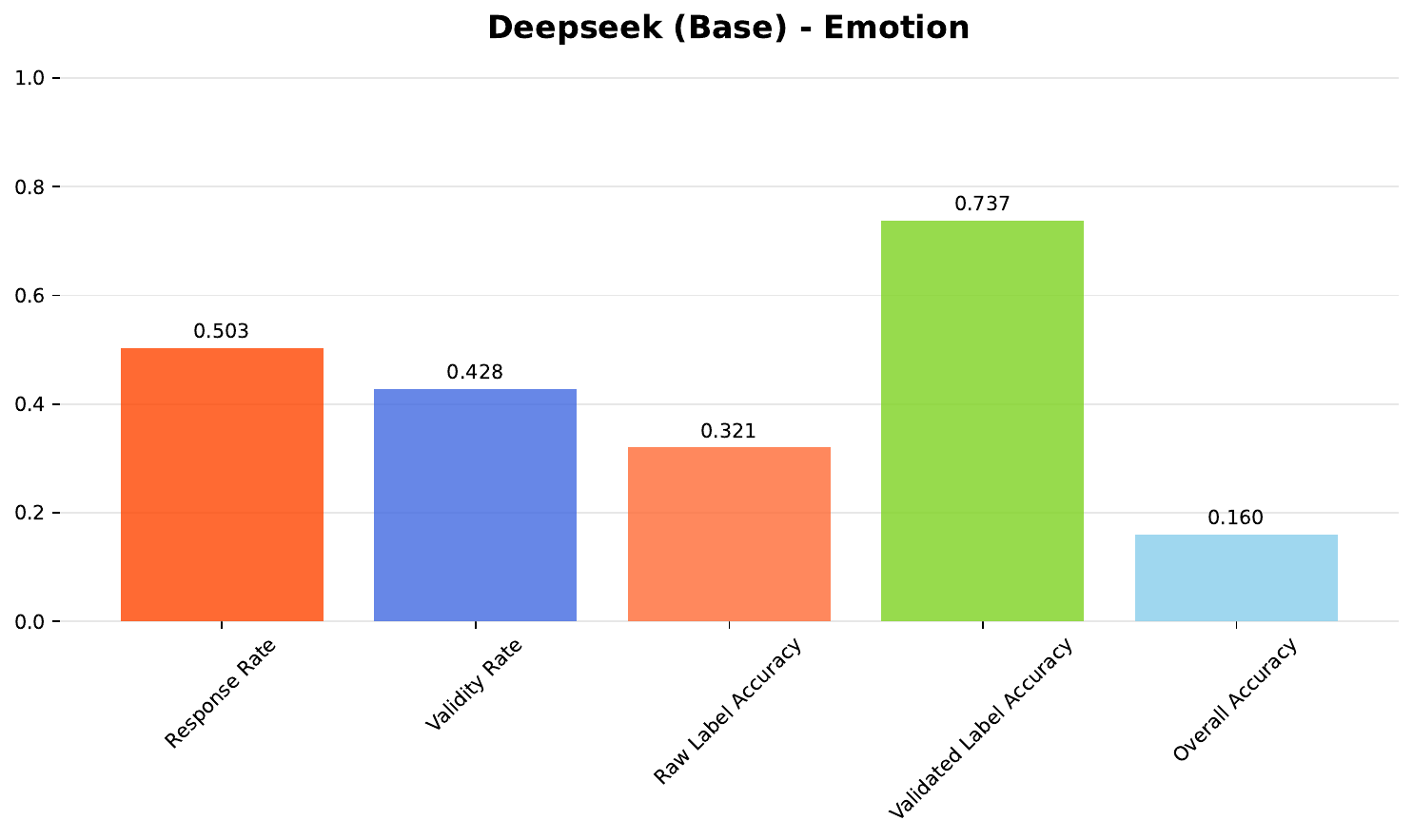}
  }\hfill
  \subcaptionbox{Expression}[.24\linewidth]{%
    \includegraphics[width=\linewidth]{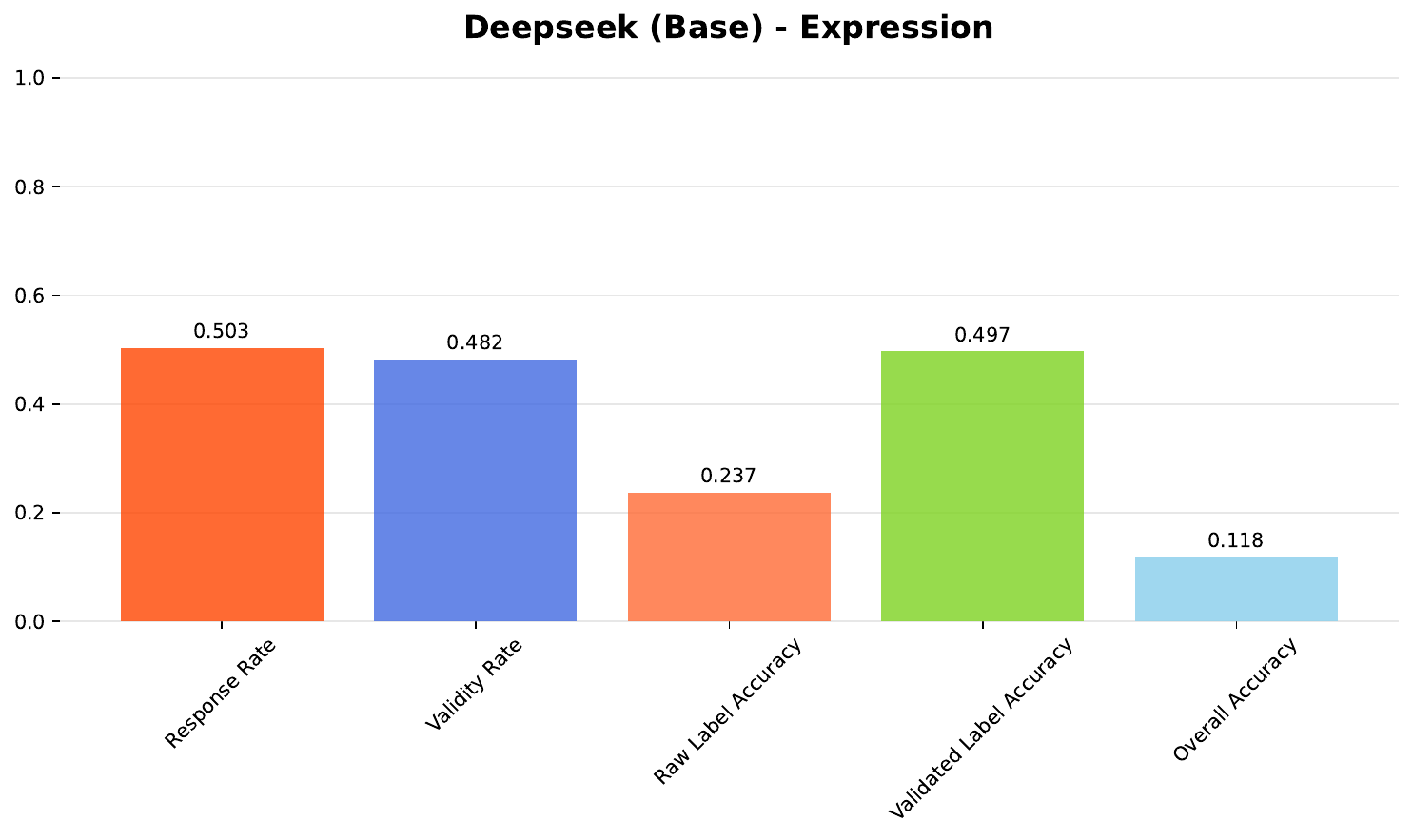}
  }\hfill
  \subcaptionbox{Voice}[.24\linewidth]{%
    \includegraphics[width=\linewidth]{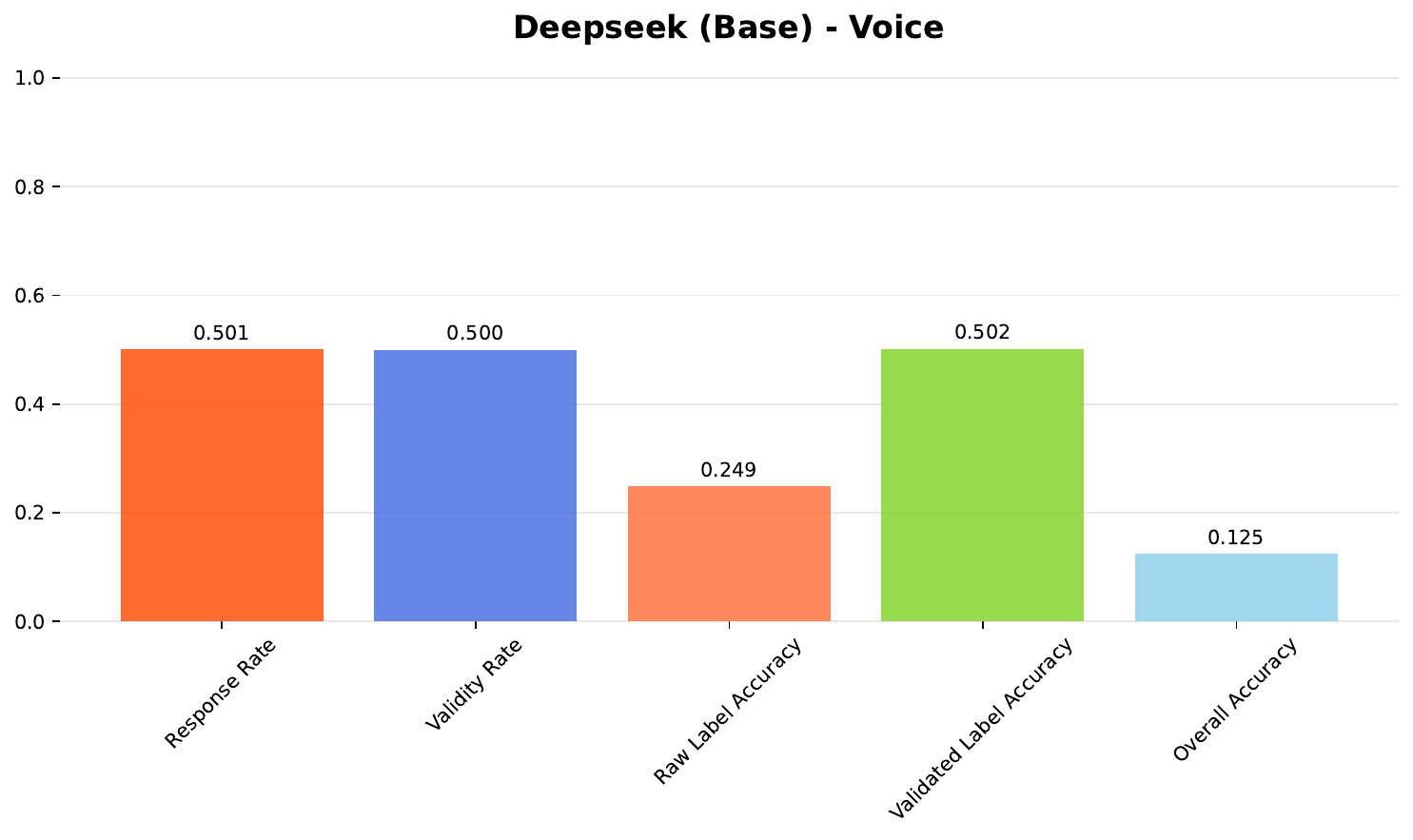}
  }

  \par\medskip

  \textbf{DeepSeek (Fine-Tuned)}\\[2pt]
  \subcaptionbox{Behavior}[.24\linewidth]{%
    \includegraphics[width=\linewidth]{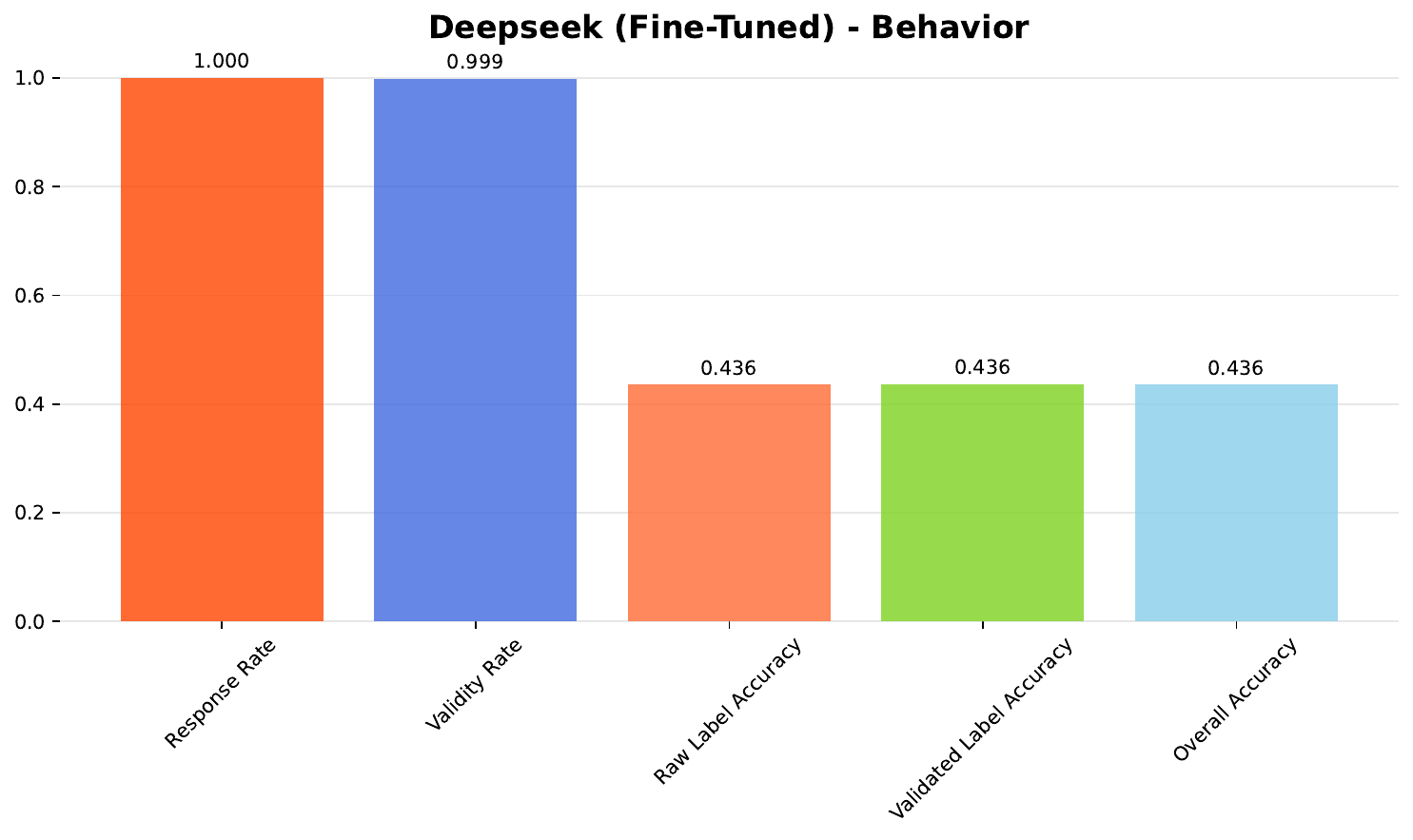}
  }\hfill
  \subcaptionbox{Emotion}[.24\linewidth]{%
    \includegraphics[width=\linewidth]{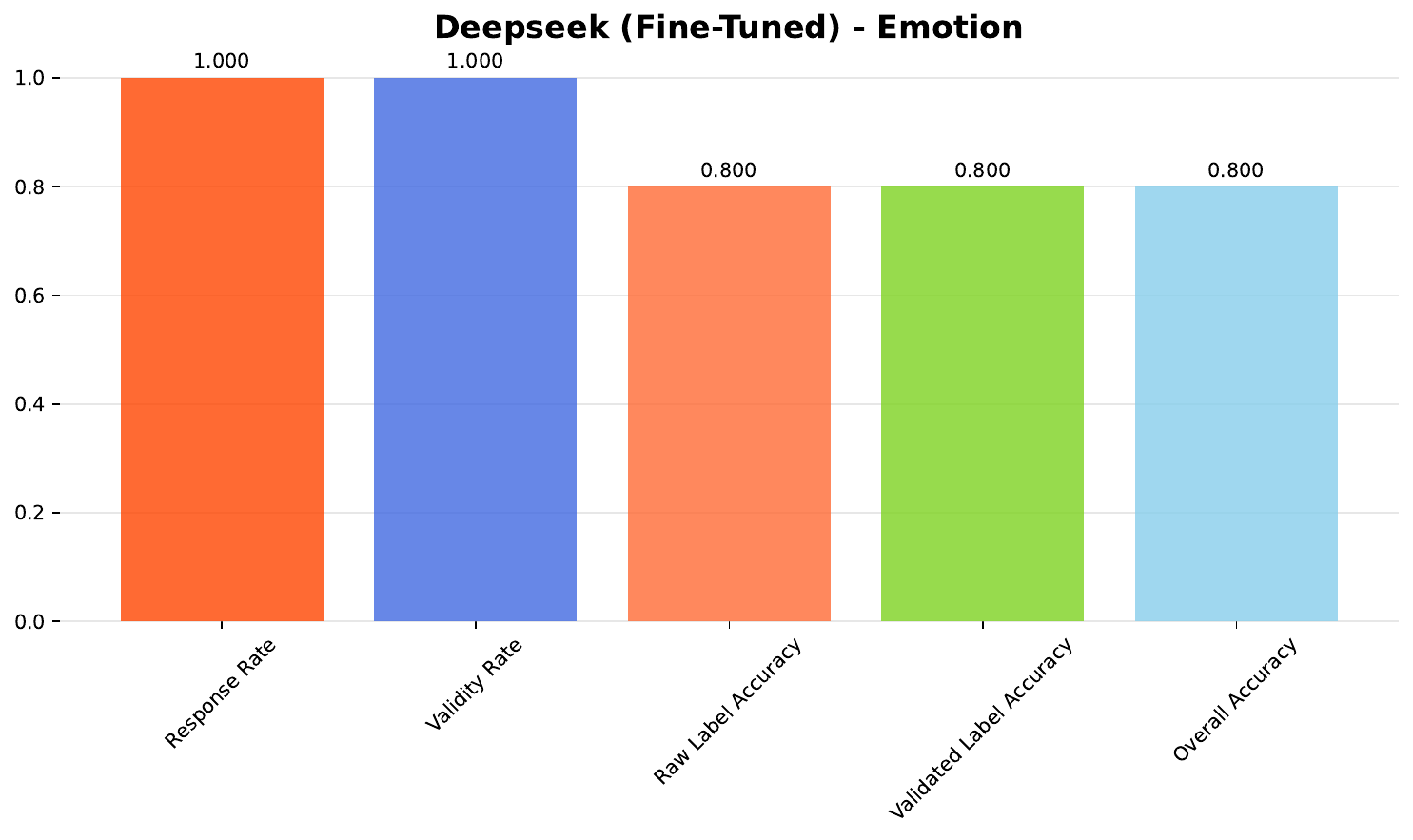}
  }\hfill
  \subcaptionbox{Expression}[.24\linewidth]{%
    \includegraphics[width=\linewidth]{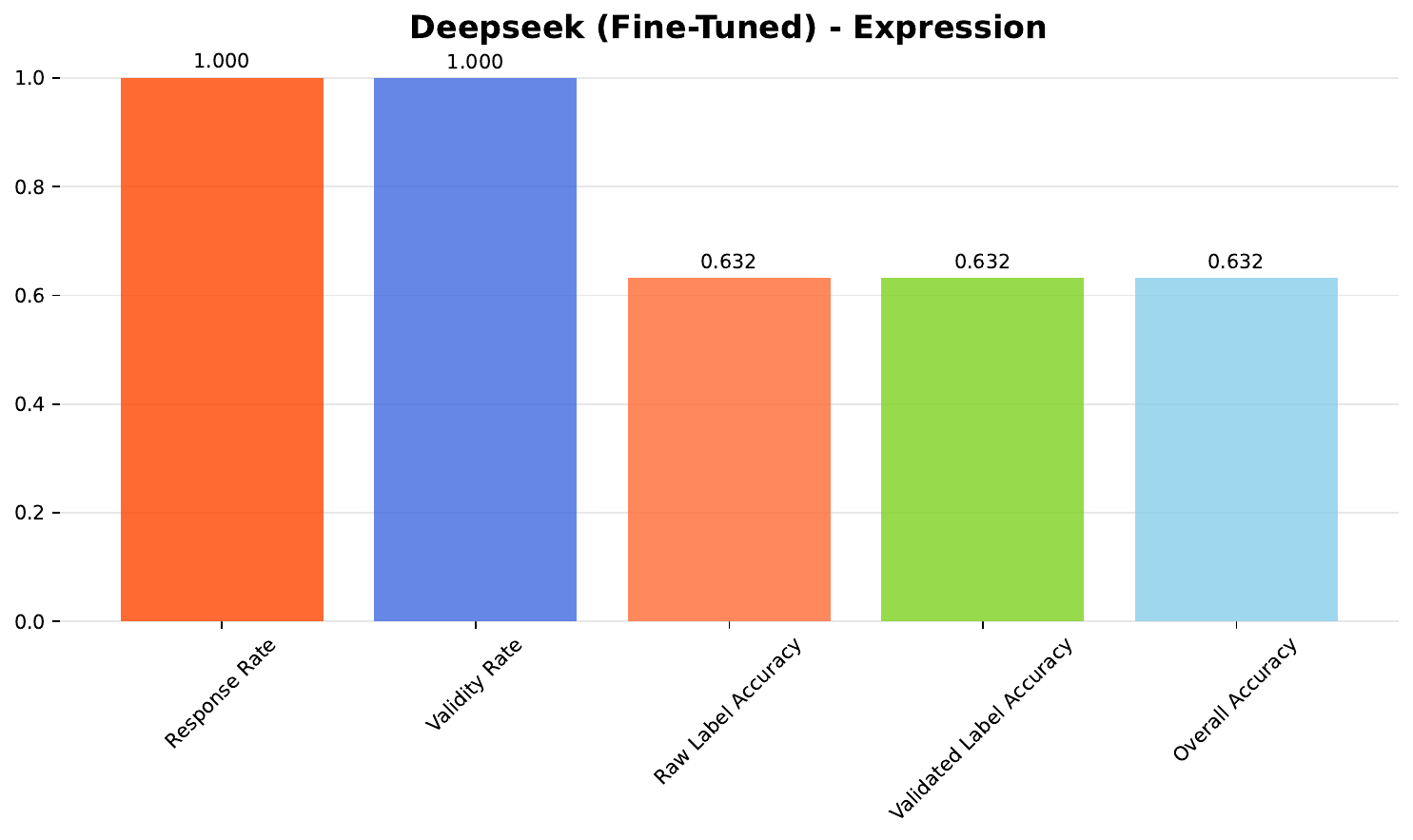}
  }\hfill
  \subcaptionbox{Voice}[.24\linewidth]{%
    \includegraphics[width=\linewidth]{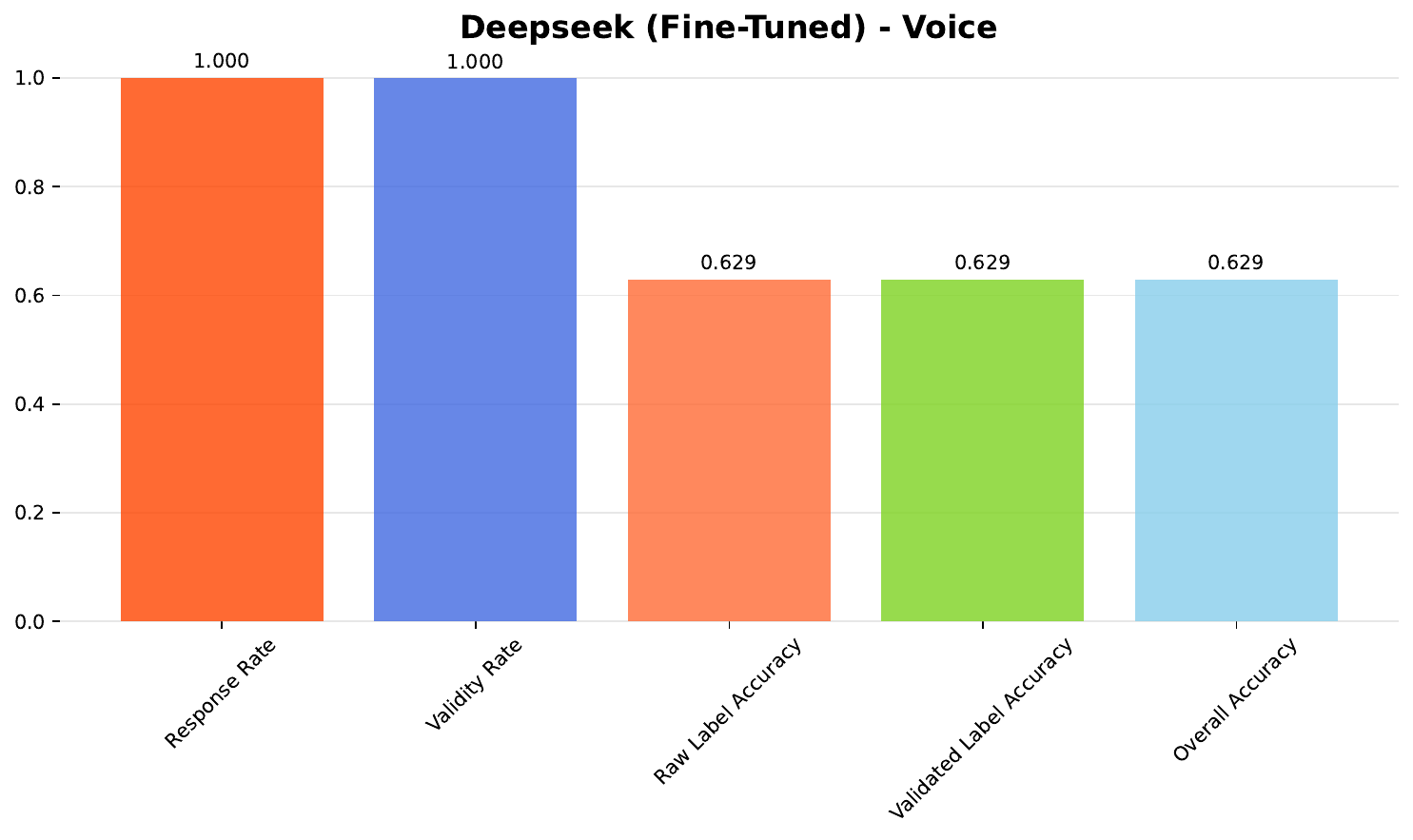}
  }

  \caption{\textbf{Task 1 — All model families (24 charts).} Weighted averages over ten personas; five metrics per chart for each model family (Qwen, InternLM3, DeepSeek) showing both base and fine-tuned variants across four dimensions (Behavior, Emotion, Expression, Voice).}
  \label{fig:task1-all}
\end{figure*}

Fig.~\ref{fig:task1-all} presents a comprehensive view of Task~1 across three model families, covering both base and fine-tuned settings, four behavioral dimensions, and five evaluation metrics. This integrated view makes it possible to trace how fine-tuning improves performance at both the overall and dimension-specific levels.

The end-to-end gains can be understood as the combined effect of three factors: whether the model produces an output at all (response coverage), whether the output is structurally valid (vocabulary compliance), and whether the valid outputs align correctly with the reference labels (in-vocabulary discrimination). For Qwen and DeepSeek, structural validity is already saturated after fine-tuning, so their improvements mainly come from higher label alignment together with strong response coverage. By contrast, InternLM3 also benefits from improved label alignment, but its relatively low response rate continues to limit its overall performance.

A consistent difficulty ordering is observed across dimensions: \textbf{Emotion is the easiest, Behavior the hardest, and Expression/Voice fall in between.} For example, under fine-tuning, Emotion reaches a mean overall accuracy above 0.65, while Behavior lags at around 0.35. This pattern indicates that residual errors are not random but stem from the inherent challenge of classifying pedagogical acts in Behavior, which requires stronger discourse- and intent-level modeling.

Family-wise, Qwen and DeepSeek converge after fine-tuning, both achieving strict end-to-end accuracy around 0.62. InternLM3, despite showing the largest relative improvement, remains constrained by its limited response coverage. These results suggest that, in this task, the performance ceiling is determined more by dataset design and task structure than by model scale alone.

Looking forward, since structural validity is already near perfect, the most promising directions for further improvement are: \textbf{(i) increasing response coverage to reduce empty outputs}, and \textbf{(ii) strengthening fine-grained guidance and exemplars for the Behavior dimension to improve label discrimination.} All of these trends are clearly reflected in the 24 subplots of Fig.~\ref{fig:task1-all}.

\clearpage

\subsection{Additional Analysis for Task 2}
\label{subapp:task2_figs}

\begin{figure*}[!htbp]
  \centering
  \captionsetup[subfigure]{font=small,justification=centering,singlelinecheck=false}
  
  % First row: HA, HC, HE, HN, HO
  \subcaptionbox{High\\Agreeableness}[.19\linewidth]{%
    \includegraphics[width=\linewidth]{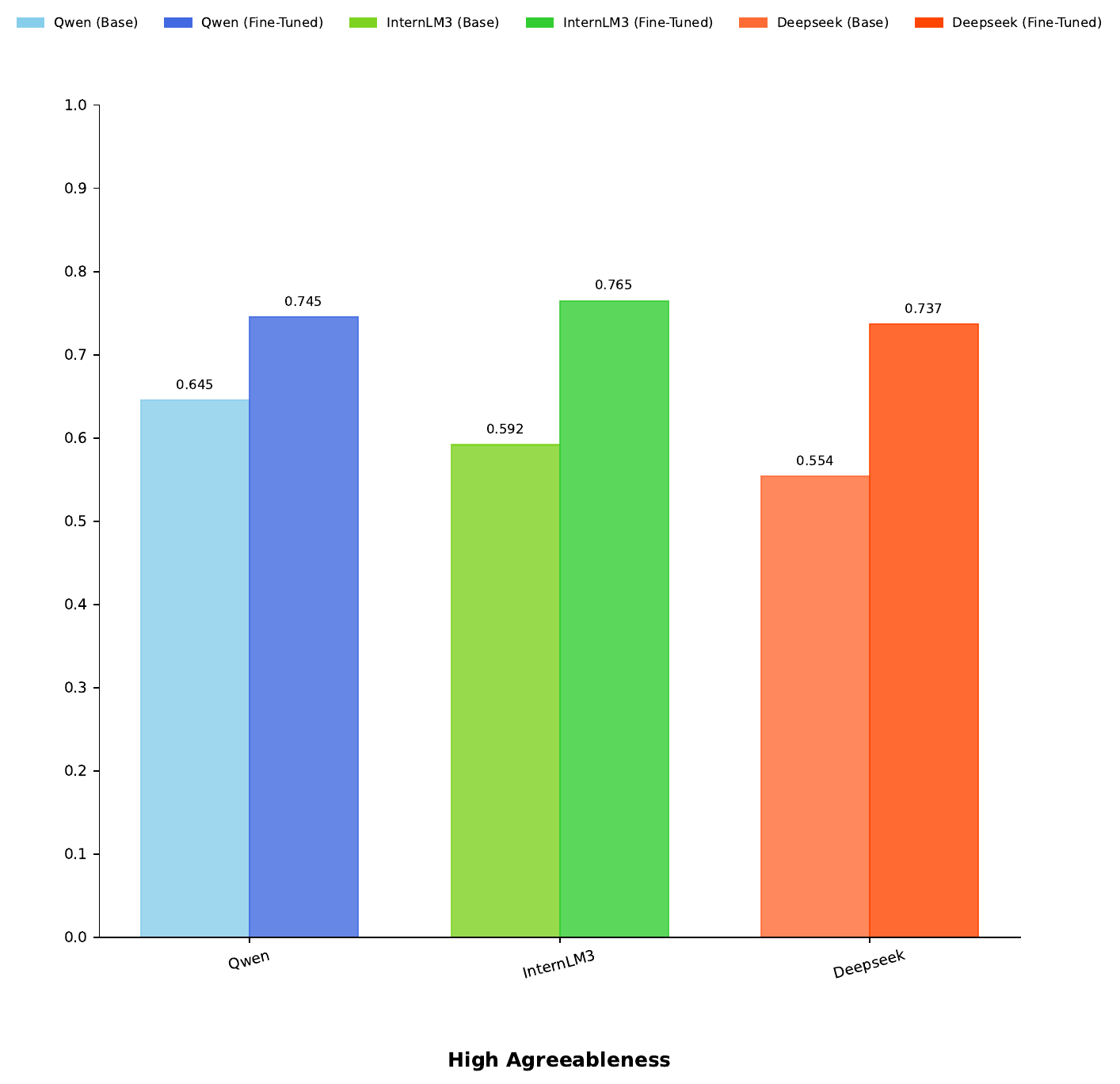}
  }\hfill
  \subcaptionbox{High\\Conscientiousness}[.19\linewidth]{%
    \includegraphics[width=\linewidth]{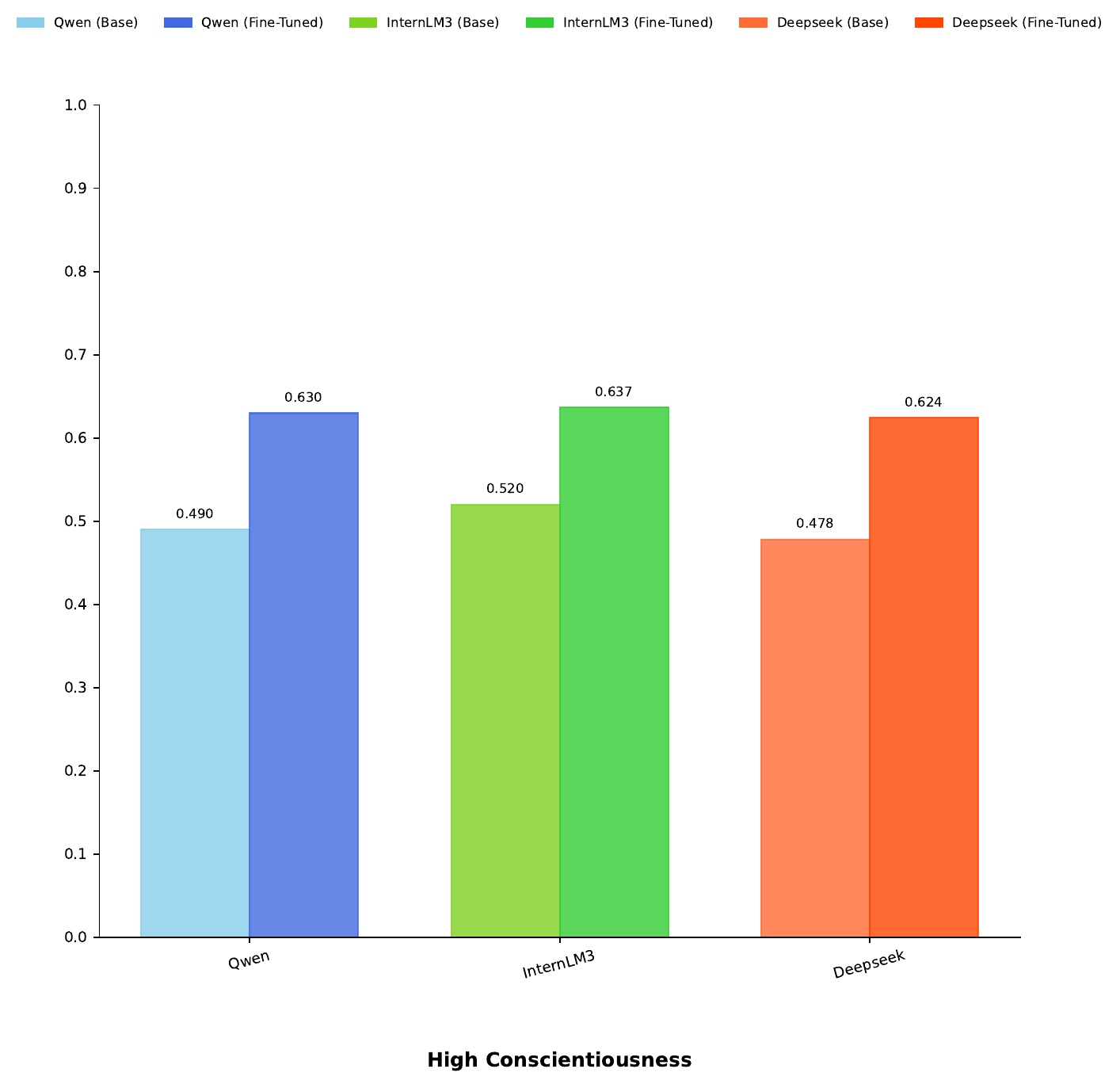}
  }\hfill
  \subcaptionbox{High\\Extraversion}[.19\linewidth]{%
    \includegraphics[width=\linewidth]{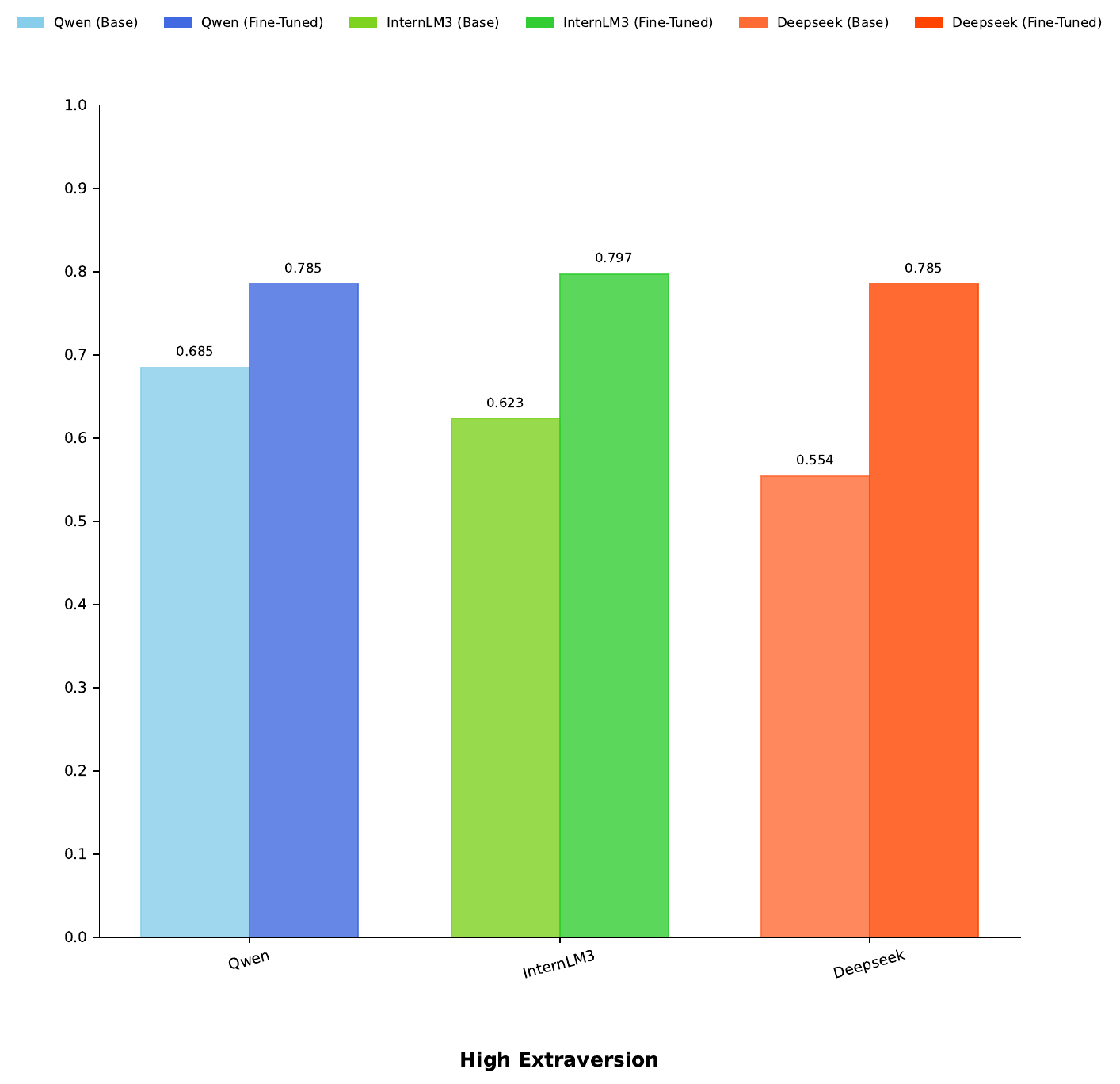}
  }\hfill
  \subcaptionbox{High\\Neuroticism}[.19\linewidth]{%
    \includegraphics[width=\linewidth]{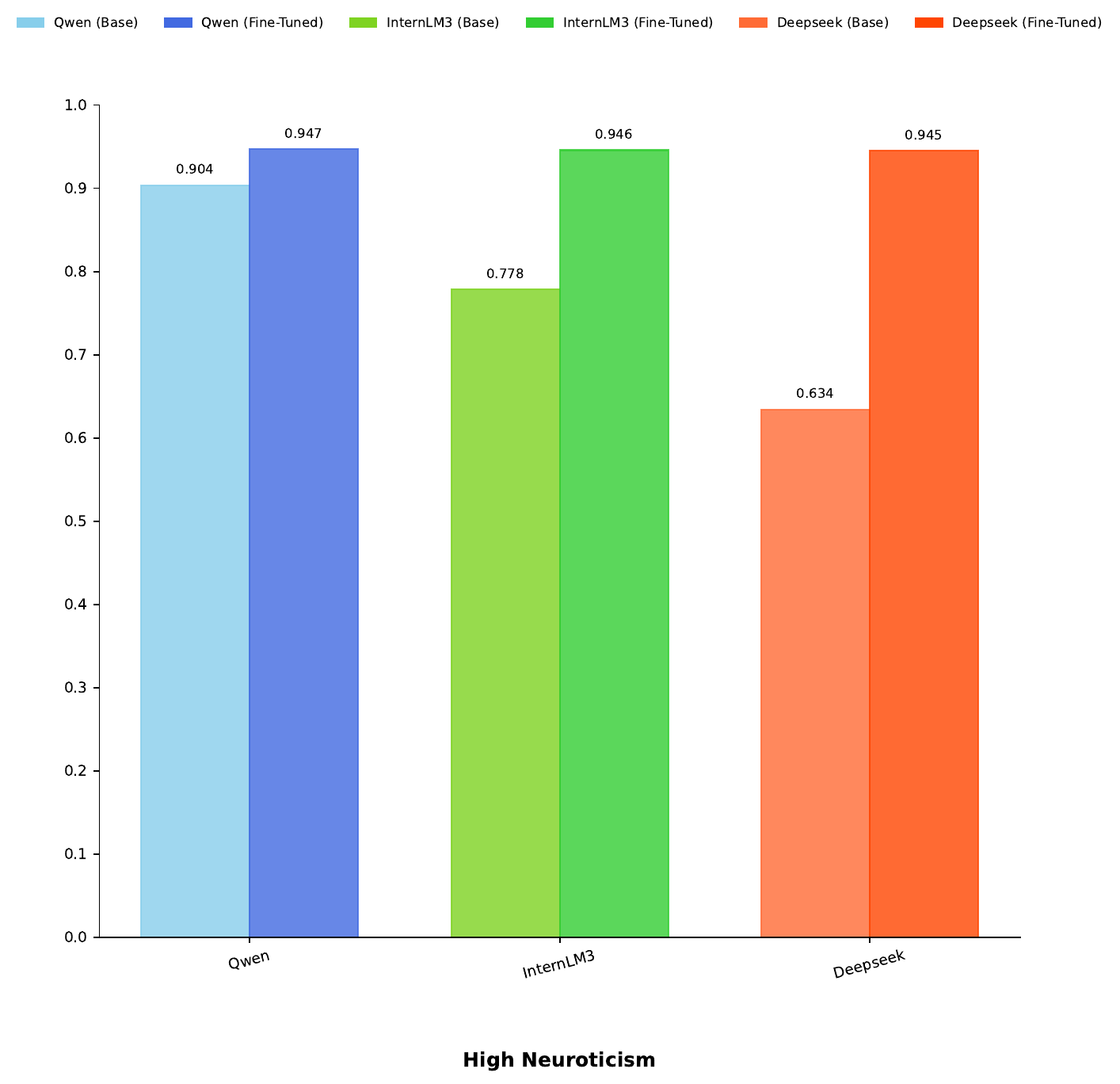}
  }\hfill
  \subcaptionbox{High\\Openness}[.19\linewidth]{%
    \includegraphics[width=\linewidth]{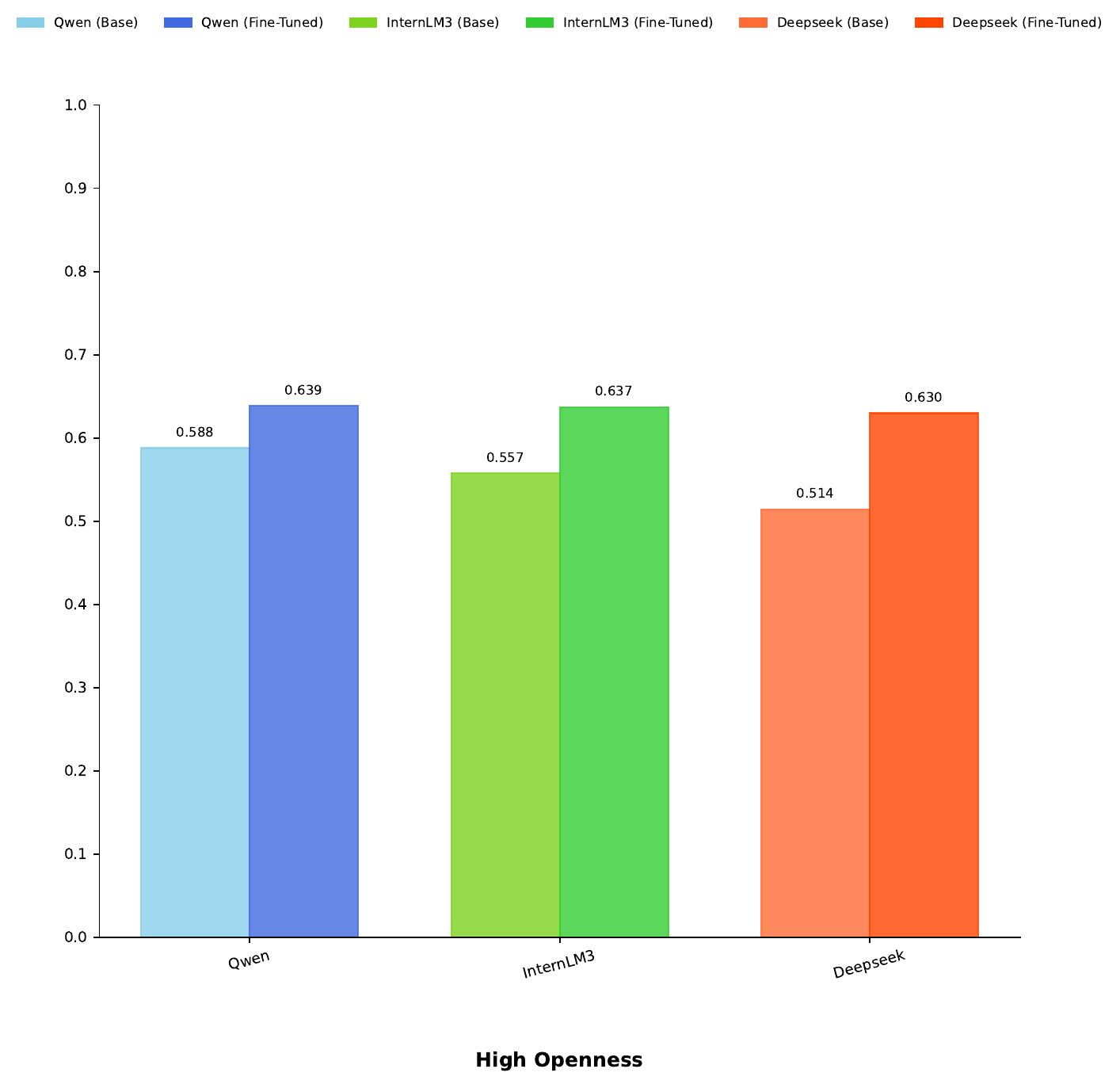}
  }

  \par\medskip
  \centering

  % Second row: LA, LC, LE, LN, LO
  \subcaptionbox{Low\\Agreeableness}[.19\linewidth]{%
    \includegraphics[width=\linewidth]{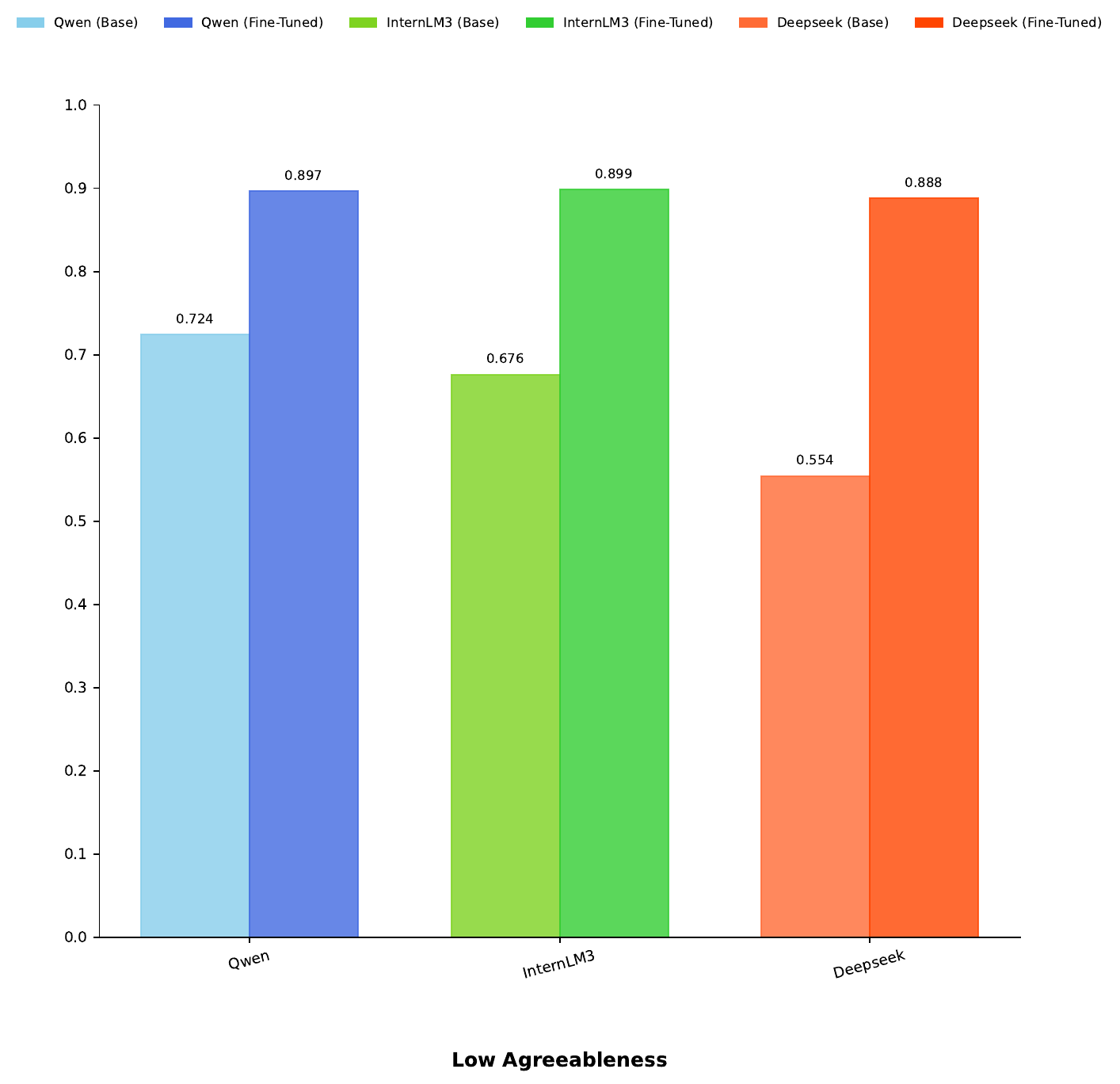}
  }\hfill
  \subcaptionbox{Low\\Conscientiousness}[.19\linewidth]{%
    \includegraphics[width=\linewidth]{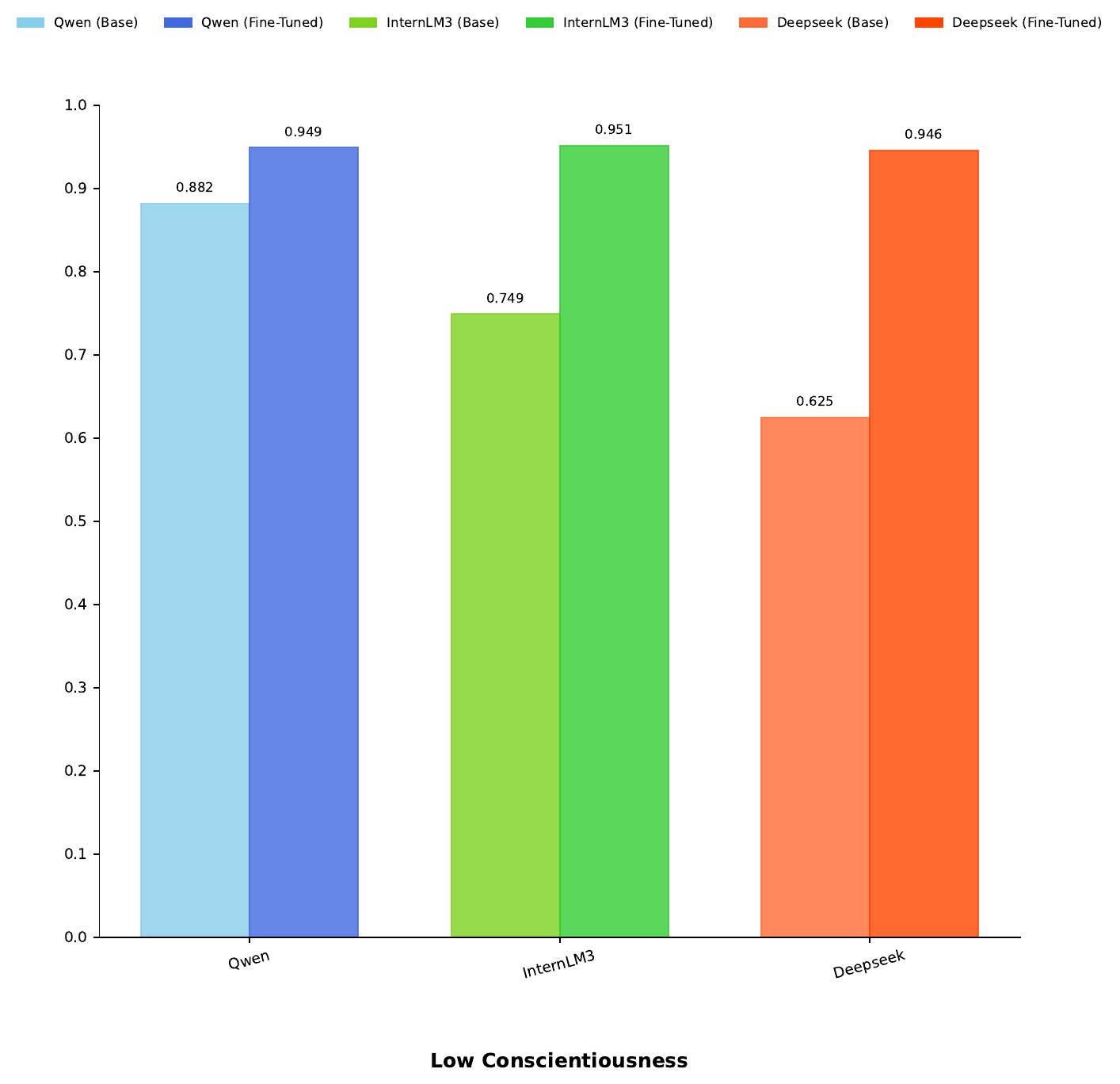}
  }\hfill
  \subcaptionbox{Low\\Extraversion}[.19\linewidth]{%
    \includegraphics[width=\linewidth]{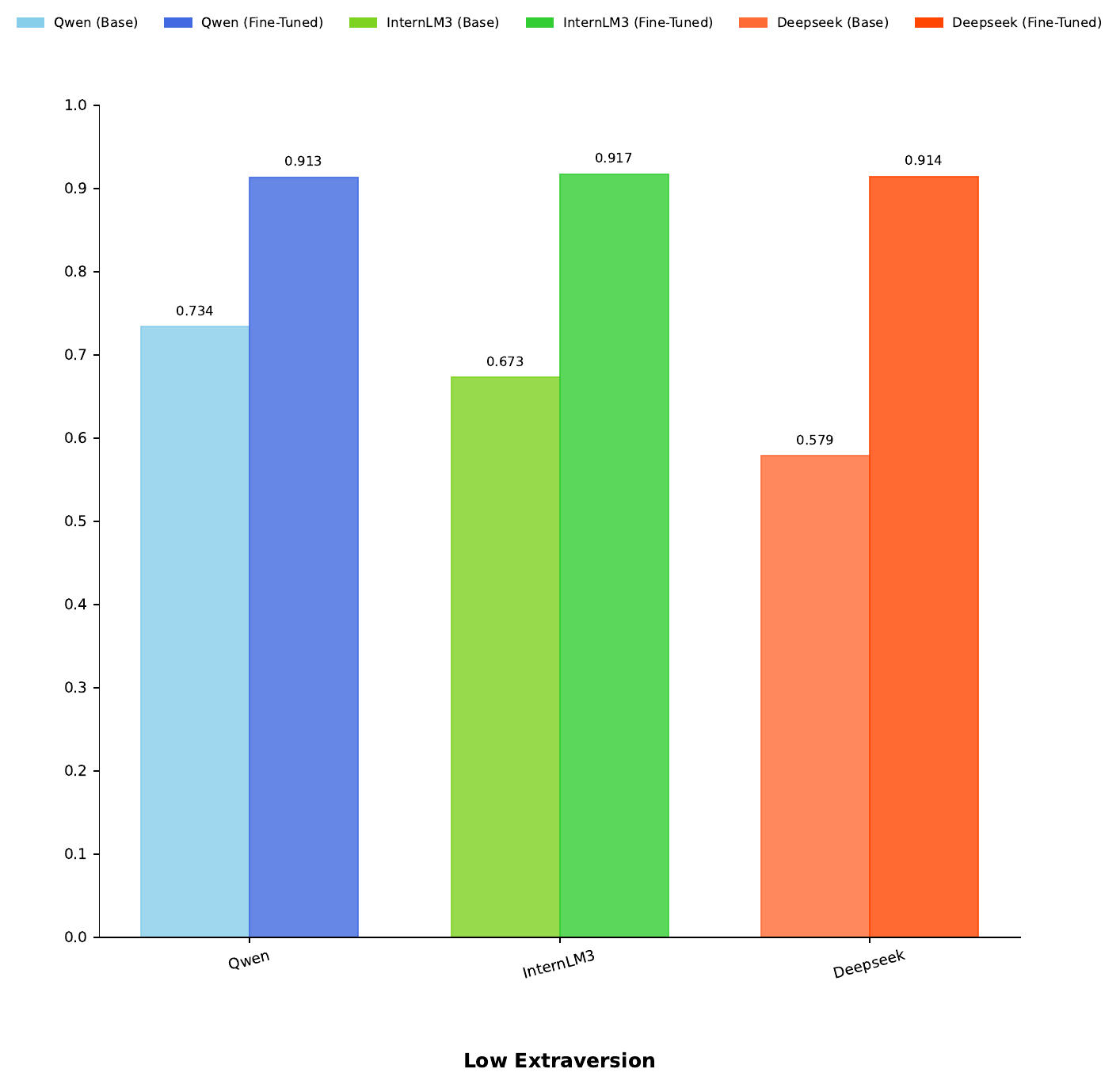}
  }\hfill
  \subcaptionbox{Low\\Neuroticism}[.19\linewidth]{%
    \includegraphics[width=\linewidth]{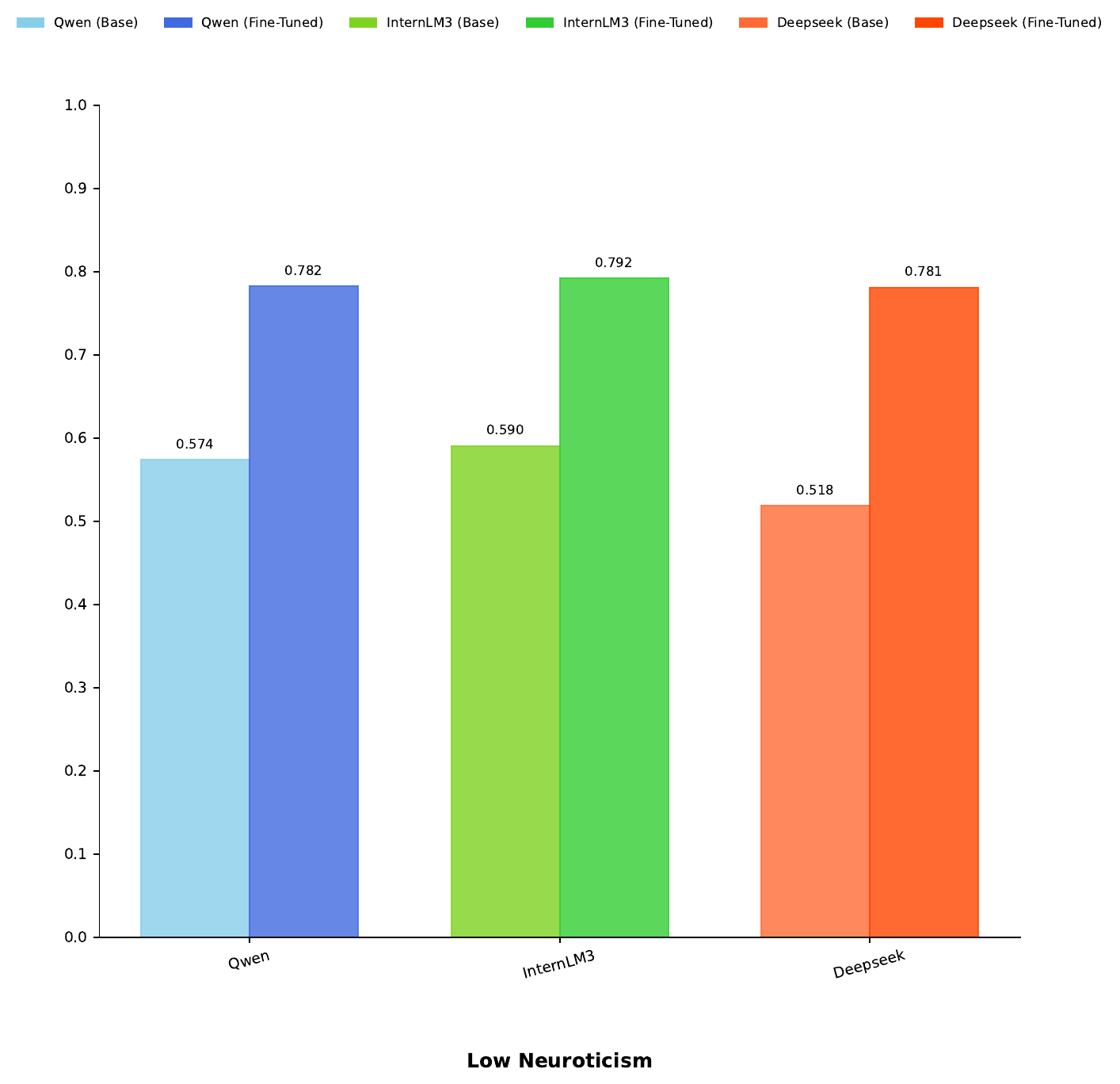}
  }\hfill
  \subcaptionbox{Low\\Openness}[.19\linewidth]{%
    \includegraphics[width=\linewidth]{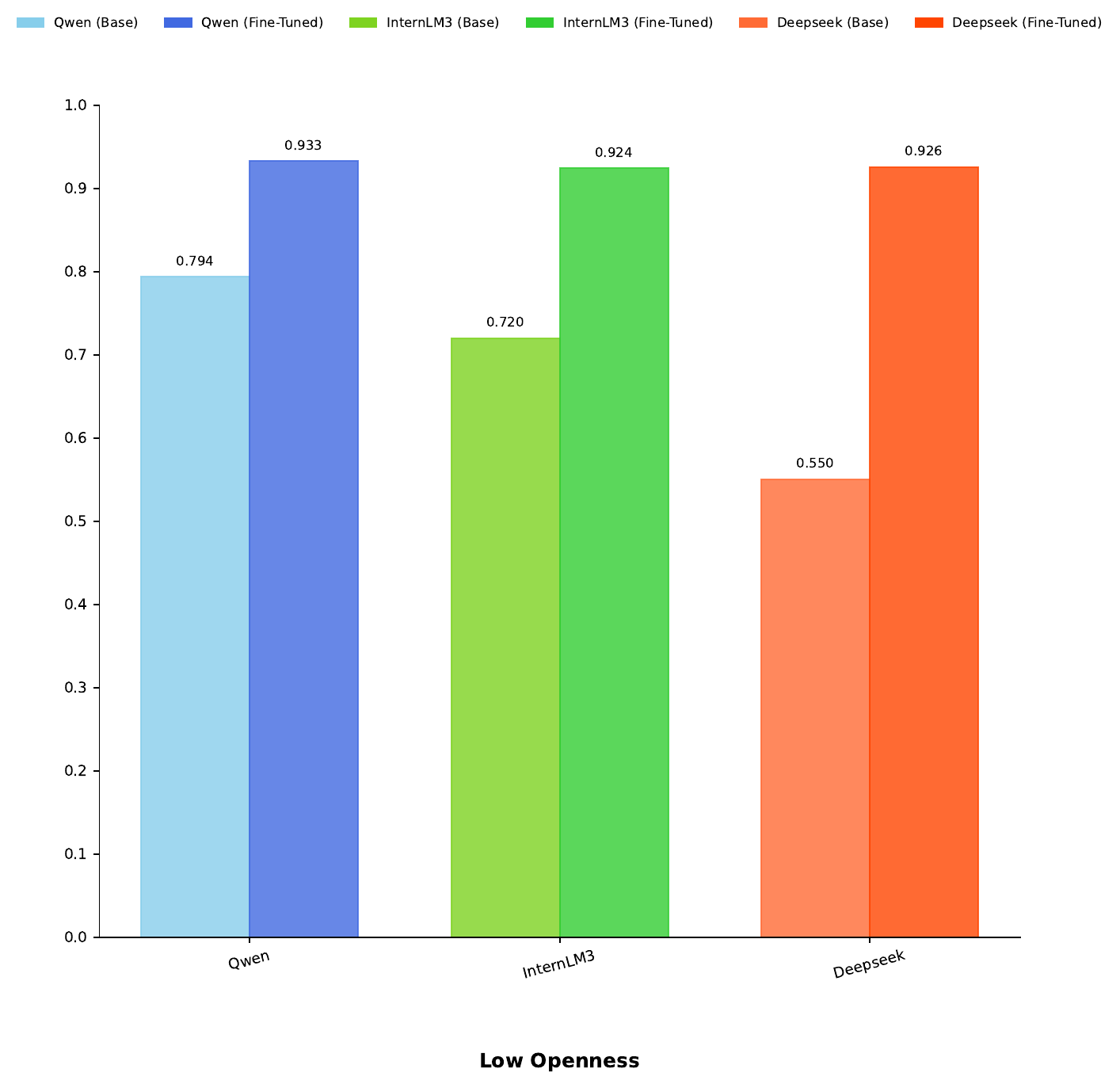}
  }

  \caption{\textbf{Task 2 — Student Realism across all personas (10 charts).} Each chart shows model performance comparison for a specific persona type, displaying realism scores across three model families (Qwen, InternLM3, DeepSeek) in both base and fine-tuned conditions.}
  \label{fig:task2-all-personas}
\end{figure*}

Fig.~\ref{fig:task2-all-personas}  reports Task~2 results for the ten Big Five–based personas across Qwen, InternLM3, and DeepSeek, under both base and fine-tuned conditions. Overall, \textbf{fine-tuning consistently improves realism across all models, with scores converging around 0.82} (DeepSeek: \textbf{0.556$\rightarrow$0.822}, InternLM3: \textbf{0.648$\rightarrow$0.821}, Qwen: \textbf{0.702$\rightarrow$0.826}).

At the persona level, the best post-tuning alignment is observed for High Neuroticism (HN, 0.891), Low Neuroticism (LN, 0.879), and Low Openness (LO, 0.871), reflecting their ability to naturally capture uncertainty and emotional variation. By contrast, High Conscientiousness (HC, 0.748) and High Openness (HO, 0.764) remain the most challenging personas, yielding the lowest realism scores even after adaptation. This suggests that highly structured traits often overlap with LLMs’ default answer-first tendency, making outputs appear more machine-like than student-like.

In terms of gains, the largest improvements occur for Low Openness (+0.176), High Neuroticism (+0.166), and Low Neuroticism (+0.156), while High Conscientiousness (+0.090) and High Openness (+0.093) improve the least. Model-wise, DeepSeek shows the most dramatic increase (+0.266) despite its low baseline, Qwen maintains a stable advantage with strong pre-tuning performance, and InternLM3 improves moderately but converges with the others in the end. Taken together, these results indicate that \textbf{student realism is most effectively enhanced through personas reflecting natural uncertainty or variability, while structured or idealized personas remain difficult to simulate authentically}.

\vspace{6pt}
\noindent\textbf{Cross-task linkage with Task~1.}
Relating Task~1 (basic coherence) and Task~2 (student realism) reveals complementary insights. Task~1 captures low-level observable alignment signals, while Task~2 evaluates higher-level subjective perception.

First, response coverage from Task~1 directly constrains realism in Task~2. For instance, InternLM3 maintains a low post-tuning response rate (0.4261), which limits its realism score compared to Qwen and DeepSeek, despite improvement.

Second, the dimension-level difficulty ordering in Task~1 aligns with persona-level differences in Task~2: Emotion is the easiest in Task~1 (OverallAcc=0.65), matching the high realism scores of High/Low Neuroticism and Low Openness personas; Behavior is the hardest (OverallAcc=0.355), consistent with the poor realism of High Conscientiousness and High Openness personas, which demand strict adherence to classroom norms.

Finally, both tasks exhibit post-tuning convergence across families but at different levels: Qwen and DeepSeek converge at OverallAcc=0.62 in Task~1, with InternLM3 trailing due to limited response coverage; in Task~2, all three converge tightly around 0.82. This indicates that EduPersona exerts stronger corrective effects on high-level perception, while low-level structural bottlenecks remain.

In summary, the additional analysis of Task~2 demonstrates that \textbf{basic coherence is a prerequisite but not sufficient for student realism; achieving realism further depends on the authentic reproduction of persona-specific behaviors and classroom dynamics}. Future improvements should emphasize training data and strategies that incorporate “imperfect, human-like student behaviors” to simultaneously enhance observable alignment and perceived authenticity.

\clearpage

\subsection{Additional Analysis for Task 3}

\label{subapp:task3_figs}

\begin{figure*}[!htbp]
  \centering
  \captionsetup[subfigure]{font=small,justification=centering,singlelinecheck=false}
  
  % First row: HA, HC, HE, HN, HO
  \subcaptionbox{High\\Agreeableness}[.19\linewidth]{%
    \includegraphics[width=\linewidth]{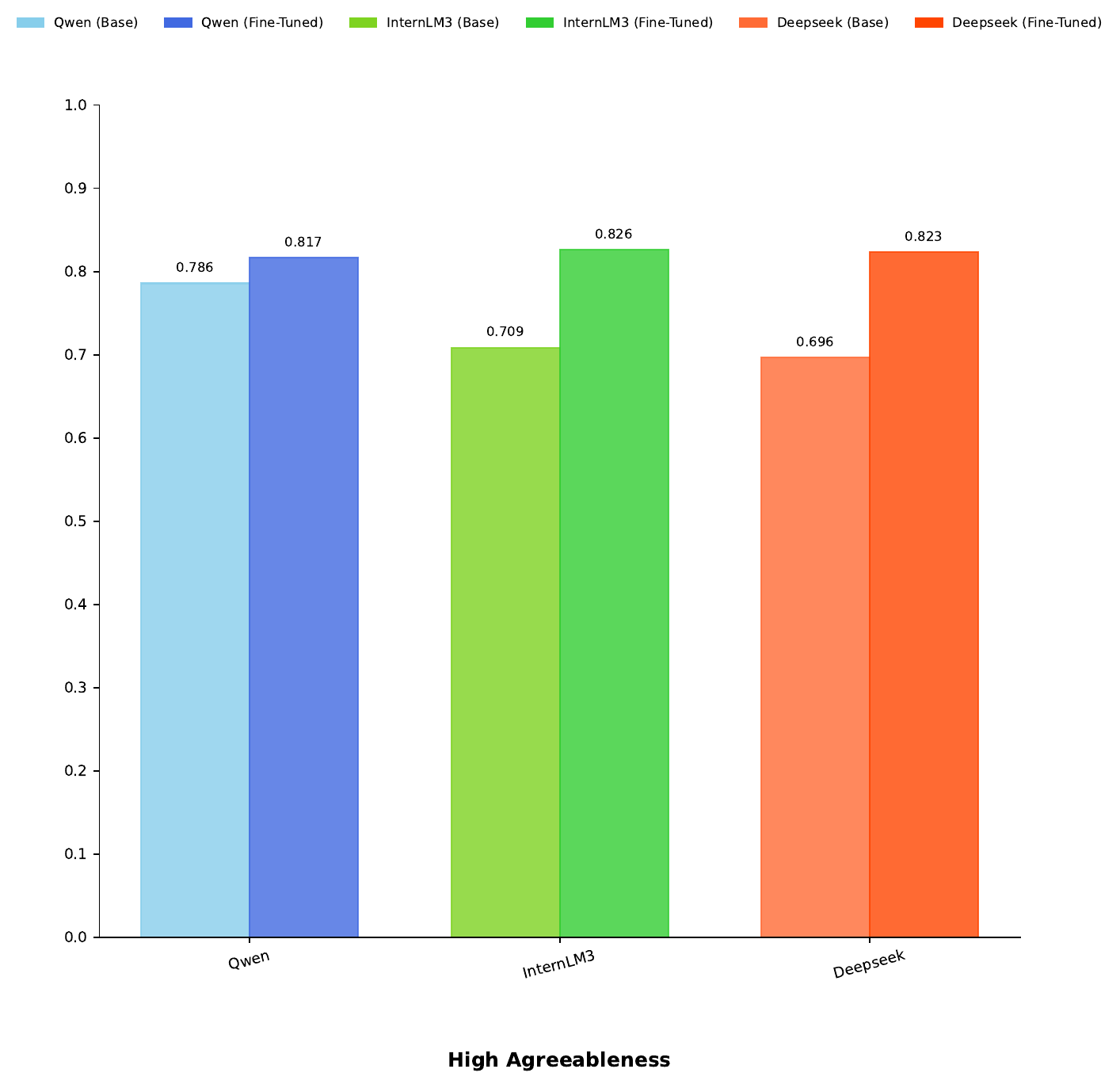}
  }\hfill
  \subcaptionbox{High\\Conscientiousness}[.19\linewidth]{%
    \includegraphics[width=\linewidth]{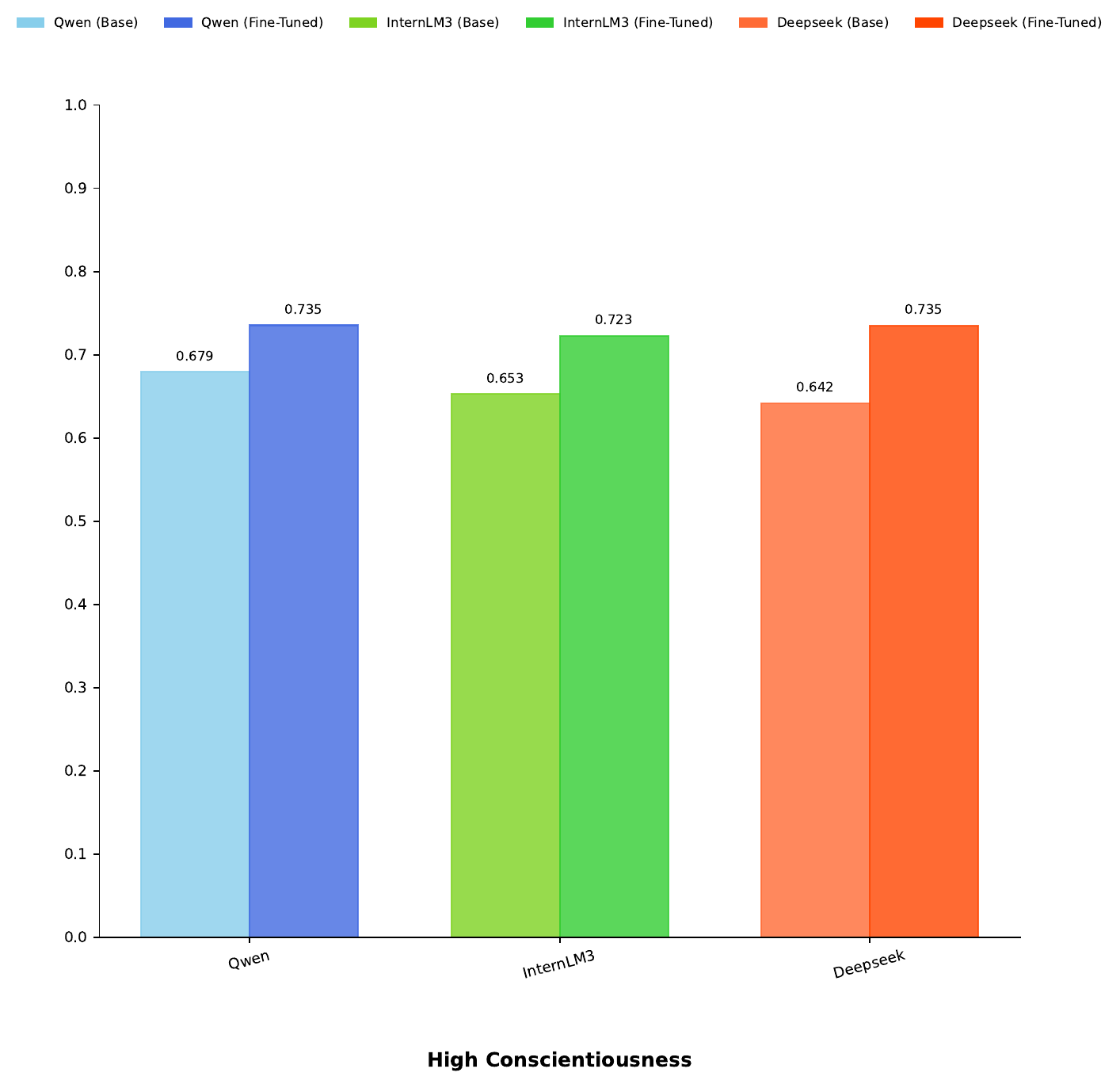}
  }\hfill
  \subcaptionbox{High\\Extraversion}[.19\linewidth]{%
    \includegraphics[width=\linewidth]{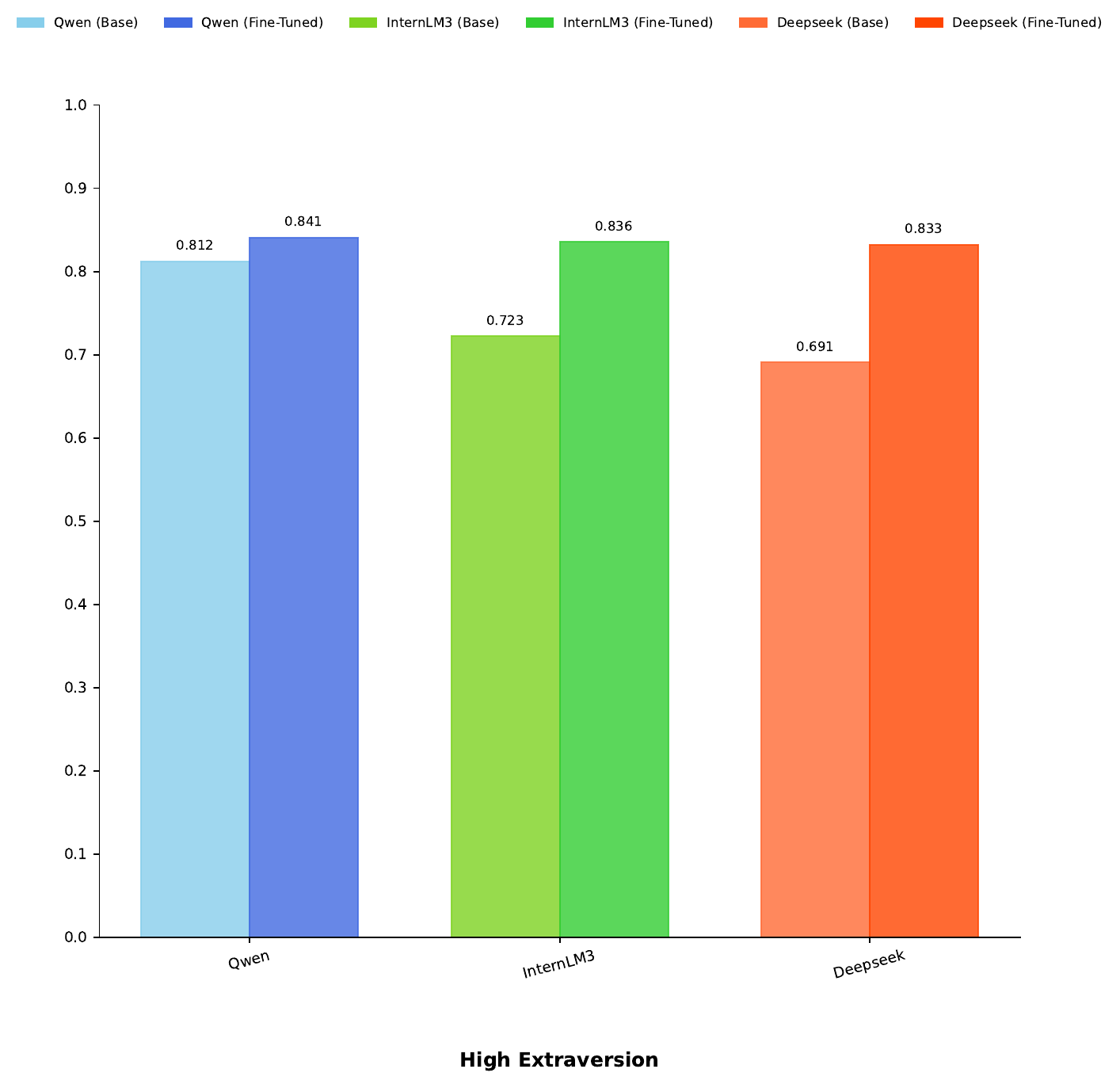}
  }\hfill
  \subcaptionbox{High\\Neuroticism}[.19\linewidth]{%
    \includegraphics[width=\linewidth]{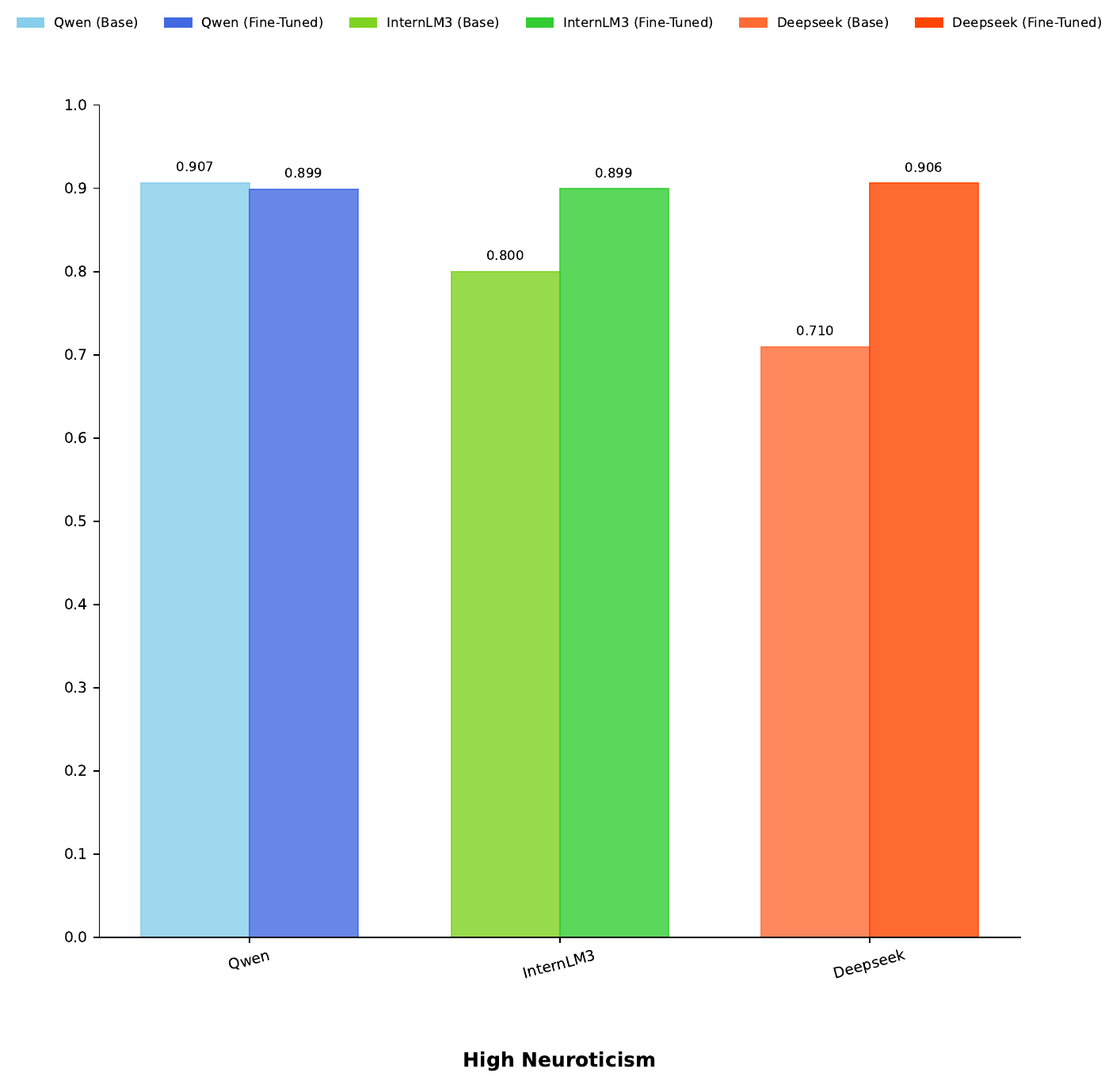}
  }\hfill
  \subcaptionbox{High\\Openness}[.19\linewidth]{%
    \includegraphics[width=\linewidth]{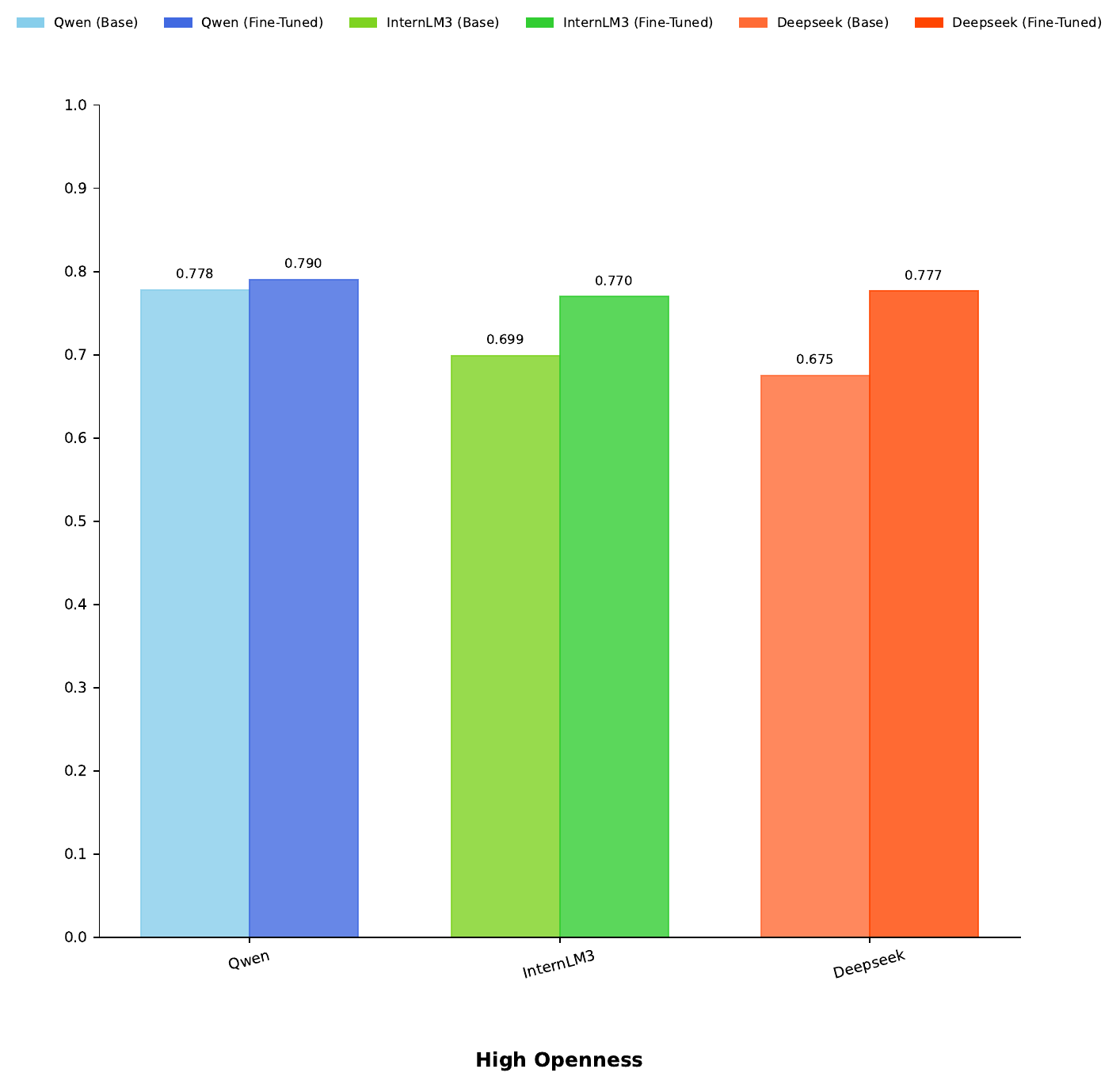}
  }

  \par\medskip
  \centering

  % Second row: LA, LC, LE, LN, LO
  \subcaptionbox{Low\\Agreeableness}[.19\linewidth]{%
    \includegraphics[width=\linewidth]{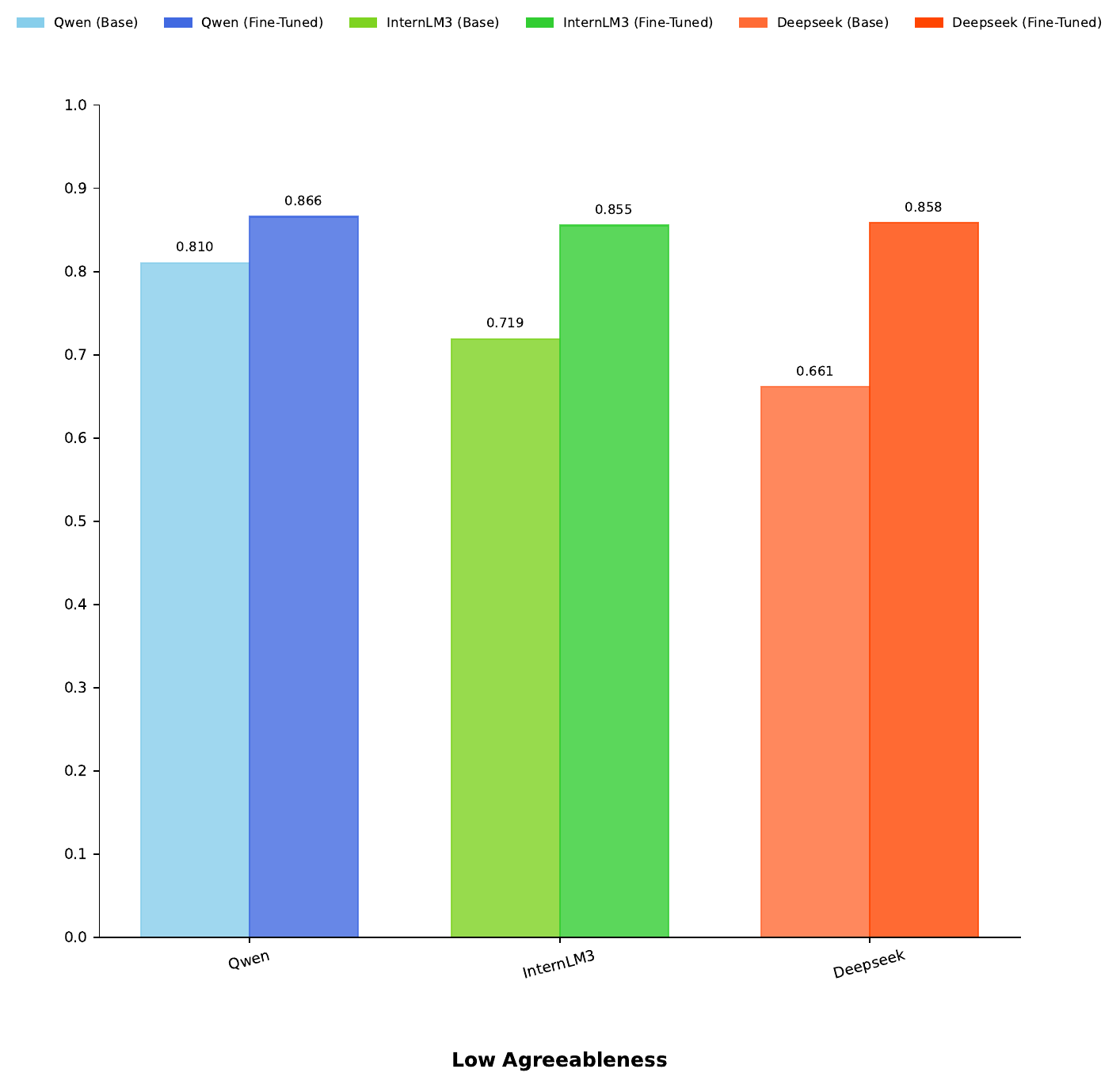}
  }\hfill
  \subcaptionbox{Low\\Conscientiousness}[.19\linewidth]{%
    \includegraphics[width=\linewidth]{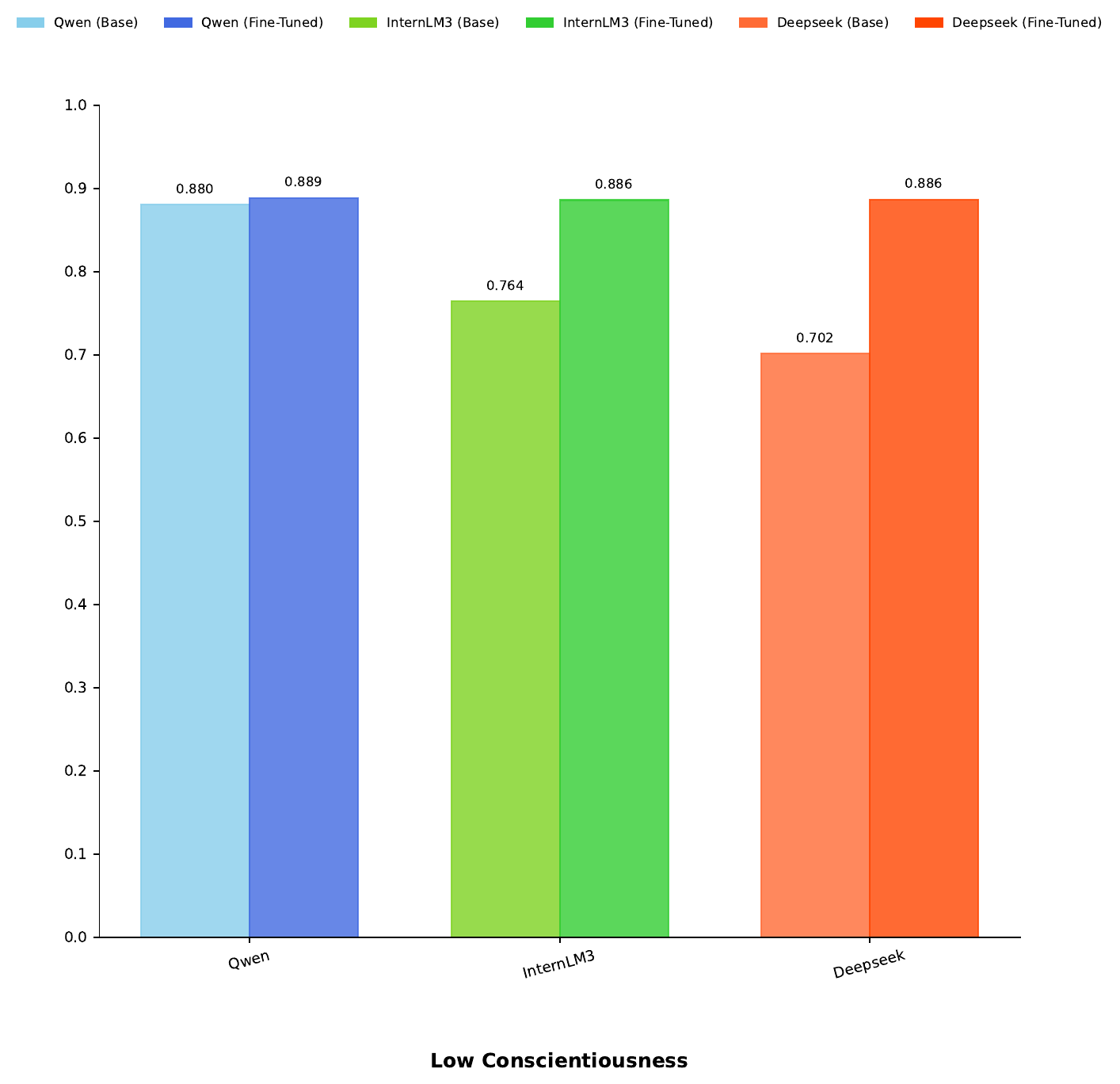}
  }\hfill
  \subcaptionbox{Low\\Extraversion}[.19\linewidth]{%
    \includegraphics[width=\linewidth]{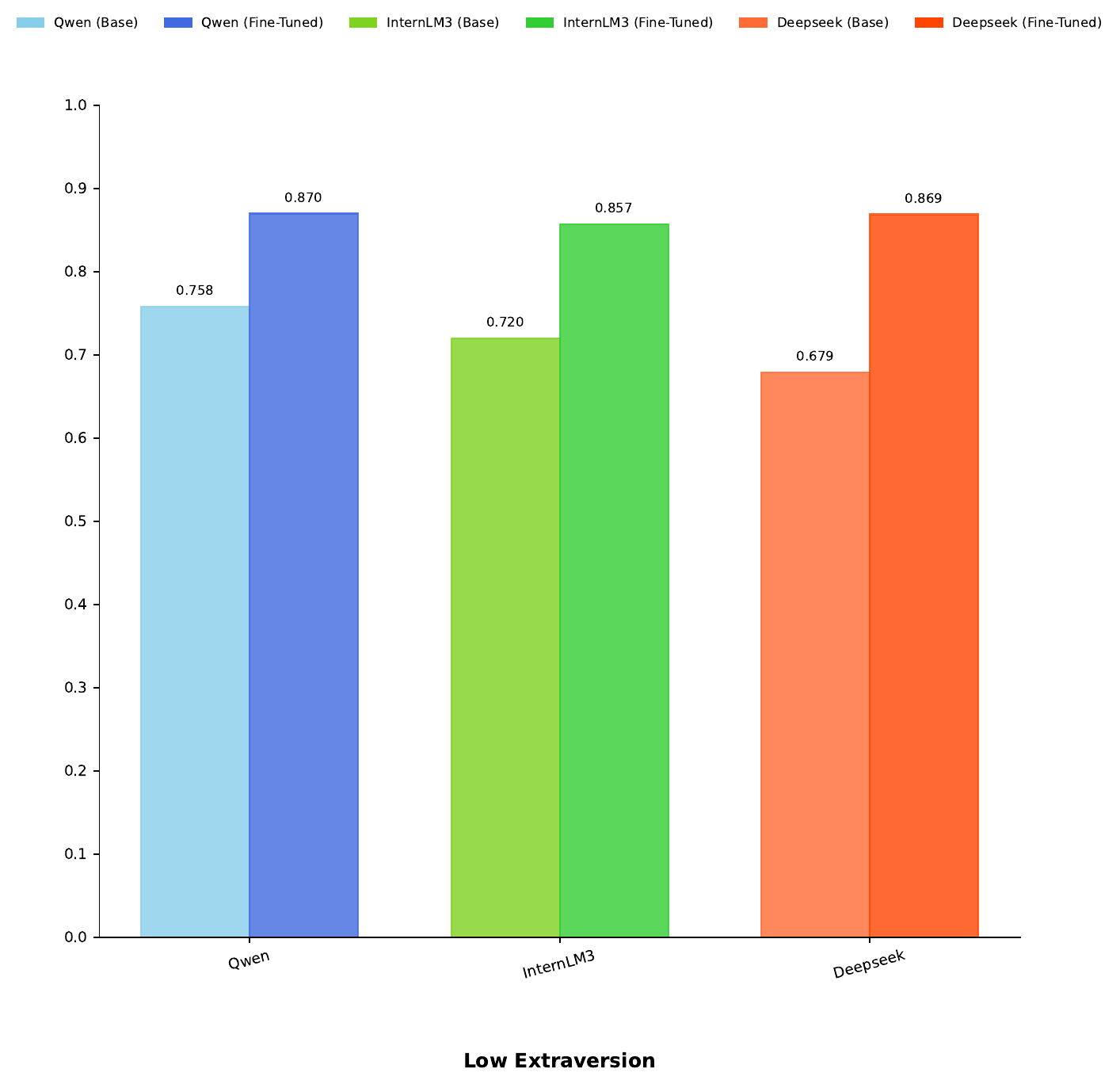}
  }\hfill
  \subcaptionbox{Low\\Neuroticism}[.19\linewidth]{%
    \includegraphics[width=\linewidth]{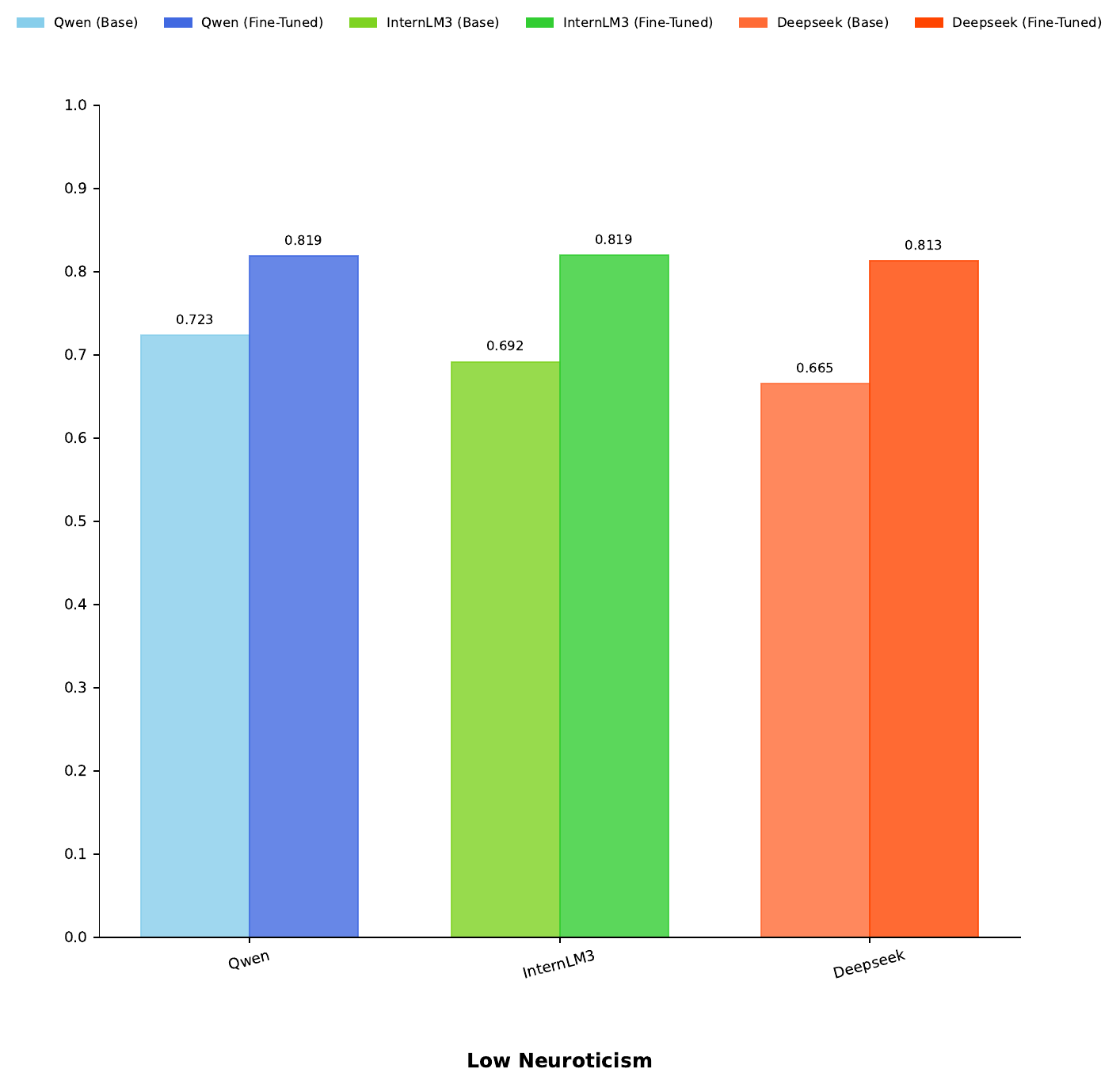}
  }\hfill
  \subcaptionbox{Low\\Openness}[.19\linewidth]{%
    \includegraphics[width=\linewidth]{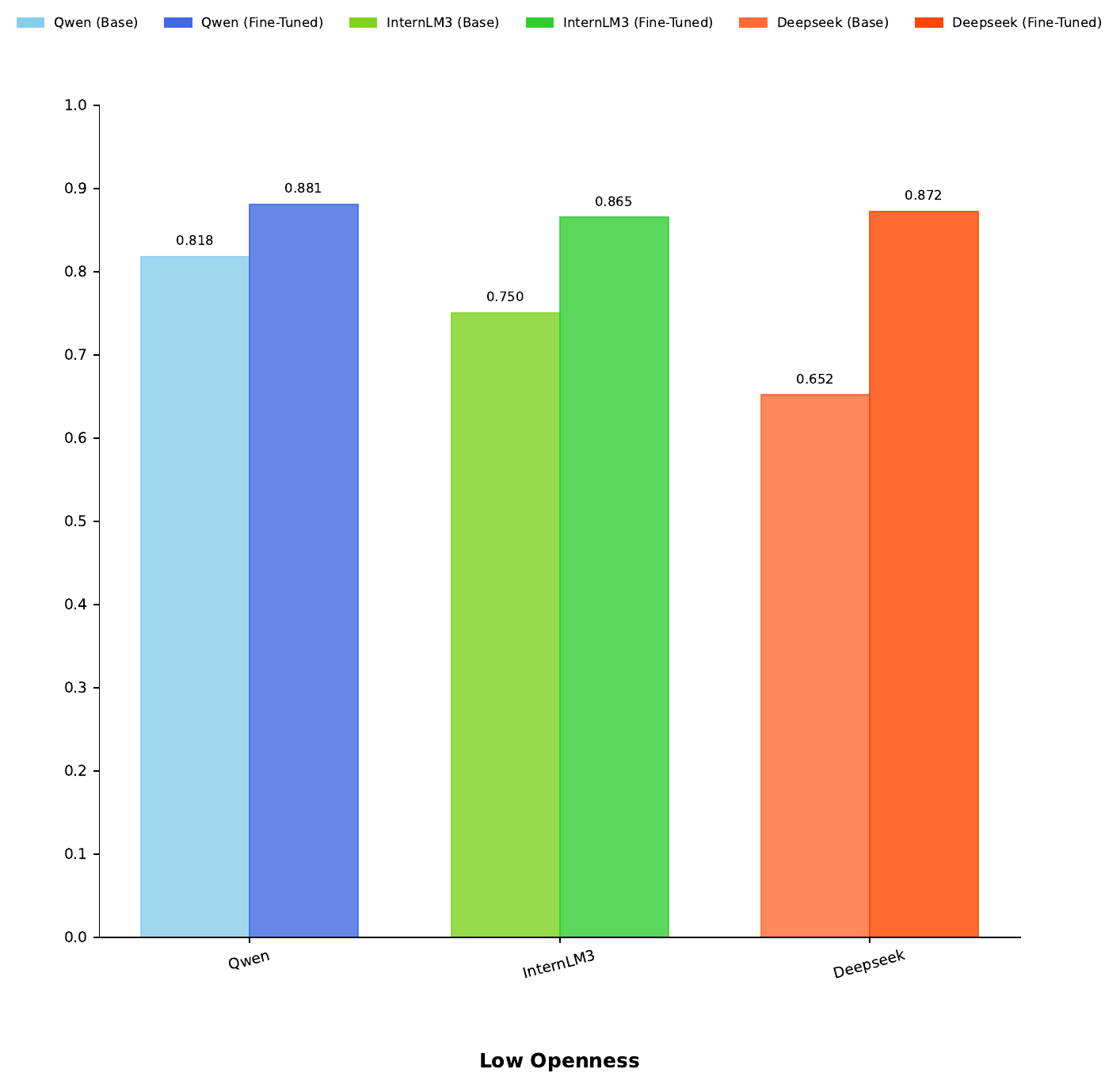}
  }

  \caption{\textbf{Task 3 — Persona Consistency across all personas (10 charts).} Each chart shows model performance comparison for a specific persona type, displaying persona consistency scores across three model families (Qwen, InternLM3, DeepSeek) in both base and fine-tuned conditions.}
  \label{fig:task3-all-personas}
\end{figure*}

Fig.~\ref{fig:task3-all-personas}  presents persona consistency results across the ten personas and three model families, comparing base and fine-tuned conditions. Overall, fine-tuning consistently improves consistency scores and brings the three families to a converged range around \textbf{0.84} (DeepSeek: \textbf{0.677$\rightarrow$0.837}, InternLM3: \textbf{0.723$\rightarrow$0.834}, Qwen: \textbf{0.795$\rightarrow$0.841}). This mirrors the findings from Task 1 and Task 2: in Task 1, Qwen and DeepSeek converged in basic coherence after fine-tuning, while in Task 2, all models reached similar levels of student realism. Together, these results confirm that EduPersona fine-tuning reliably improves model performance across layers of subjective ability while reducing inter-model disparities.

At the persona level, a stable difficulty hierarchy emerges. \textit{High Neuroticism} (0.901), \textit{Low Conscientiousness} (0.887), and \textit{Low Openness} (0.873) achieve the highest post-tuning consistency, reflecting hesitation, partial answers, or self-corrections that align well with authentic student behavior. In contrast, \textit{High Conscientiousness} (0.731) and \textit{High Openness} (0.779) remain the most challenging to sustain, even after fine-tuning. This pattern echoes Task 2, where the same personas also scored lowest in student realism, indicating that structured, idealized personas are consistently difficult for models to simulate both authentically and consistently.

In terms of improvement magnitude, the largest gains occur for \textit{Low Extraversion} (+0.146), \textit{Low Openness} (+0.133), and \textit{Low Agreeableness} (+0.130), while \textit{High Openness} (+0.062) and \textit{High Conscientiousness} (+0.073) improve the least. This highlights that EduPersona fine-tuning is particularly effective for enhancing “non-idealized” student traits, whereas idealized personas remain a persistent challenge.

At the model level, baseline disparities are substantial (Qwen 0.795, InternLM3 0.723, DeepSeek 0.677), but after fine-tuning they narrow dramatically to a range of 0.833–0.841. This again parallels Tasks 1 and 2, reinforcing that EduPersona fine-tuning not only boosts absolute performance but also reduces variance across both models and personas.

In summary, Task 3 demonstrates that \textbf{persona consistency is more demanding than student realism, yet EduPersona fine-tuning significantly enhances overall stability while reducing inter-model and inter-persona variance}. Cross-task comparisons reveal consistent bottlenecks (High Conscientiousness and High Openness) as well as easier-to-model traits (High Neuroticism, Low Conscientiousness, Low Openness), providing a coherent picture of how virtual student agents can be systematically improved.

\end{document}